\documentclass[fleqn,10pt]{wlscirep}

\def\maketitlesupplementaryalt
   {
   \newpage
   \onecolumn
    \begin{center}
    \Large
    \textbf{An interpretable approach to automating the assessment of biofouling in video footage}\\
    \vspace{0.5em}Supplementary Material \\
    \end{center}
   }

\usepackage{soul}
\usepackage{color}

\DeclareRobustCommand{\hlcyan}[1]{{\sethlcolor{cyan}\hl{#1}}}
\DeclareRobustCommand{\hlred}[1]{{\sethlcolor{red}\hl{#1}}}
\DeclareRobustCommand{\hlgreen}[1]{{\sethlcolor{green}\hl{#1}}}
\DeclareRobustCommand{\hlblue}[1]{{\sethlcolor{blue}\hl{#1}}}
\DeclareRobustCommand{\hlyellow}[1]{{\sethlcolor{yellow}\hl{#1}}}

\usepackage{placeins}

\usepackage[capitalise]{cleveref}
\usepackage{lineno}

\usepackage[utf8]{inputenc}
\usepackage[T1]{fontenc}

\usepackage[percent]{overpic}

\usepackage{trimclip}
\newcommand\picdims[4][]{%
  \setbox0=\hbox{\includegraphics[#1]{#4}}%
  \clipbox{.5\dimexpr\wd0-#2\relax{} %
           .5\dimexpr\ht0-#3\relax{} %
           .5\dimexpr\wd0-#2\relax{} %
           .5\dimexpr\ht0-#3\relax}{\includegraphics[#1]{#4}}}
\usepackage{graphicx}
\usepackage{booktabs}

\usepackage{caption}
\usepackage{subcaption}

\usepackage{tikz}
\usepackage{amsmath}
\usepackage{bm}
\usepackage{mathtools}
\usetikzlibrary{shapes,arrows}

\title{An interpretable approach to automating the assessment of biofouling in video footage}

\author[1,2,*]{Evelyn J. Mannix}
\author[3]{Bartholomew A. Woodham}
\affil[1]{Centre of Excellence for Biosecurity Risk Analysis, The University of Melbourne, Melbourne, Australia.}
\affil[2]{Melbourne Centre for Data Science, The University of Melbourne, Melbourne, Australia.}
\affil[3]{Biosecurity Animal Division, Department of Agriculture, Fisheries and Forestry, Canberra, Australia.}

\affil[*]{evelyn.mannix@unimelb.edu.au}

\begin{abstract}
Biofouling---communities of organisms that grow on hard surfaces immersed in water---provides a pathway for the spread of invasive marine species and diseases. To address this risk, international vessels are increasingly being obligated to provide evidence of their biofouling management practices. Verification that these activities are effective requires underwater inspections, using divers or underwater remotely operated vehicles (ROVs), and the collection and analysis of large amounts of imagery and footage. Automated assessment using computer vision techniques can significantly streamline this process, and this work shows how this challenge can be addressed efficiently and effectively using the interpretable Component Features (ComFe) approach with a DINOv2 Vision Transformer (ViT) foundation model. ComFe is able to obtain improved performance in comparison to previous non-interpretable Convolutional Neural Network (CNN) methods, with significantly fewer weights and greater transparency---through identifying which regions of the image contribute to the classification, and which images in the training data lead to that conclusion. All code, data and model weights are publicly released.

\end{abstract}

\begin{document}

\flushbottom
\maketitle
%
%
\thispagestyle{empty}

\section{Introduction}

The spread of exotic marine organisms and diseases pose a severe threat to people's way of life, the global environment, and local economies.\cite{international2012guidelines}  Biofouling on vessels travelling internationally is a significant pathway for the anthropogenic spread and introduction of these species,\cite{georgiades2021role} and can consist of a wide range of organisms adapted to colonising hard surfaces in water, such as barnacles, tubeworms and algae.\cite{callow2002marine} While the shipping industry prioritises the management of biofouling to reduce fuel costs,\cite{schultz2011economic} the incentives to manage biofouling for hydrodynamic purposes do not provide sufficient coverage of biosecurity risks.\cite{davidson2016mini, zabin2018will}

Several jurisdictions, such as Australia, New Zealand and California,\cite{DAFF2023biofouling, scianni2021yes} require internationally arriving vessels to provide pre-arrival information that documents the steps they have taken to manage their biofouling. By encouraging proactive biofouling management, the associated biosecurity risk from vessels to these countries is reduced. However, to assess the efficacy of management practices or provide evidence that a vessel does not pose an unacceptable biosecurity risk, the underwater hull of the vessel may need to be inspected. Further, incoming vessels may request or be asked to provide recent underwater footage that supports that their hull is clean and effectively managed.

Manual assessment of footage from biofouling inspections by the appropriate decision maker can be a resource intensive exercise. There can also be a large degree of variability in biofouling assessment, with different experts having different thresholds for particular biofouling categories.\cite{mannix2021automating} Previous work\cite{krause2023semantic} has shown that it is feasible to use automated approaches to identify the presence and severity of biofouling on vessel hulls. However, these approaches are often not designed to identify fouling on commericial vessels---rather they are designed to identify particular groups of fouling organisms on test plates, that are used to study the behaviour of biofouling on surfaces. While there has been some work aiming to detect biofouling on vessel hulls, this has only focused on detecting biofouling in still images, and the approach was not extended to video data.\cite{mannix2021automating}

Previous deep learning approaches have also been black-box methods which are unable to explain their predictions.\cite{rudin2019stop} Standard image classification approaches, such as training Convolutional Neural Networks (CNNs) with image annotations as done in prior works,\cite{mannix2021automating} can often learn to detect classes using spurious features.\cite{Neuhaus_2023_ICCV} These are features that are correlated with the variable of interest that are used by a model to artificially improve training accuracy. In training a model to detect biofouling, these spurious features could be niche areas like sea chests or antifoulant coating damage, which are both known to promote biofouling growth.\cite{moser2017quantifying} In contrast, interpretable-by-design methods can (i) localise which regions of an image are important for making a prediction, and (ii) identify representative images from the training data that lead to a particular categorisation.\cite{chen2019looks} This can improve the trust of a biosecurity manager in the model and provide opportunities to identify flaws---such as if training data is incorrectly labelled or spurious features are important for a prediction.

In this paper, we present the application of Component Features (ComFe),\cite{mannix2024scalable} a state-of-the-art interpretable-by-design image classification approach, to the challenge of identifying the presence of biofouling within footage of a vessel's hull. The ComFe approach takes a pretrained foundation model, such as a model from the DINOv2 family,\cite{oquab2023dinov2} and is able to identify key directions in the embedding space of this model that encode important concepts.\cite{bhalla2024interpreting} By clustering regions of the image and comparing the regional embeddings to these concepts, ComFe is able to identify where biofouling occurs in an image, and explain this by referring to key images in the training data. In this work, our main contributions are:

\begin{itemize}
    \item Presenting the first application of an interpretable computer vision model to the problem of identifying biofouling in imagery of vessel hulls, that is able to outperform previous large CNN ensembles while using significantly less parameters and being more efficient to compute.
    \item Demonstrating the utility of the model to summarize videos taken using underwater Remotely Operated Vehicles (ROVs), to enable the potential biosecurity risk from biofouling on a vessel to be quickly evaluated.
    \item The public release of the dataset used to train and evaluate the models, as well as the models themselves.
\end{itemize}

In \cref{sec:methods} we provide a high level explanation of the ComFe approach, describe the datasets used for training the models and the training process, and outline how the models are applied to video footage. In \cref{sec:results} we demonstrate the utility of ComFe for detecting biofouling in still image and video data, and discuss further considerations in \cref{sec:discussion}.

\tikzstyle{decision} = [diamond, draw, fill=blue!20, 
text width=6em, text badly centered, node distance=3cm, inner sep=0pt]
\tikzstyle{block} = [rectangle, draw, fill=blue!20, 
text width=7em, text centered, rounded corners, minimum height=4em]
\tikzstyle{blockdashed} = [rectangle, draw, dashed, 
text width=7em, text centered, rounded corners, minimum height=4em]
\tikzstyle{wideblock} = [rectangle, draw, fill=blue!20, 
text width=10em, text centered, rounded corners, minimum height=4em]
\tikzstyle{blockbare} = [text width=8.1em, text centered, minimum height=4em]
\tikzstyle{wideblockbare} = [text width=10em, text centered, minimum height=4em]
\tikzstyle{block2} = [rectangle, draw, fill=yellow!20, 
text width=9em, text centered, rounded corners, minimum height=4em]
\tikzstyle{line} = [draw, -latex']


\tikzstyle{cloud} = [draw, ellipse,fill=red!20, node distance=3cm,
minimum height=4em]

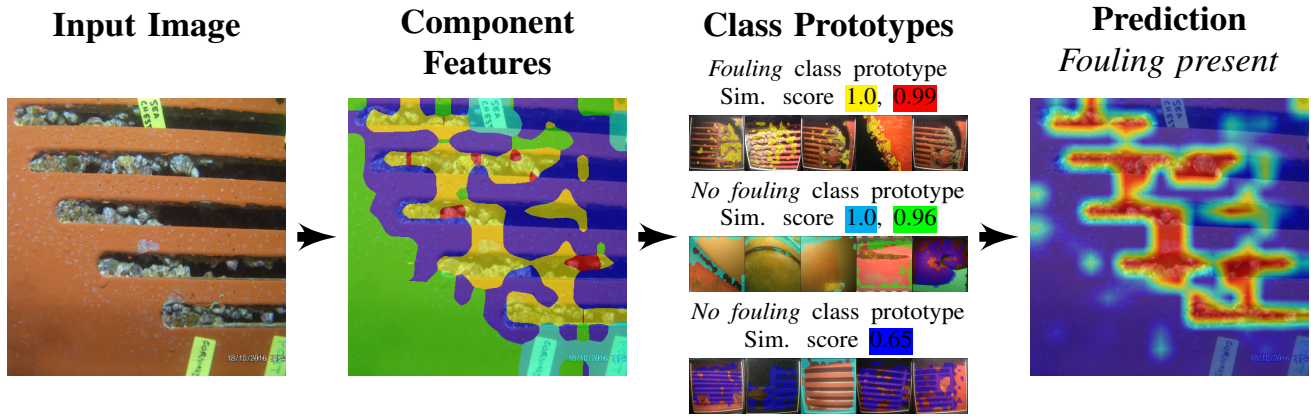
\begin{figure*}[!tb]
  \centering
  \resizebox{\textwidth}{!}{%
  \begin{tikzpicture}[node distance = 2cm, auto]


    \node [blockbare] (image) {
    \includegraphics[width=\textwidth]{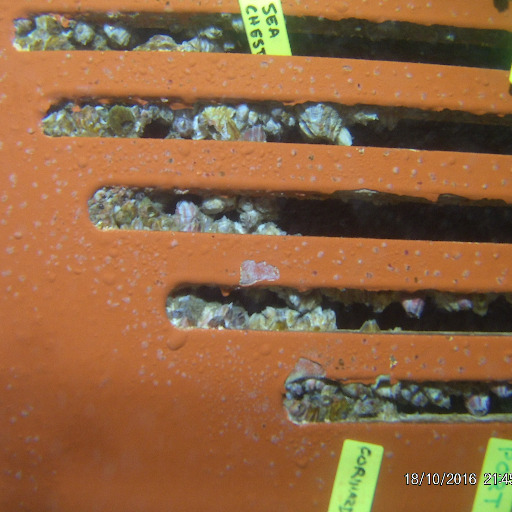}
    };
    \node [blockbare, above of = image, yshift=0.15cm] (imagetext) { \textbf{Input Image}
    };
    
    \node [blockbare, right of = image, node distance = 3.5cm] (image_prototypes) {
    \includegraphics[width=\textwidth]{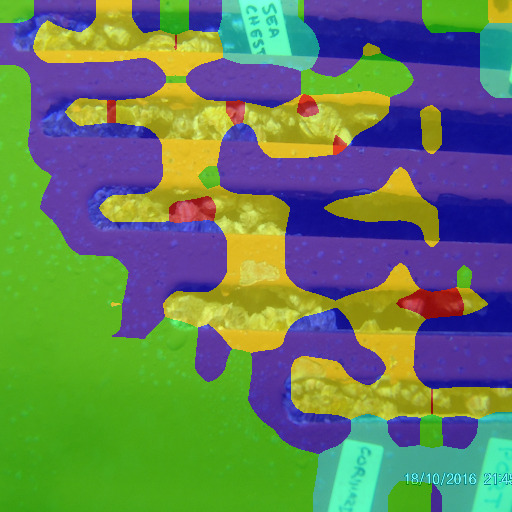}
    };
    \node [blockbare, above of = image_prototypes, yshift=0.0cm] (image_prototypestext) { \textbf{Component Features}
    };
    
    \node [blockbare, right of = image_prototypes, node distance = 3.5cm] (class_prototypes) {
    \scriptsize
    \textit{Fouling} class prototype\\
    Sim. score \hlyellow{1.0}, \hlred{0.99}\\
    \includegraphics[width=\textwidth]{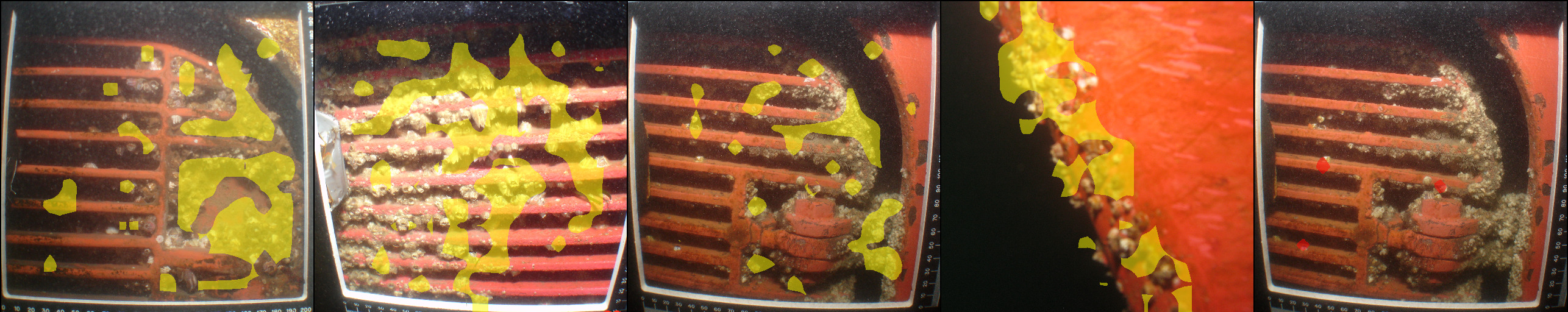}
    \textit{No fouling} class prototype\\
    Sim. score \hlcyan{1.0}, \hlgreen{0.96}\\
    \includegraphics[width=\textwidth]{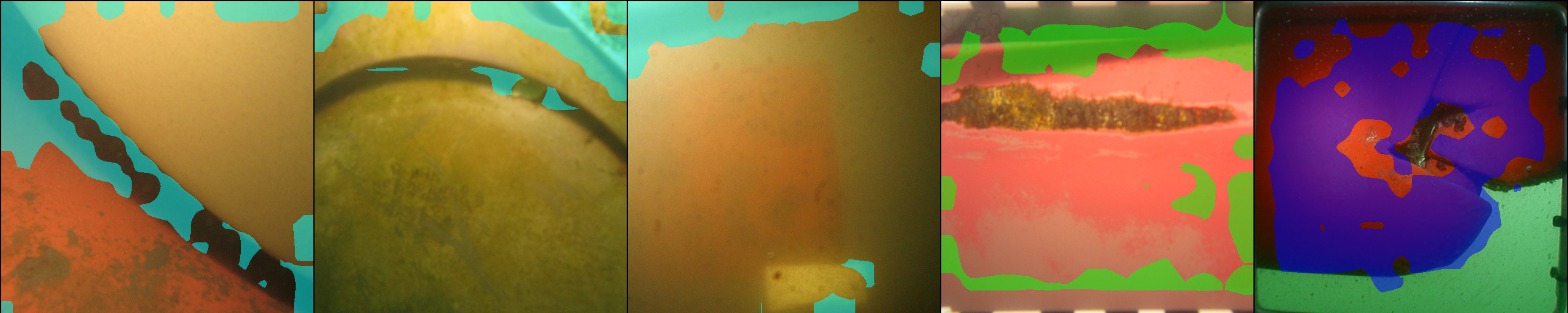}
    \textit{No fouling} class prototype\\
    Sim. score \hlblue{0.65}\\
    \includegraphics[width=\textwidth]{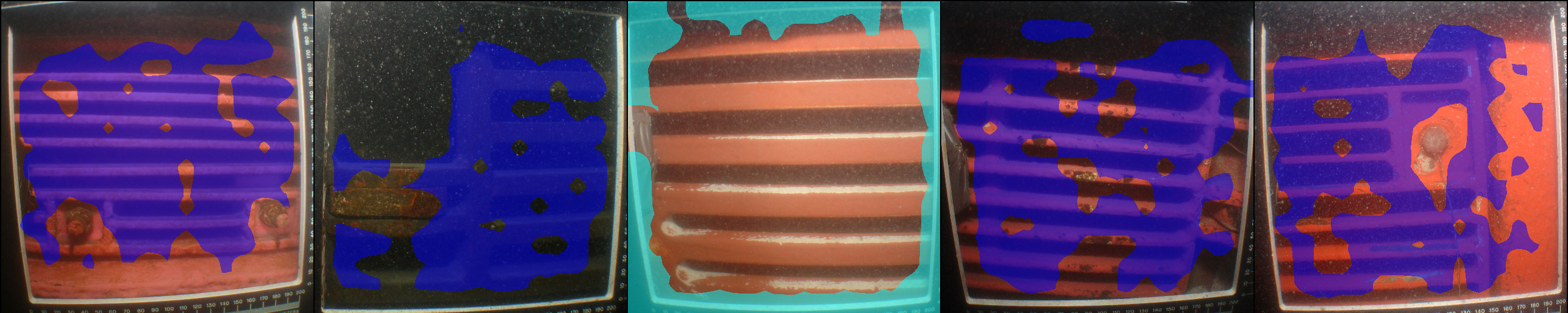}
    };
    \node [blockbare, above of = class_prototypes, yshift=0.15cm] (classtext) {
    \textbf{Class Prototypes} \par
    };
    
    \node [blockbare, right of = class_prototypes, node distance = 3.5cm] (class) {
    \includegraphics[width=\textwidth]{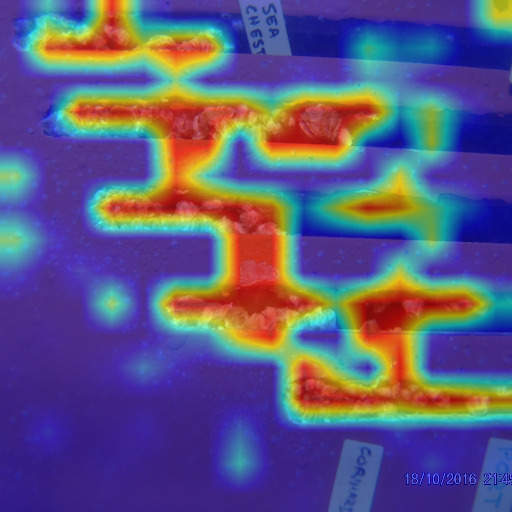}
    };
    \node [blockbare, above of = class, yshift=-0.0cm] (classtext) { \textbf{Prediction}\\\textit{Fouling present}
    };
    

    \path [line,line width=.1cm] (image) -- (image_prototypes);
    \path [line,line width=.1cm] (image_prototypes) -- (class_prototypes);
    \path [line,line width=.1cm] (class_prototypes) -- (class);

  \end{tikzpicture}}
        \caption{\textbf{Illustration of ComFe.} The image is first clustered into component features, which represent different regions with similar content. The green component feature shows the vessel hull, blue captures the sea chest grating, teal highlights the image tags, and the yellow and red component features identify the regions of the image with biofouling. These are then compared to class prototypes, which match as expected as visualised by the exemplars---training images in the dataset with similar embeddings to the fitted class prototypes. This comparison between component features and class prototypes is used to identify the salient parts of the image for predicting if biofouling is present, as shown by the class confidence heatmap in the final image, where the red regions show areas in the image with higher confidence of fouling being present.}
        \label{fig:explanation_example}
\end{figure*}

\section{Methods}
\label{sec:methods}

\subsection{The ComFe approach}
\label{sec:comfe_approach}

The ComFe\cite{mannix2024scalable} method explains a prediction through the use of component features and class prototypes, as outlined in \cref{fig:explanation_example}. The core components are a pretrained foundation model that is frozen during training, and the ComFe head---a transformer-decoder network that identifies component features, and the class prototypes, a set of learnable vectors within the embedding space of the foundation model that are used to represent the classes within the training data. These parts of ComFe are used in different ways:
\begin{itemize}
    \item The clustering head identifies the component features within an image---regions which share similar characteristics. In \cref{fig:explanation_example} the green component feature captures the flat vessel hull, the blue identifies the sea chest grating, the teal denotes the image tags used in the survey for this particular vessel, while the red and yellow component features capture the biofouling organisms present in the image. 
    \item The class prototypes are compared to the embeddings of the component features, and are associated to each class of interest (e.g. fouling or not fouling), so that a prediction can be made as to whether fouling is detected. As shown in the class prototype panel in \cref{fig:explanation_example}, the component features are grouped as expected, and in the final prediction panel the red regions show areas with higher confidence of fouling being present.
    \item The class prototypes are visualised using the component features within the training data that have similar embeddings, which are called the class prototype exemplars. The class prototypes are visualised in \cref{fig:explanation_example} using five exemplars.
\end{itemize}
Like other interpretable computer vision approaches, ComFe can identify the important patches within an image for making a prediction, as well as identify the images from the training set that support this prediction. However, with a DINOv2 foundation model backbone,\cite{oquab2023dinov2} ComFe obtains better performance, does not require hyper-parameter optimisation to obtain good results on a diverse range of datasets, and improves upon the robustness and generalisability of the backbone model in comparison to the standard black-box approach of fitting a linear head.\cite{mannix2024scalable}

In this paper, we train ComFe models with DINOv2 backbones to identify the presence of fouling, and present results for detecting whether a vessel hull or structure is present within the image in the supplementary materials. For a more detailed explanation of the ComFe approach, see the paper introducing the model.\cite{mannix2024scalable} Code for training ComFe models is available on \href{github.com/emannix/comfe-component-features}{GitHub}.\cite{mannix2024scalable}

\subsection{Datasets}
\label{sec:datasets}

\begin{table}[!tb]
\caption{The Simplified Level of Fouling (SLoF) scale,\cite{mannix2021automating} with a description of correspondence to the more detailed Level of Fouling Scale.\cite{floerl2005risk}}
\label{tbl:slof}
\centering
\begin{tabular}{l|p{13cm}}
\toprule
Rank & Description \\
\midrule
0 & No fouling organisms, but biofilm or slime may be present. (LoF \textbf{0-1})  \\
1 & Fouling organisms (e.g. barnacles, mussels, seaweed or tubeworms) are visible but patchy (1-15\% of surface covered). (LoF \textbf{2-3})  \\
2 & A large number of fouling organisms are present (16-100\% of surface covered). (LoF \textbf{4-5})  \\
\bottomrule
\end{tabular}
\end{table}

\begin{figure*}[!tb]
\centering
\includegraphics[width=1\textwidth]{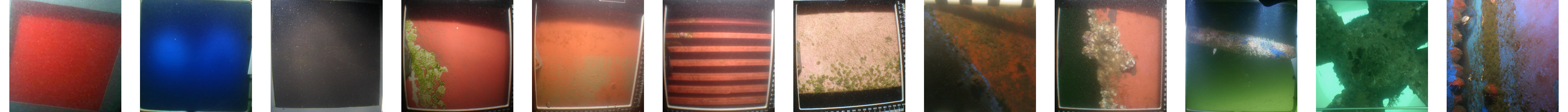}
\includegraphics[width=1\textwidth]{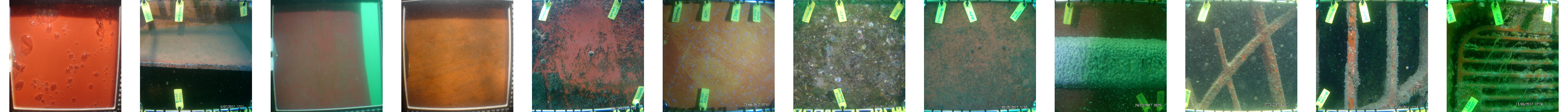}
\caption[Example images.]{Example training (first row) and testing (second row) images from the biofouling dataset.\cite{mannix2021automating} }
\label{fig:dataset_2021_cropped_examples}
\end{figure*}

\begin{table*}[!tb]
\caption[Brief summary of SLoF counts in the biofouling dataset.]{Brief summary of SLoF counts in the biofouling dataset.\cite{mannix2021automating}}
\label{tbl:dataset_2021_cropped}
\centering
\begin{tabular}[t]{lrrrr}
\toprule
\textbf{Category} & SLoF 0 & SLoF 1 & SLoF 2 & Total \\
\midrule
\textbf{Training}  & 7328 & 1503 & 591 & 9422  \\ 
\textbf{Test}  & 494 &  193 & 154 & 841 \\ 
\bottomrule
\end{tabular}
\end{table*}

The ComFe models are trained on a dataset of 10,263 images from underwater surveys of around 300 commercial and recreational vessel hulls, as introduced in prior work.\cite{mannix2021automating} The dataset comprises images provided from a number of different jurisdictions, including Australia, New Zealand and California. These images were graded according to the Simplified Level of Fouling (SLoF) scale, as described in \cref{tbl:slof}, by experts from Ramboll New Zealand, who held qualifications in marine biology and had extensive experience working with biofouling imagery. 

Example images are shown in \cref{fig:dataset_2021_cropped_examples}, highlighting the variability in the imagery in terms of lighting, antifoulant coating colour, quality and fouling organisms. Overall, the SLoF annotations (\cref{tbl:dataset_2021_cropped}) were highly imbalanced, with most of the images having no fouling (SLoF 0), compared to ~20\% with a small degree of fouling (SLoF 1) and ~10\% having a large degree of fouling (SLoF 2). To account for this in evaluating the performance of the models, the average precision metric was used to measure model performance. Average precision has been shown to provide a better indication of model
performance in this context in comparison to other choices, such as accuracy or the Area Under the Receiver Operator Curve Curve (AUROC).\cite{liu2019binormal}

Following previous studies,\cite{mannix2021automating} the dataset was divided into the same training and test splits to allow comparisons to be made with prior works. The majority of the test split was selected from all of the images belonging to 14 particular vessels, that were chosen to test the generalisation of models to imagery from vessels that had not been previously seen. These vessels were selected on the basis capturing the diversity present within the data in terms of biofouling burden, niche area styles, and types of biofouling communities. 

Further video footage was also collected using an underwater ROV, from a marina environment with high turbidity and poor visibility. This dataset is used within this paper to qualitatively test the utility of the trained models in an operational setting. Both datasets are available for download from \href{https://figshare.com/articles/dataset/Automating_the_assessment_of_biofouling_in_images/26537158}{figshare}.\cite{Mannix2024figshare}

\subsection{Implementation details}
\label{sec:implementation}

The ComFe models were fitted using DINOv2 backbones\cite{oquab2023dinov2} closely following the original method.\cite{mannix2024scalable} We found that performance was sensitive to the initialisation of the model, but that the models could be fit quickly and required fewer epochs---training took less than an hour on an NVIDIA A100 GPU for all cases considered. We trained ten ComFe models with different initialisation seeds for ten epochs (with two warmup epochs), and the run with the best average precision was selected to be the final model. For the ViT-S/14 DINOv2 variant, we used the original DINOv2 model, but for the ViT-B/14 and ViT-L/14 variants we employed the architectures with registers, as this improved performance.\cite{darcet2023vision, mannix2024scalable}

For identifying biofouling we considered the background (non-informative) class to be a no-fouling class, and assumed every image would have atleast one patch not containing the variable of interest (e.g. a patch not containing macrofouling). These patches could contain timestamps, tags, survey frames, or clean areas of hull or open water. This meant that rather than using the ordinal SLoF labels for training as done in previous works,\cite{mannix2021automating} in this paper we focused on identifying the presence of fouling (SLoF 1 and 2) versus absence (SLoF 0). We explore estimating severity using the coverage of biofouling over the whole image identified by the ComFe model within the results section.

The hyperparameters used for fitting the ComFe models were mostly unchanged from the original ComFe paper,\cite{mannix2024scalable} including the image augmentation and preprocessing steps applied. However, due to the smaller number of classes considered in comparison to most standard computer vision benchmarks, the number of class prototypes was increased to ten per class. 

For comparison to a non-interpretable alternative, we also train linear heads using the standard approach to fitting them with a frozen DINOv2 backbone.\cite{oquab2023dinov2} Modifications were made to this fitting process to account for dataset imbalance as discussed in previous work,\cite{mannix2021automating} such as selecting the approach with the best average precision and using a class-weighted crossentropy loss function.

\subsection{Application to video footage}
\label{sec:implementation_video}

A stream of still images can be extracted from video footage, allowing a model only trained to assess single frames to be used to identify key features in a video. In this work, the challenge of identifying the SLoF level present in an image is defined on the basis of a single frame, so it is natural for videos to also be assessed in this way. However, this introduces additional complexities, as videos can be made up of a significant number of frames, and neural networks have been found to be sensitive to the perturbations between neighbouring frames within a video stream and predictions can be noisy.\cite{shankar2021image} To address this, the models are evaluated on only every $n$\textit{th} frame to reduce computational costs, and Gaussian kernel smoothing\cite{hardle1992kernel} is used to average the predicted confidence and coverage of biofouling over neighbouring frames. This latter approach has the effect of stabilising predictions over time, and an example of smoothed versus unsmoothed outputs is shown in the supporting information. 

To support quick decision making from the model outputs, we further employ the Simple $k$-Means Prototype Selection (SKMPS) strategy\cite{mannix2024efficient} to identify a set of representative frames that can summarise the content within a video. This approach identifies clusters of visually similar frames using the normalised global embeddings of the backbone model, and selects the image closest to the centre of each cluster as the representative example. This allows for the diversity of images with and without fouling to be quickly visualised and examined, without the footage needing to be manually assessed.

Footage may also contain a large amount of imagery that is not relevant to identifying the presence of biofouling, such as video of open water and on-dock activities. To filter out these non-relevant sections, a further hull detection model is trained using ComFe as described in the supporting information. 
A version of these models suitable for deployment is available for download from \href{https://figshare.com/articles/dataset/Automating_the_assessment_of_biofouling_in_images/26537158}{figshare}, as highlighted on \href{https://github.com/emannix/automating-the-assessment-of-biofouling}{GitHub}.\cite{Mannix2024figshare}

\section{Results}
\label{sec:results}

\subsection{ComFe models are interpretable}

\begin{figure*}[!tb]
     \centering
     \begin{subfigure}[b]{0.12\textwidth}
         \centering
        \begin{overpic}[height=\textwidth, width=\textwidth]{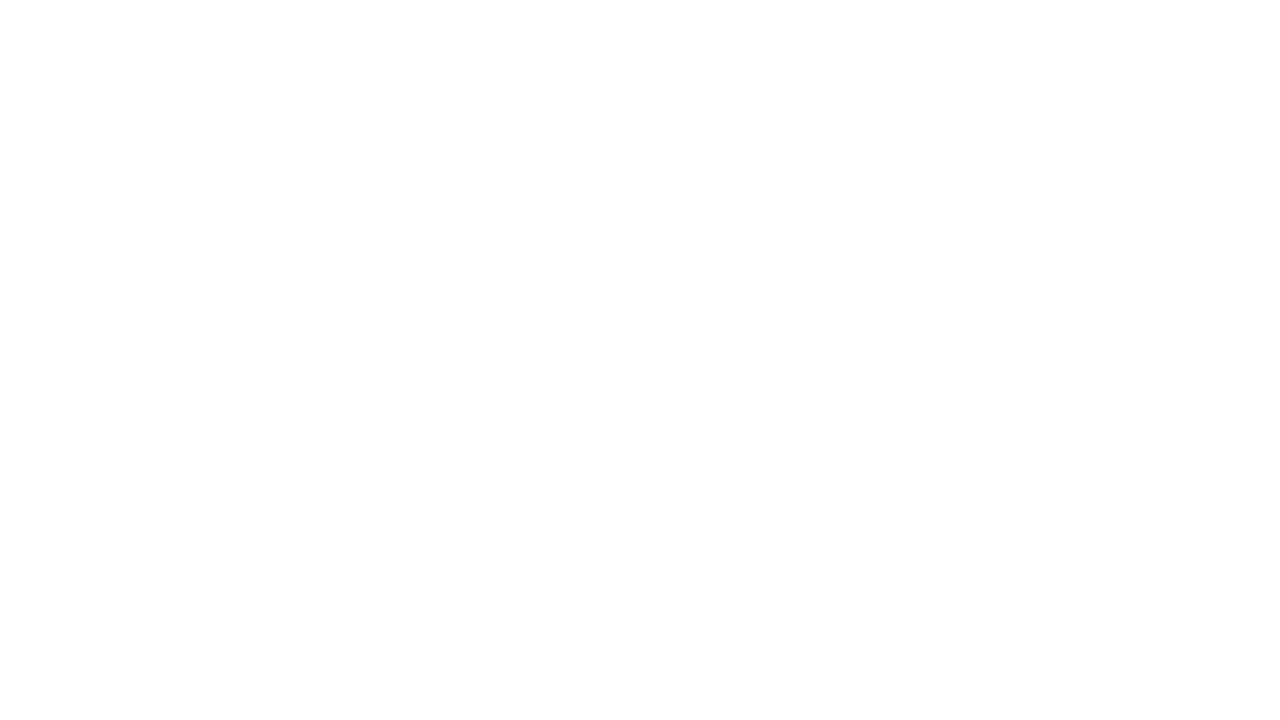}
         \put (0,50) {\parbox{2cm}{\centering Input image}}
        \end{overpic}
        \begin{overpic}[height=\textwidth, width=\textwidth]{example_images/1280px-HD_transparent_picture.png}
         \put (0,50) {\parbox{2cm}{\centering Class\\prediction}}
        \end{overpic}
     \end{subfigure}
     \begin{subfigure}[b]{0.12\textwidth}
         \centering
         \caption{Clean}
         \includegraphics[width=\textwidth]{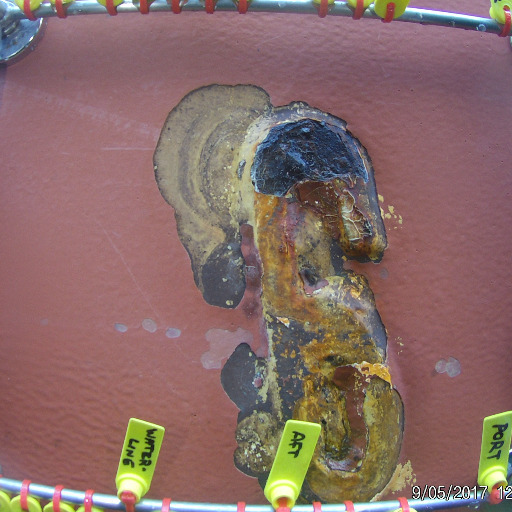}
         \includegraphics[width=\textwidth]{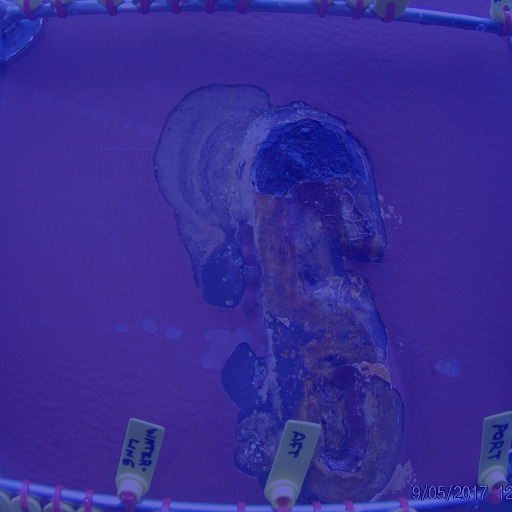}
     \end{subfigure}
     \begin{subfigure}[b]{0.12\textwidth}
         \centering
         \caption{Clean}
         \includegraphics[width=\textwidth]{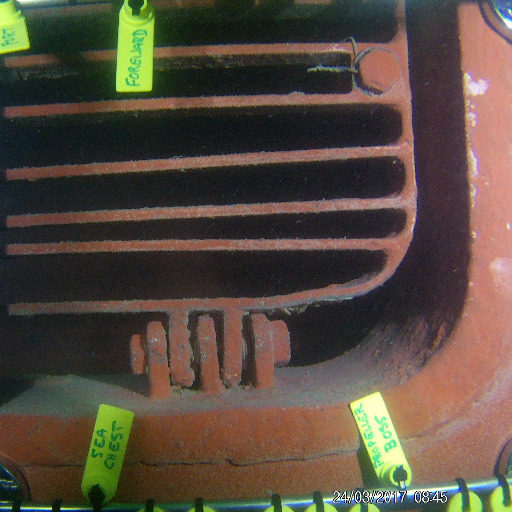}
         \includegraphics[width=\textwidth]{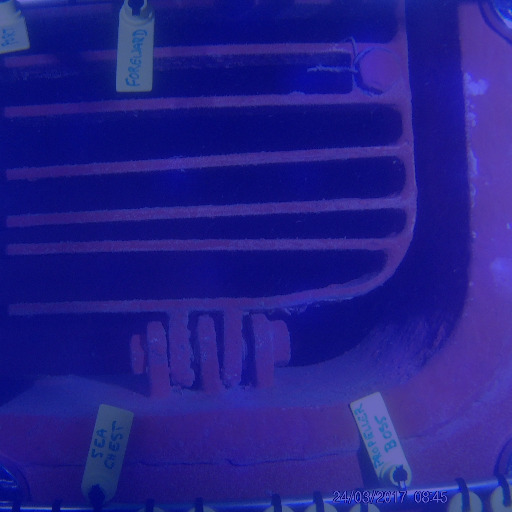}
     \end{subfigure}
     \begin{subfigure}[b]{0.12\textwidth}
         \centering
         \caption{Clean}
         \includegraphics[width=\textwidth]{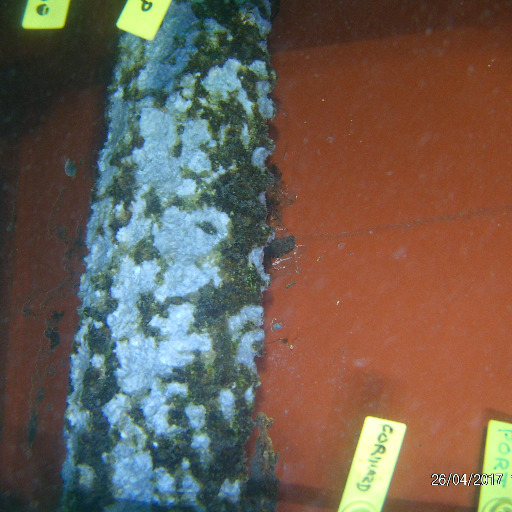}
         \includegraphics[width=\textwidth]{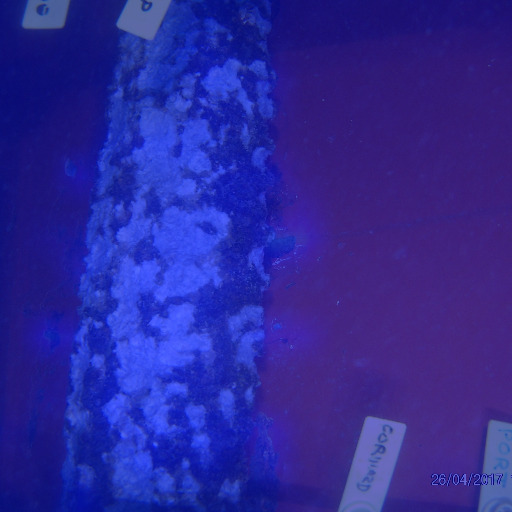}
     \end{subfigure}
     \begin{subfigure}[b]{0.12\textwidth}
         \centering
         \caption{Fouled}
         \includegraphics[width=\textwidth]{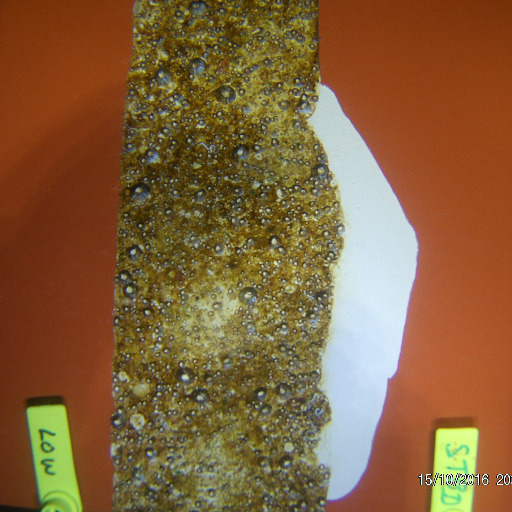}
         \includegraphics[width=\textwidth]{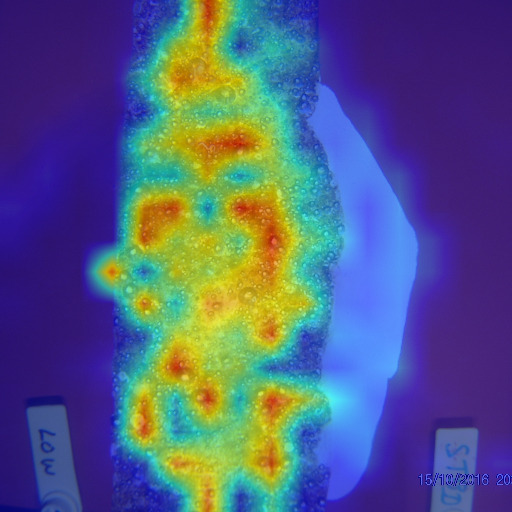}
     \end{subfigure}
     \begin{subfigure}[b]{0.12\textwidth}
         \centering
         \caption{Fouled}
         \includegraphics[width=\textwidth]{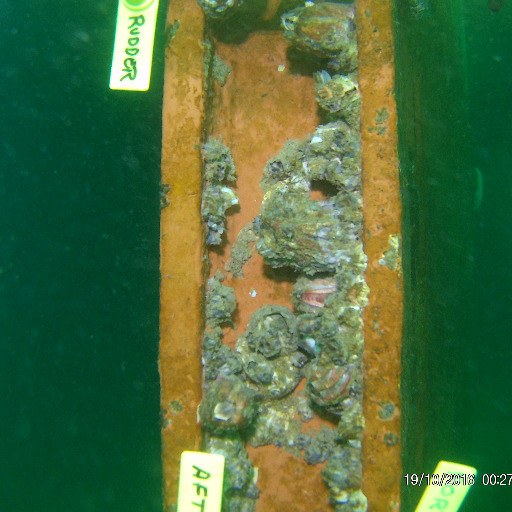}
         \includegraphics[width=\textwidth]{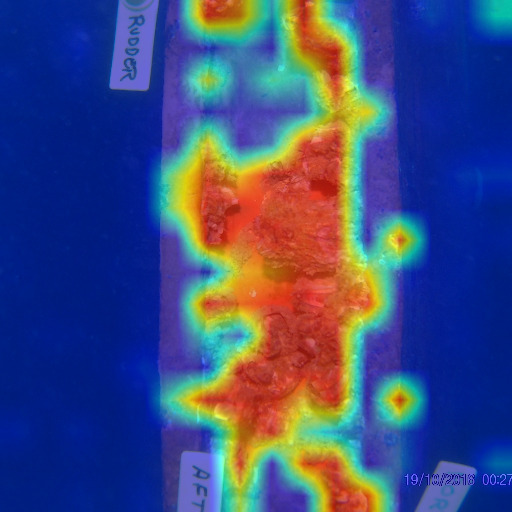}
     \end{subfigure}
     \begin{subfigure}[b]{0.12\textwidth}
         \centering
         \caption{Fouled}
         \includegraphics[width=\textwidth]{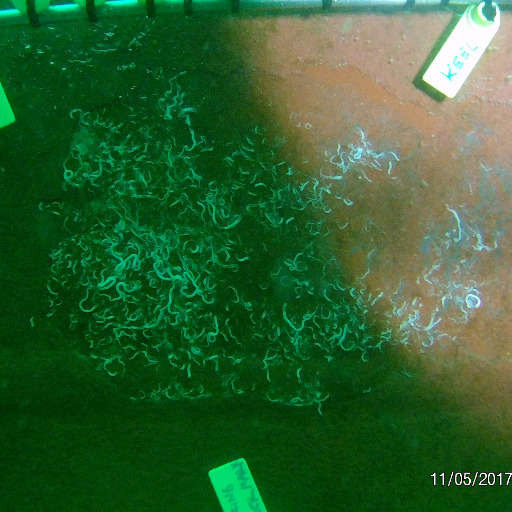}
         \includegraphics[width=\textwidth]{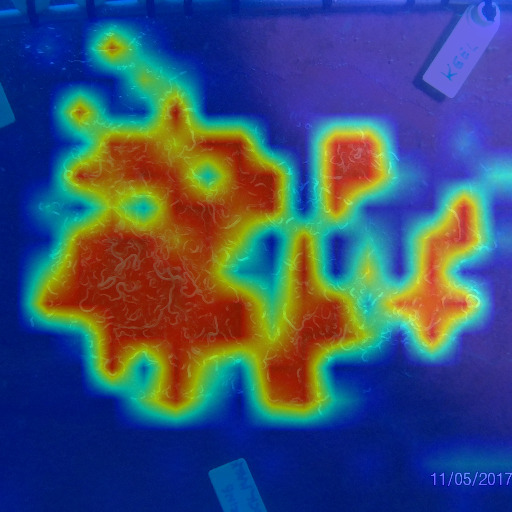}
     \end{subfigure}
        \caption{Example detections of biofouling versus not biofouling within an image using the DINOv2 ViT-B/14 w/reg network with a ComFe head. The red regions show areas with a high confidence of fouling being present, and the blue regions highlight areas of zero confidence. }
        \label{fig:salient_features_fouling}
\end{figure*}

In \cref{fig:explanation_example} an example ComFe explanation is shown, which identifies the image regions which result in a \textit{fouling present} classification being made, and highlights exemplars from the training dataset that support this conclusion. The full set of class prototypes and their exemplars is shown in the supporting information. 
Further examples are shown in \cref{fig:salient_features_fouling}, demonstrating the ability of the ComFe model to identify where in an image biofouling occurs when making a prediction, without being explicitly trained with segmentation annotations.


\subsection{ComFe models outperform prior work}

\begin{table*}[!tb]
\caption[Performance of ComFe for identifying the presence of macrofouling compared to previous methods.]{Performance of ComFe for identifying the presence of macrofouling compared to previous methods. We report the mean (maximum) test average precision from ten runs, using the best validation epoch, to compare to previous results. }
\label{tbl:accuracy_comparison_previous}
\centering
\begin{tabular}[t]{lrlrl}
\toprule
\multicolumn{2}{c}{\textbf{Head}} & \multicolumn{2}{c}{\textbf{Backbone}} & \multicolumn{1}{c}{\textbf{AP}} \\
\midrule
Linear \cite{mannix2021automating}  &  &  CNN Ensemble  & 5$\times$182M & 0.921  \\ 
\midrule
Linear  &  &  DINOv2 ViT-S/14 (f)       & 21M & 0.902 (0.905) \\ 
 &  &  DINOv2 ViT-B/14 w/reg (f) & 86M & 0.904 (0.912)  \\ 
 &  &  DINOv2 ViT-L/14 w/reg  (f) & 300M & 0.918 (0.921)  \\ 
\midrule
ComFe  & 8M & DINOv2 ViT-S/14 (f) & 21M & 0.912 (0.920)  \\ 
 & 32M & DINOv2 ViT-B/14 w/reg (f) & 86M & 0.923 (0.931)  \\ 
 & 57M & DINOv2 ViT-L/14 w/reg (f) & 300M & \textbf{0.927} (0.930)  \\ 
\bottomrule
\end{tabular}
\end{table*}

The performance of the CNN ensemble from prior work\cite{mannix2021automating} is shown in \cref{tbl:accuracy_comparison_previous}, and compared to the performance of various DINOv2 backbones trained to identify the presence of fouling using a non-interpretable linear head,\cite{oquab2023dinov2} and the interpretable ComFe approach.\cite{mannix2024scalable} To ensure the comparability of these results, we follow the same approach \cite{mannix2021automating} and employ 5-fold cross validation, and only use the test data for evaluation. The CNN ensemble\cite{mannix2021automating} uses the combined prediction of all five cross validation fits with a number of different CNN backbones, and while this approach obtains performance on par with experts,\cite{mannix2021automating} it has a large number of parameters (910 million in total) and would be inefficient to deploy in a production environment. In comparison, a ComFe model with a DINOv2 backbone can obtain better performance with a single model, with an order of magnitude fewer parameters. Further, the ComFe models also obtain better performance than the standard approach using a linear head,\cite{oquab2023dinov2} which is consistent with previous results on a number of computer vision benchmarks.\cite{mannix2024scalable}


\subsection{Predicted coverage is correlated with SLoF category}

\begin{figure*}[!tb]
     \centering
     \begin{subfigure}[b]{0.45\textwidth}
         \centering
         \includegraphics[width=\textwidth]{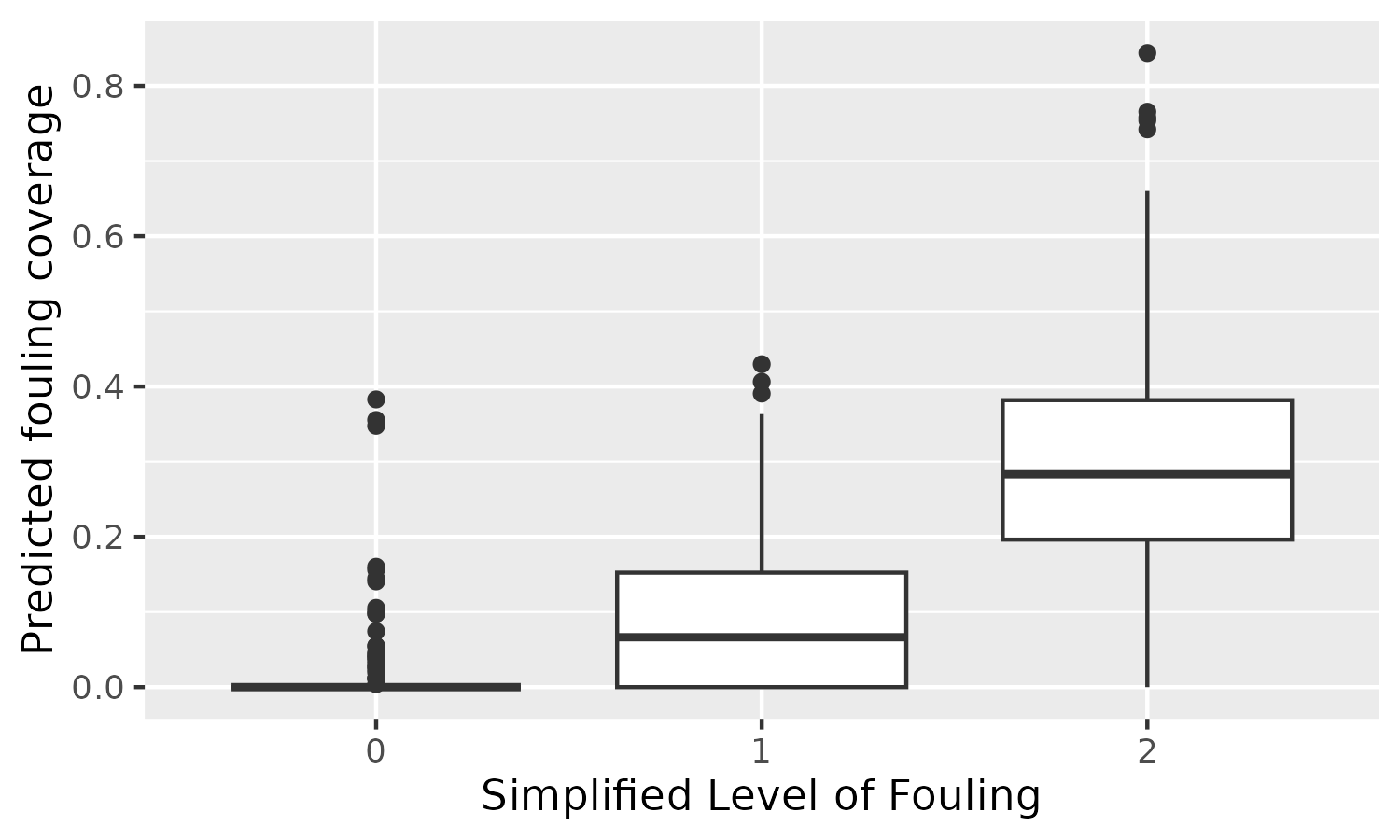}
     \end{subfigure}
        \caption{Predicted coverage within an image of fouling and paint damage by ComFe versus label categories for fouling and paint damage severity. }
        \label{fig:severity_coverage}
\end{figure*}

The ComFe models learn to identify which patches within an image relate to biofouling, or areas which are not fouled. Examples of confidence maps showing where biofouling occurs within an image are shown in \cref{fig:salient_features_fouling}. While the models are only trained with annotations that describe whether or not biofouling is present, these maps can be thresholded and used to describe biofouling coverage. In \cref{fig:severity_coverage} the SLoF class of each image in the validation set is plotted against the predicted fouling coverage by the ComFe model, and it can be seen that the thresholds described in \cref{tbl:slof} are generally captured within this box plot. Images without fouling (SLoF 0) have no predicted coverage, images categorised as SLoF 1 mostly have between 1-15\% coverage of fouling, and heavily fouled images (SLoF 2) generally have greater than 15\% coverage of biofouling. This provides further information to decision makers about not just if fouling is present, but the severity as well.


\subsection{ComFe models can summarise video footage}

\begin{figure*}[!tb]
     \centering
     \begin{subfigure}[b]{0.3\textwidth}
         \centering
         \includegraphics[width=\textwidth]{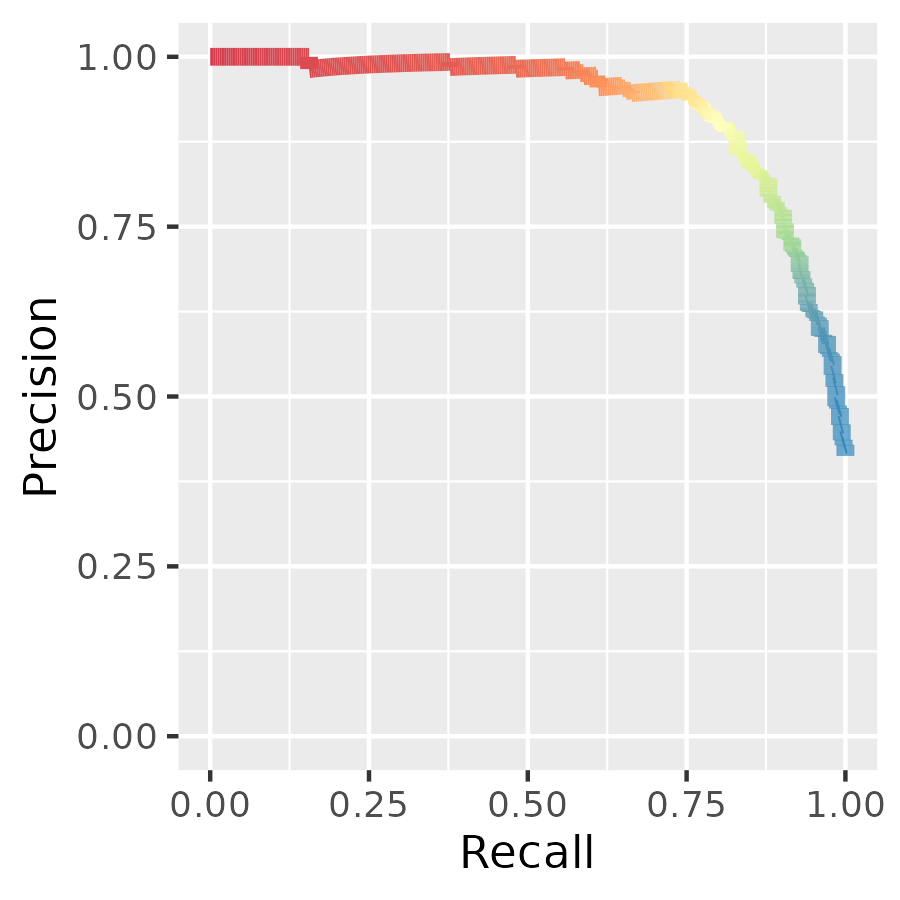}
     \end{subfigure}%
     \begin{subfigure}[b]{0.09\textwidth}
         \picdims[height=4.5cm]{\textwidth}{3.5cm}{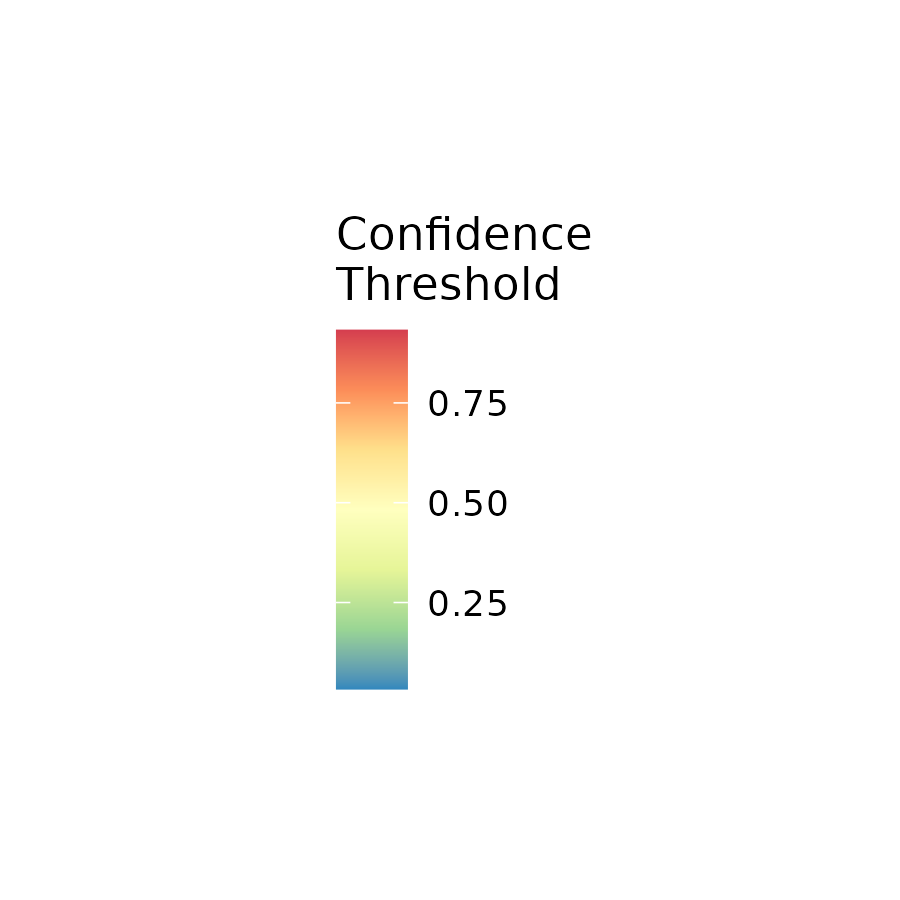}
         \vspace{1cm}
     \end{subfigure} \\

        \caption[Precision-recall curve for the best ComFe head for detecting the presence of biofouling. ]{Precision-recall curve for the best ComFe model using the DINOv2 ViT-B/14 (f) w/reg backbone for detecting the presence of biofouling.}
        \label{fig:model_suite_average_precision_biofouling}
\end{figure*}


\begin{figure*}[!tb]
     \centering
     \begin{subfigure}[b]{0.11\textwidth}
         \centering
        \begin{overpic}[height=0.6\textwidth, width=\textwidth]{example_images/1280px-HD_transparent_picture.png}
         \put (0,20) {\parbox{1.5cm}{\centering \small Fouling}}
        \end{overpic}
        \begin{overpic}[height=0.6\textwidth, width=\textwidth]{example_images/1280px-HD_transparent_picture.png}
         \put (0,20) {\parbox{1.5cm}{\centering \small No\\Fouling}}
        \end{overpic}
     \end{subfigure}%
     \begin{subfigure}[b]{0.11\textwidth}
         \centering
         \includegraphics[width=\textwidth]{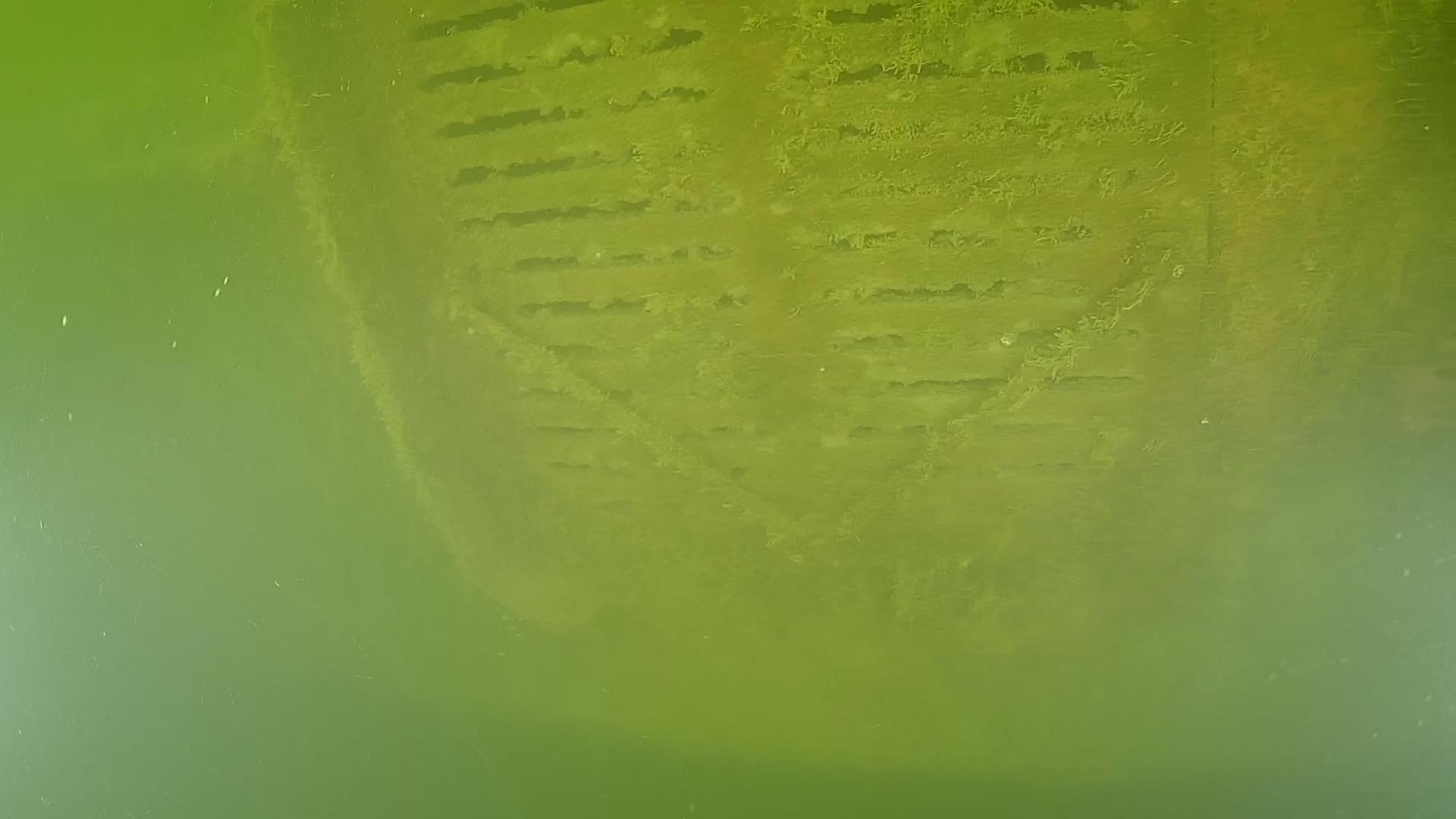}
         \includegraphics[width=\textwidth]{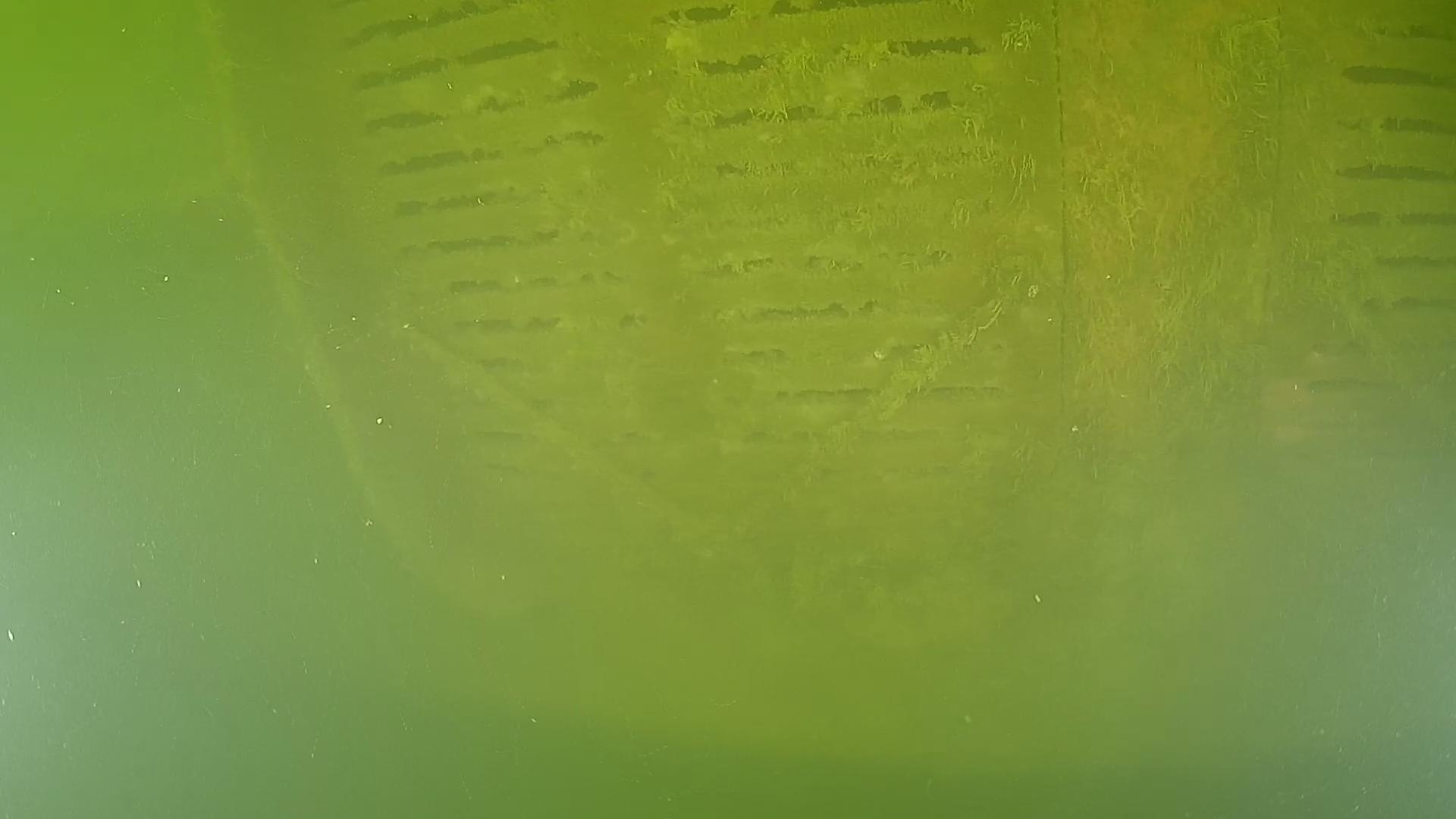}
     \end{subfigure}%
     \begin{subfigure}[b]{0.11\textwidth}
         \centering
         \includegraphics[width=\textwidth]{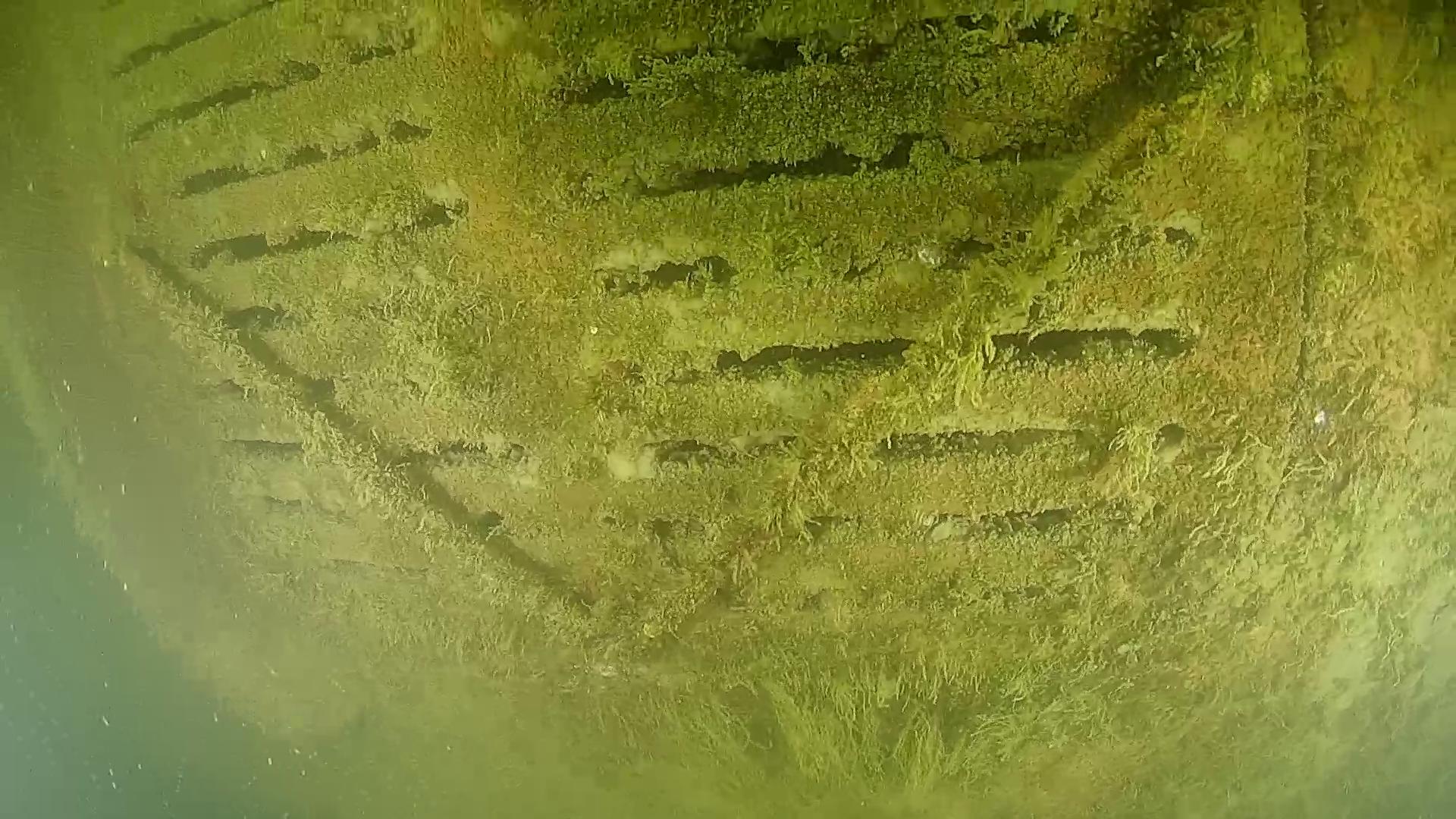}
         \includegraphics[width=\textwidth]{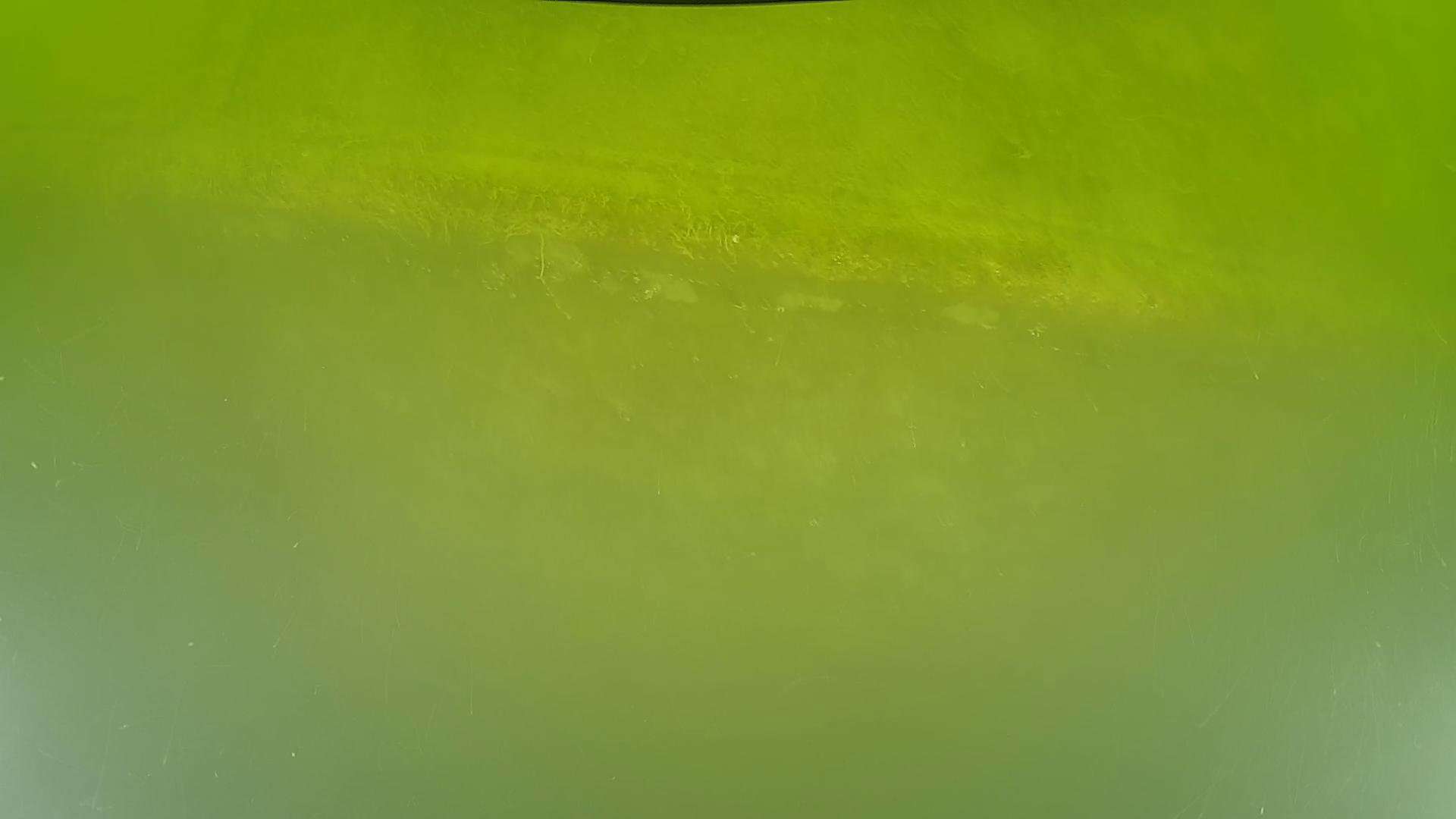}
     \end{subfigure}%
     \begin{subfigure}[b]{0.11\textwidth}
         \centering
         \includegraphics[width=\textwidth]{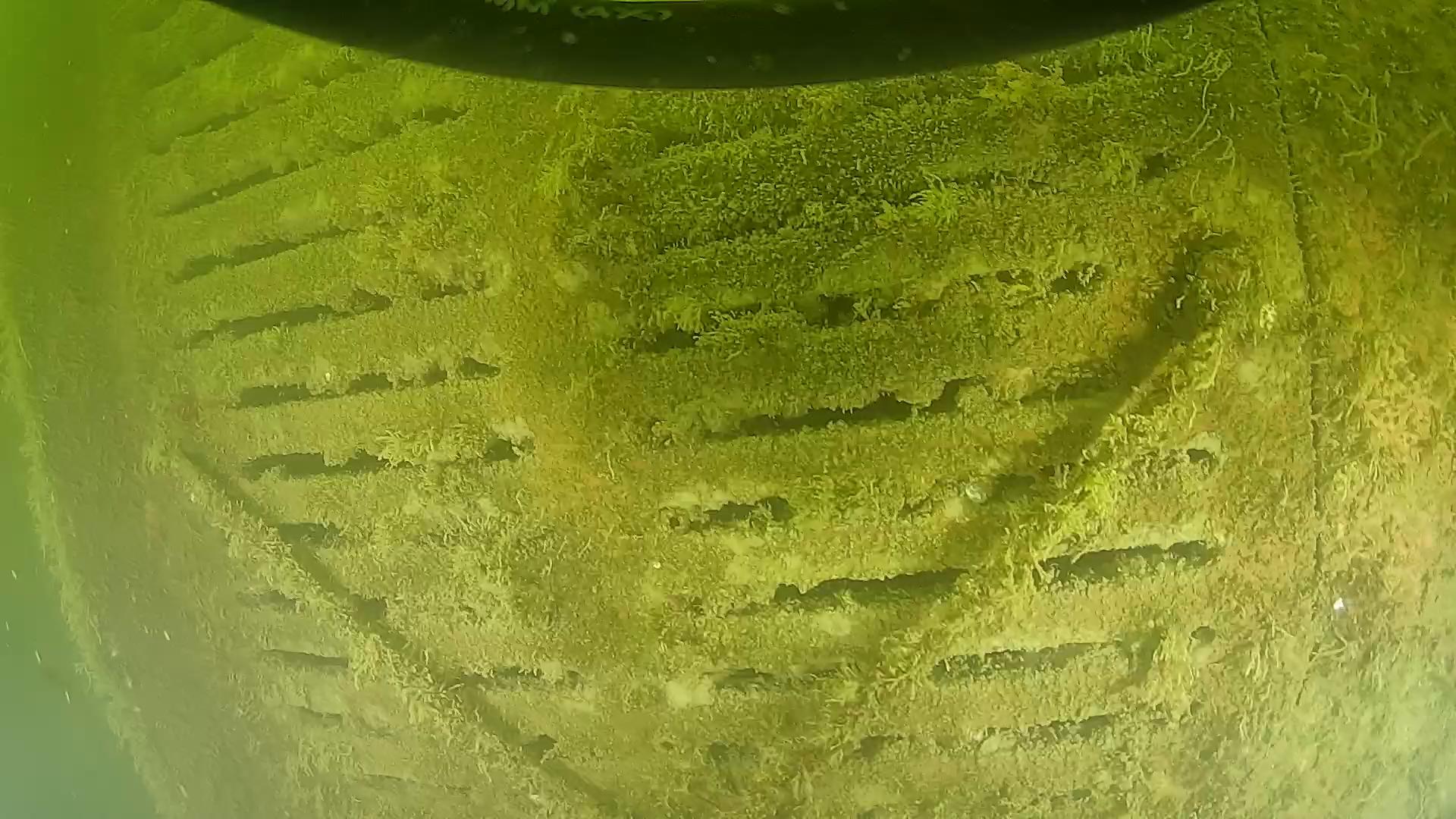}
         \includegraphics[width=\textwidth]{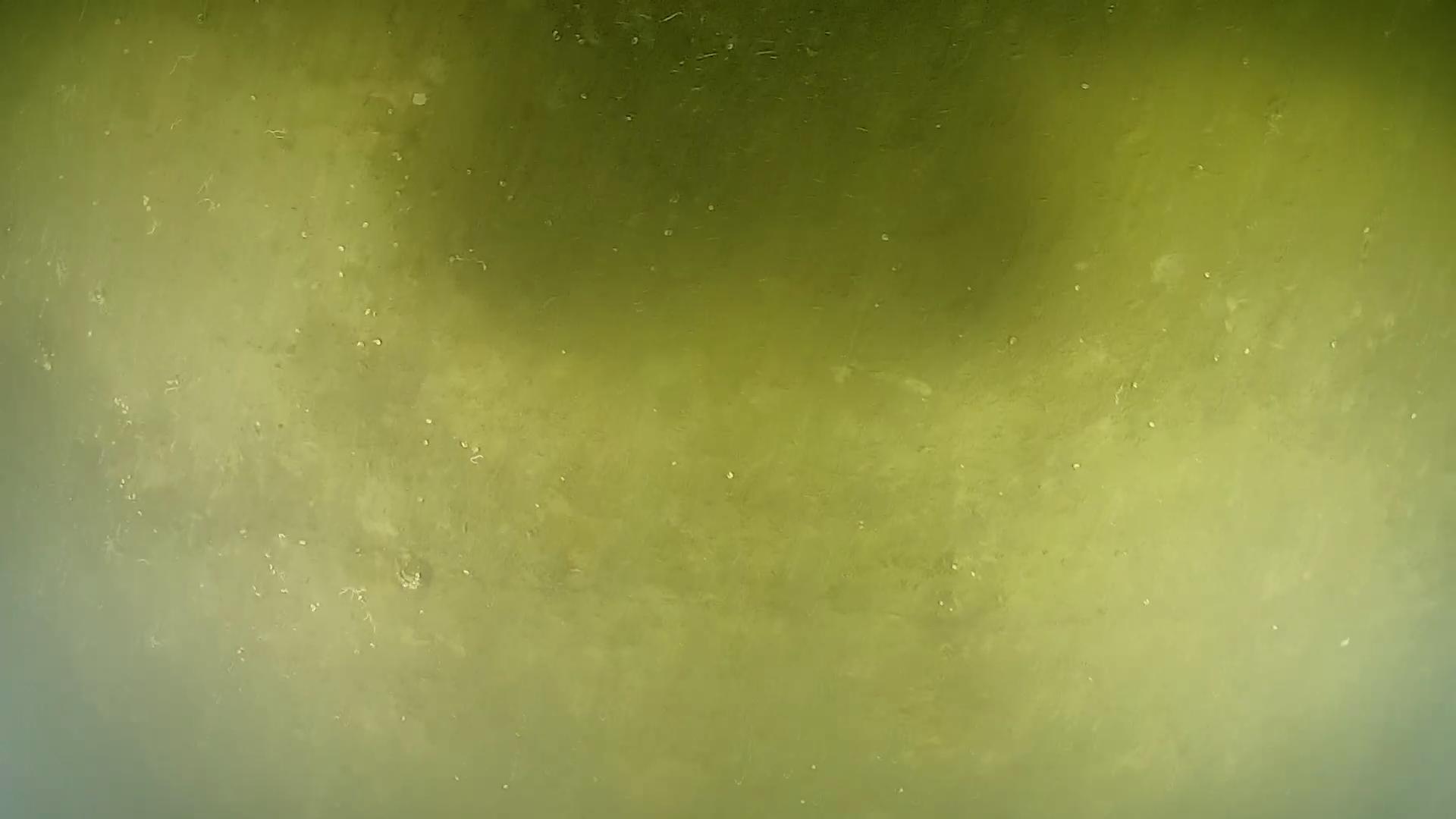}
     \end{subfigure}%
     \begin{subfigure}[b]{0.11\textwidth}
         \centering
         \includegraphics[width=\textwidth]{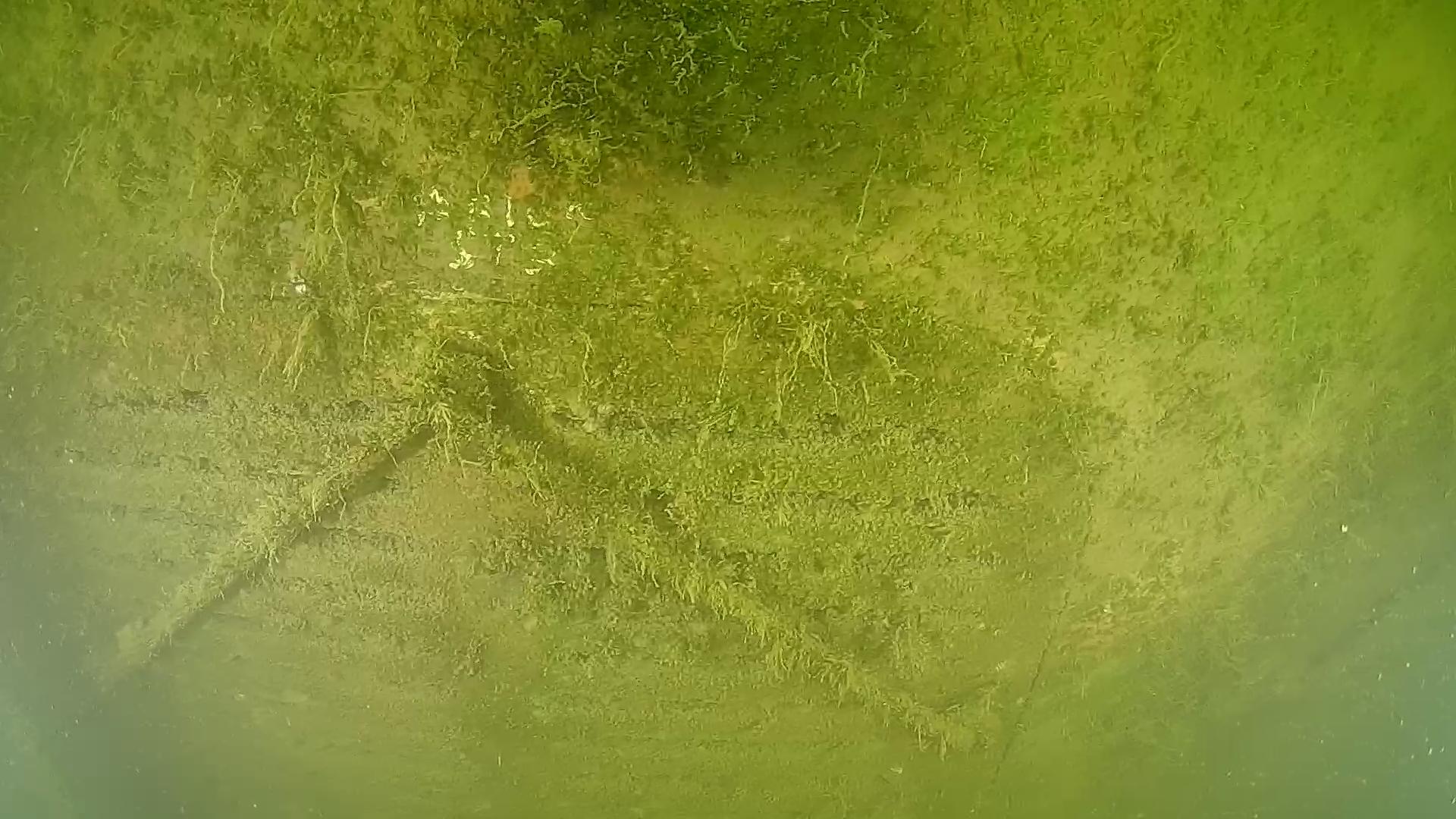}
         \includegraphics[width=\textwidth]{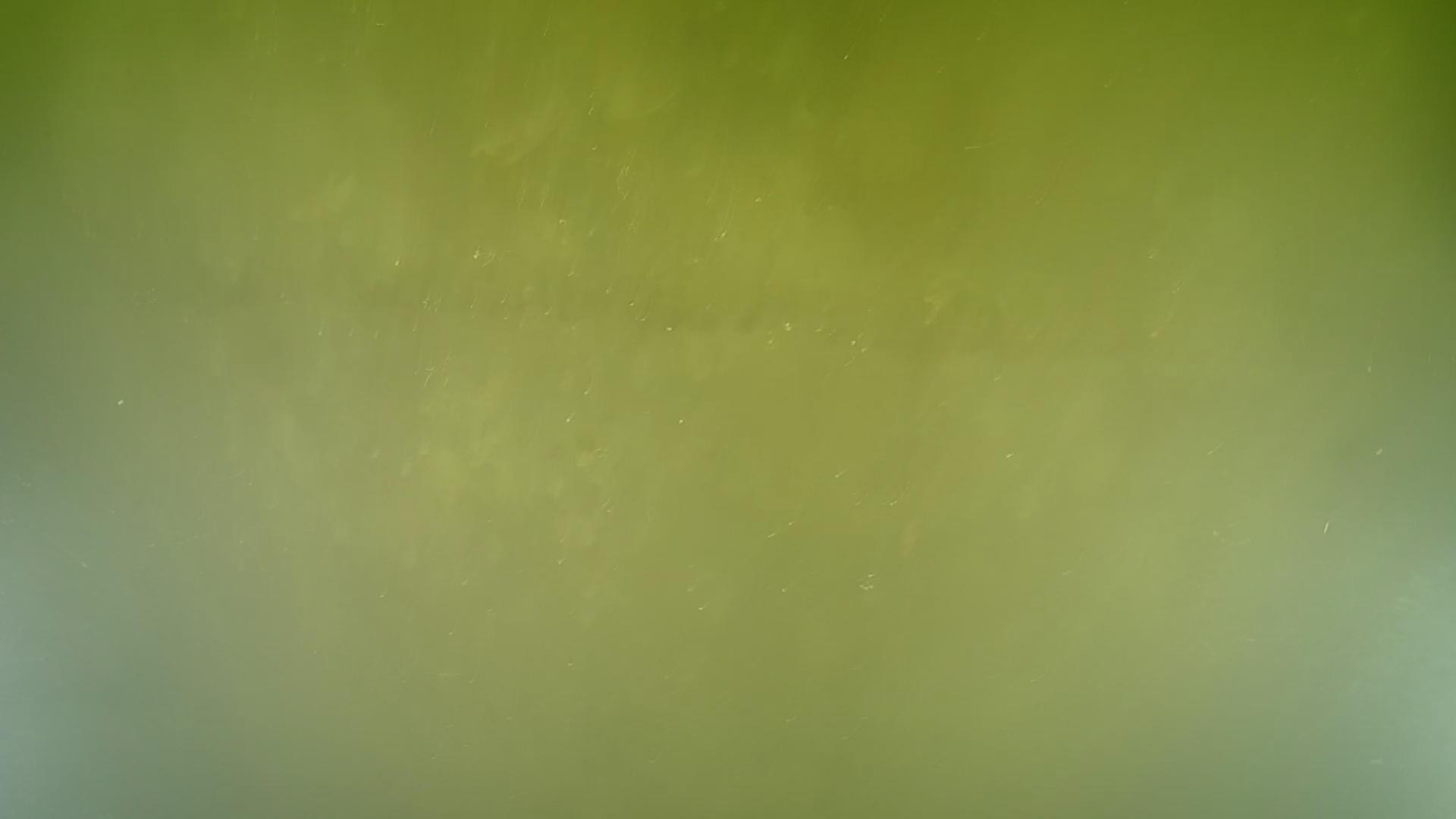}
     \end{subfigure}%
     \begin{subfigure}[b]{0.11\textwidth}
         \centering
         \includegraphics[width=\textwidth]{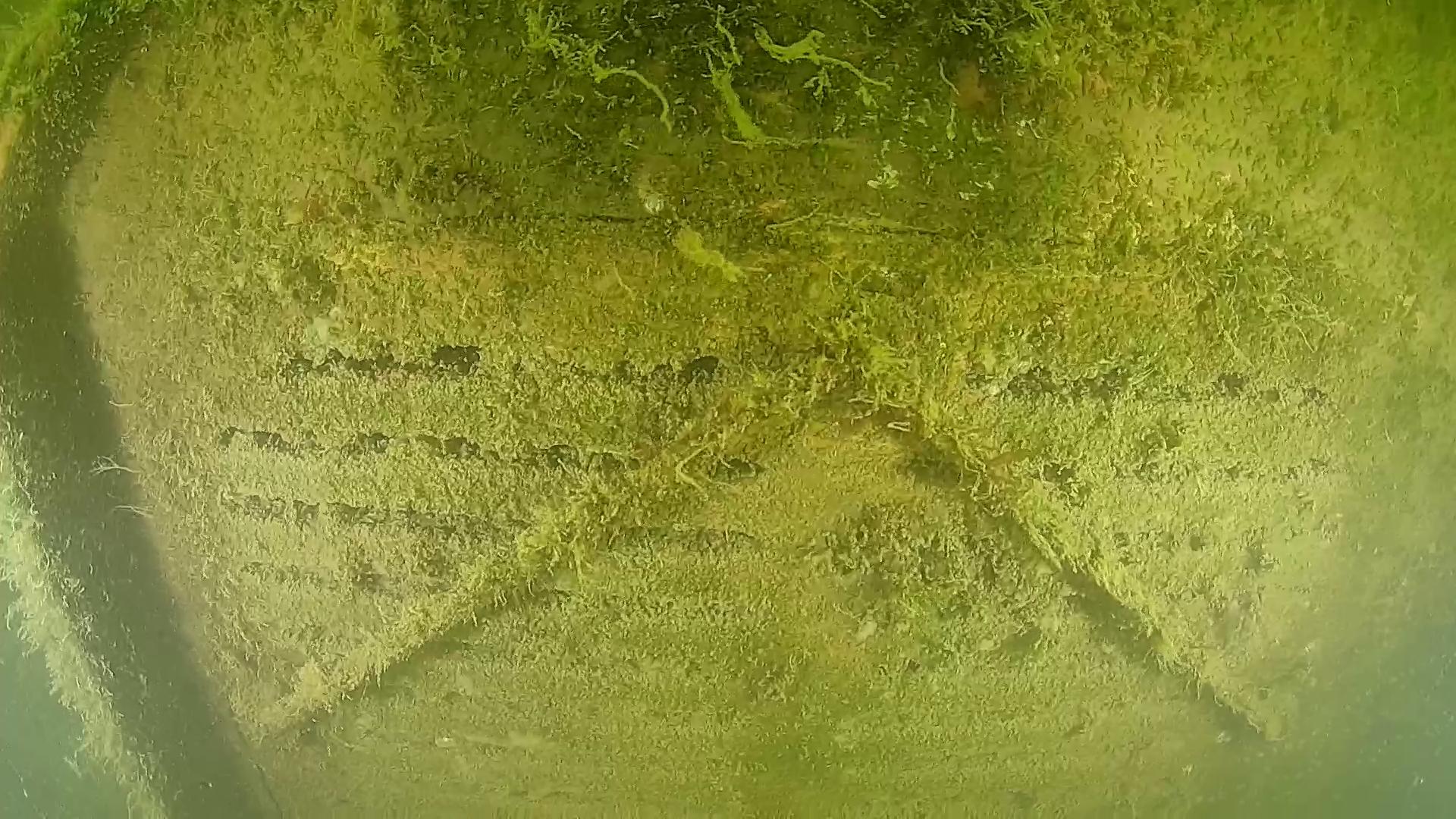}
         \includegraphics[width=\textwidth]{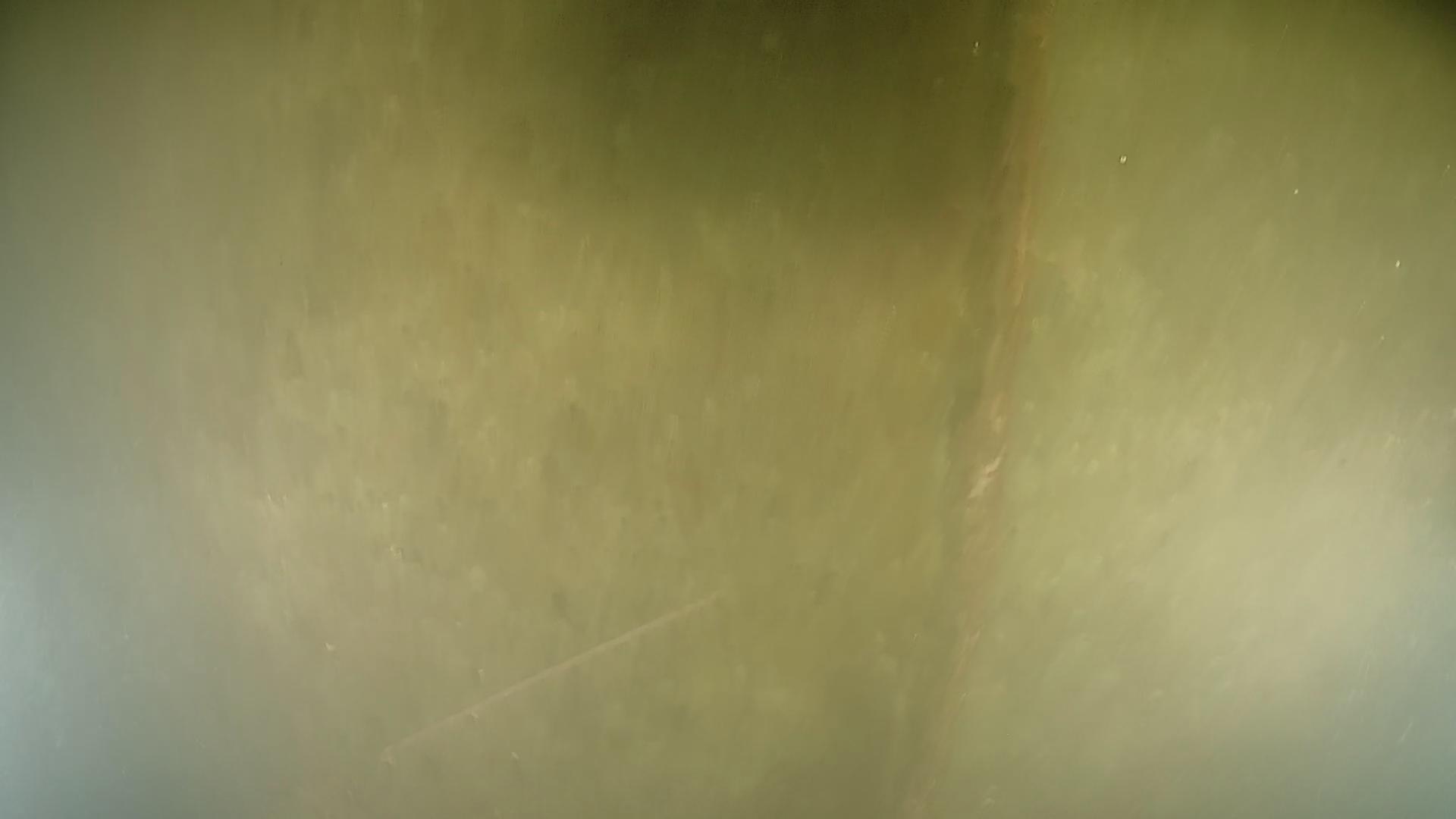}
     \end{subfigure}%
     \begin{subfigure}[b]{0.11\textwidth}
         \centering
         \includegraphics[width=\textwidth]{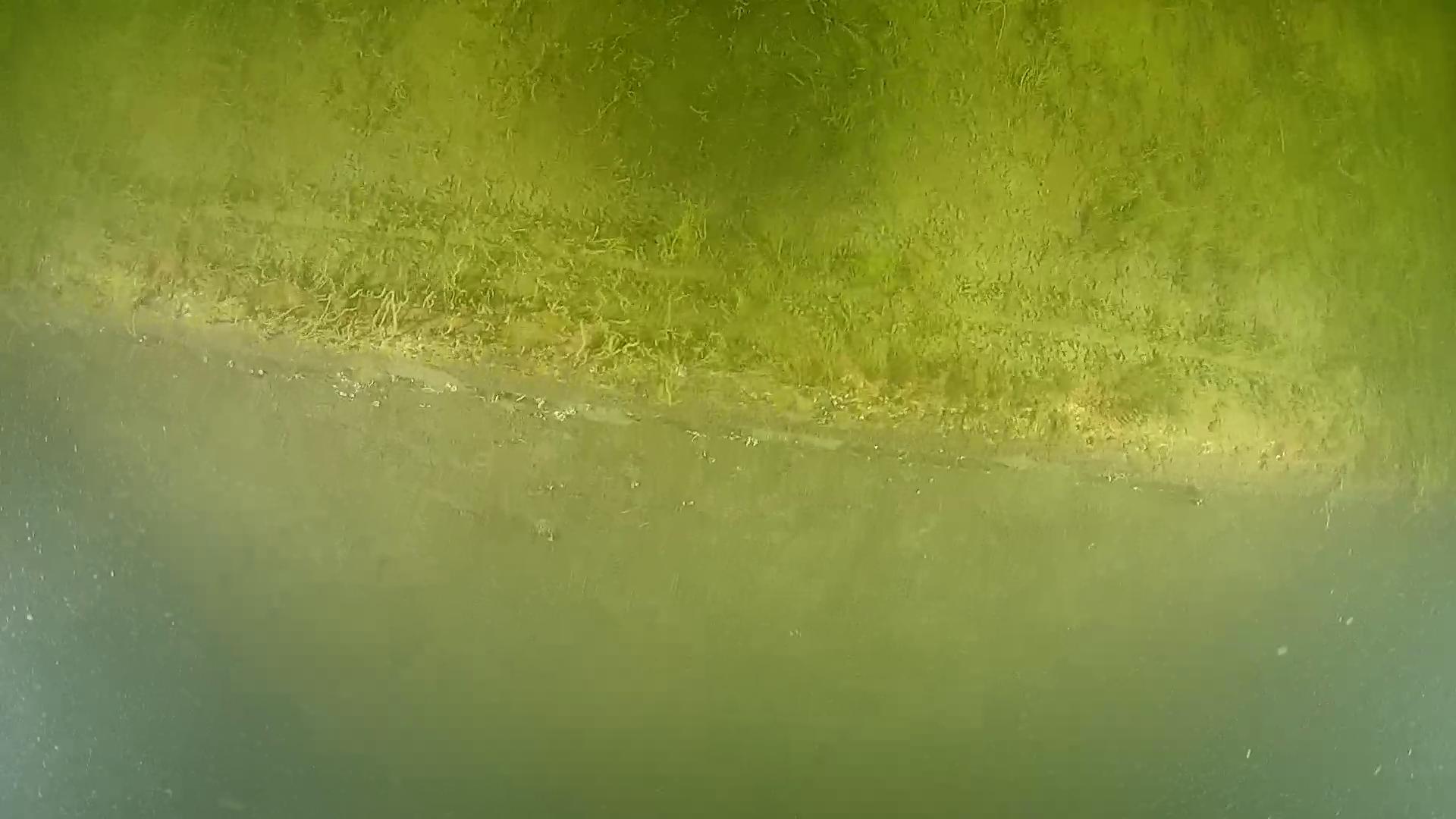}
         \includegraphics[width=\textwidth]{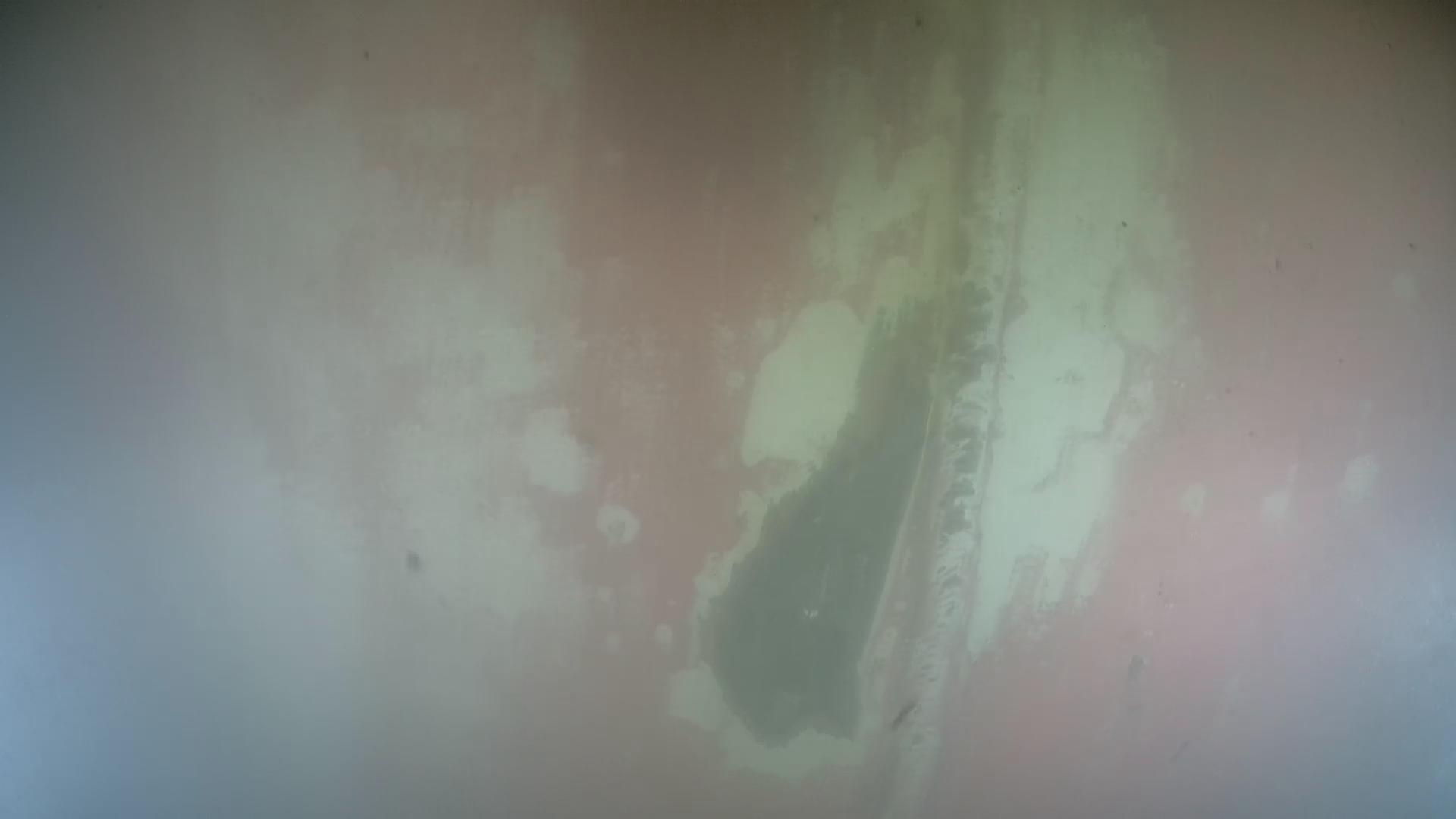}
     \end{subfigure}%
     \begin{subfigure}[b]{0.11\textwidth}
         \centering
         \includegraphics[width=\textwidth]{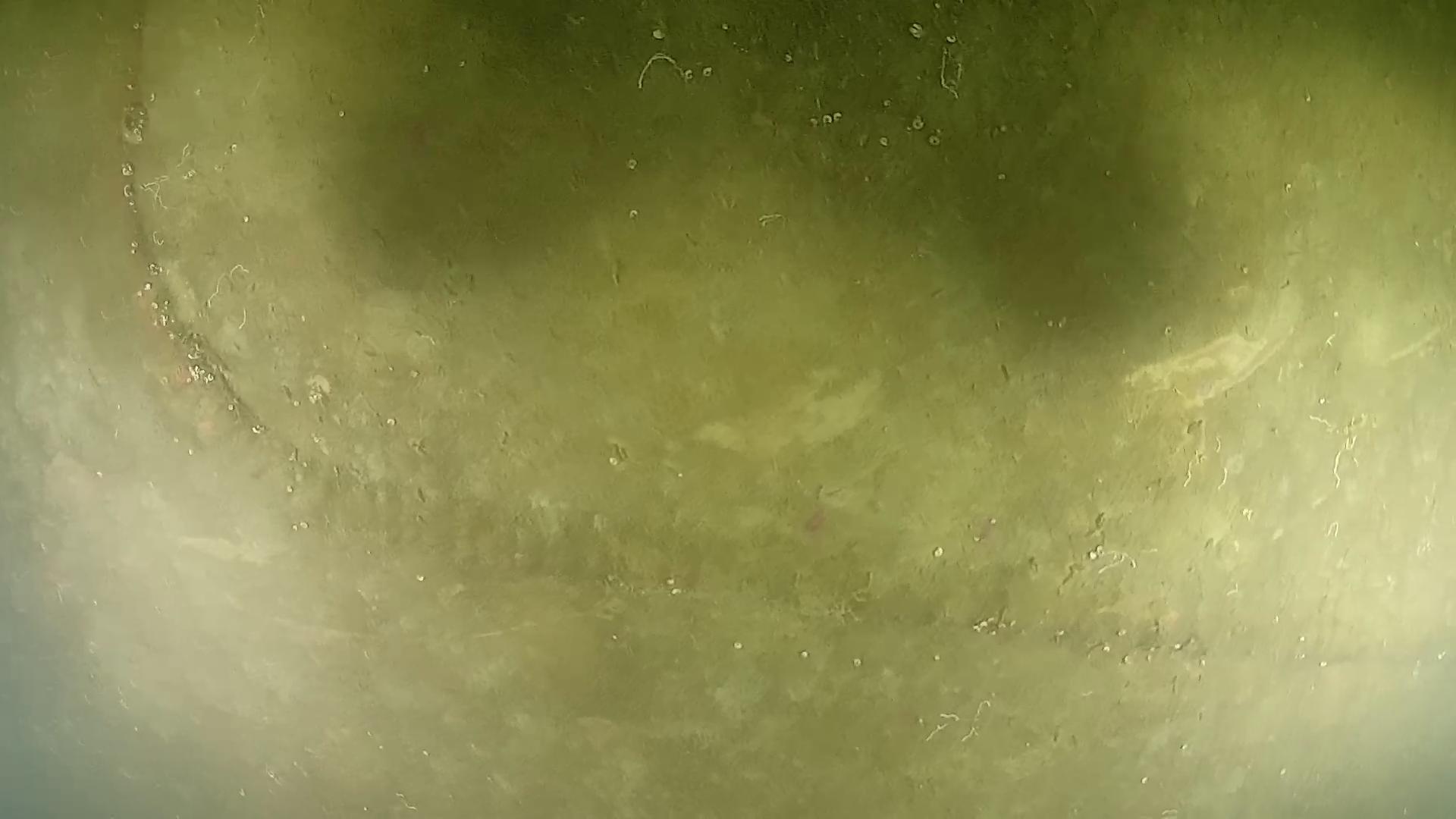}
         \includegraphics[width=\textwidth]{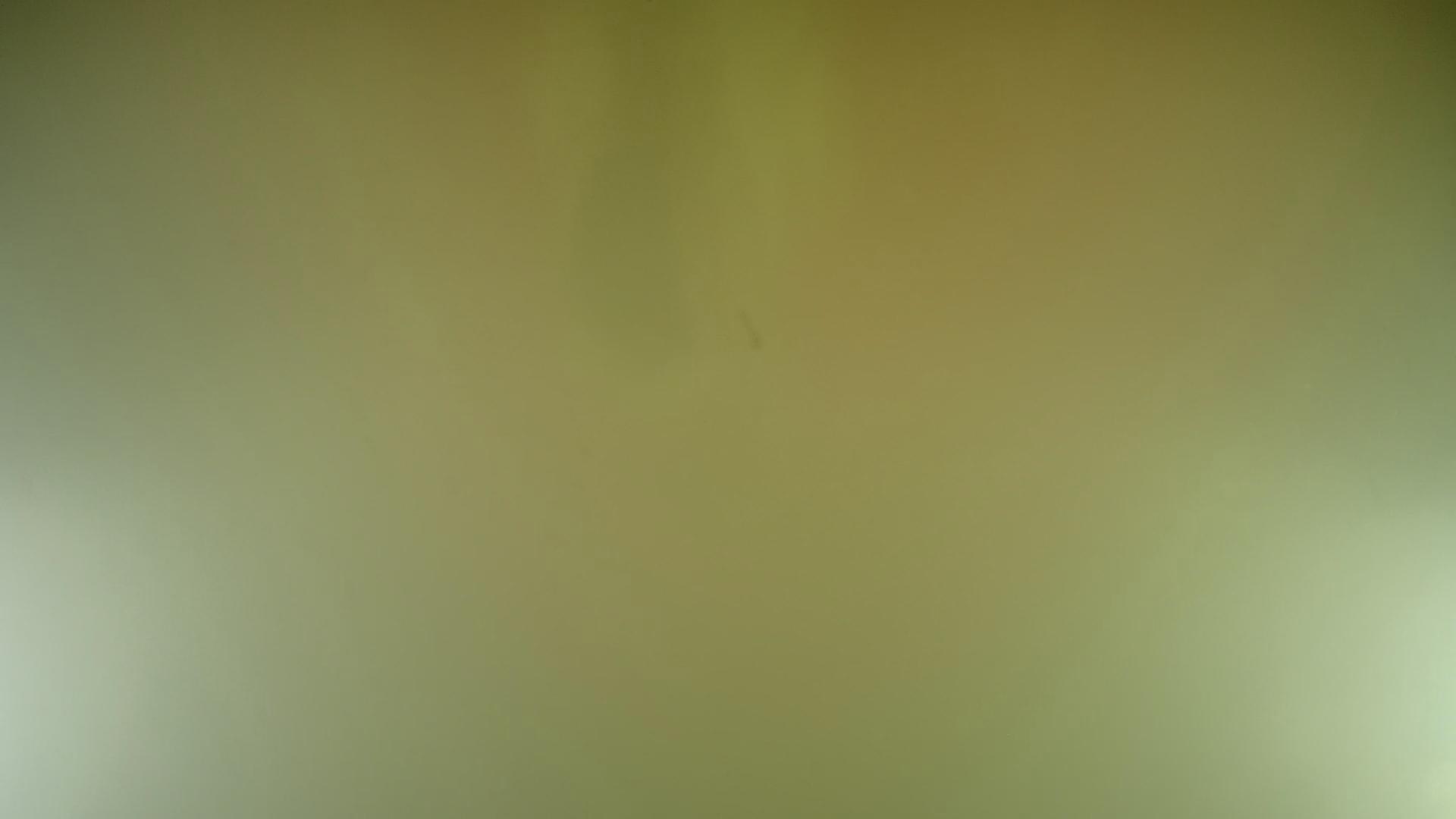}
     \end{subfigure}%
     \begin{subfigure}[b]{0.11\textwidth}
         \centering
         \includegraphics[width=\textwidth]{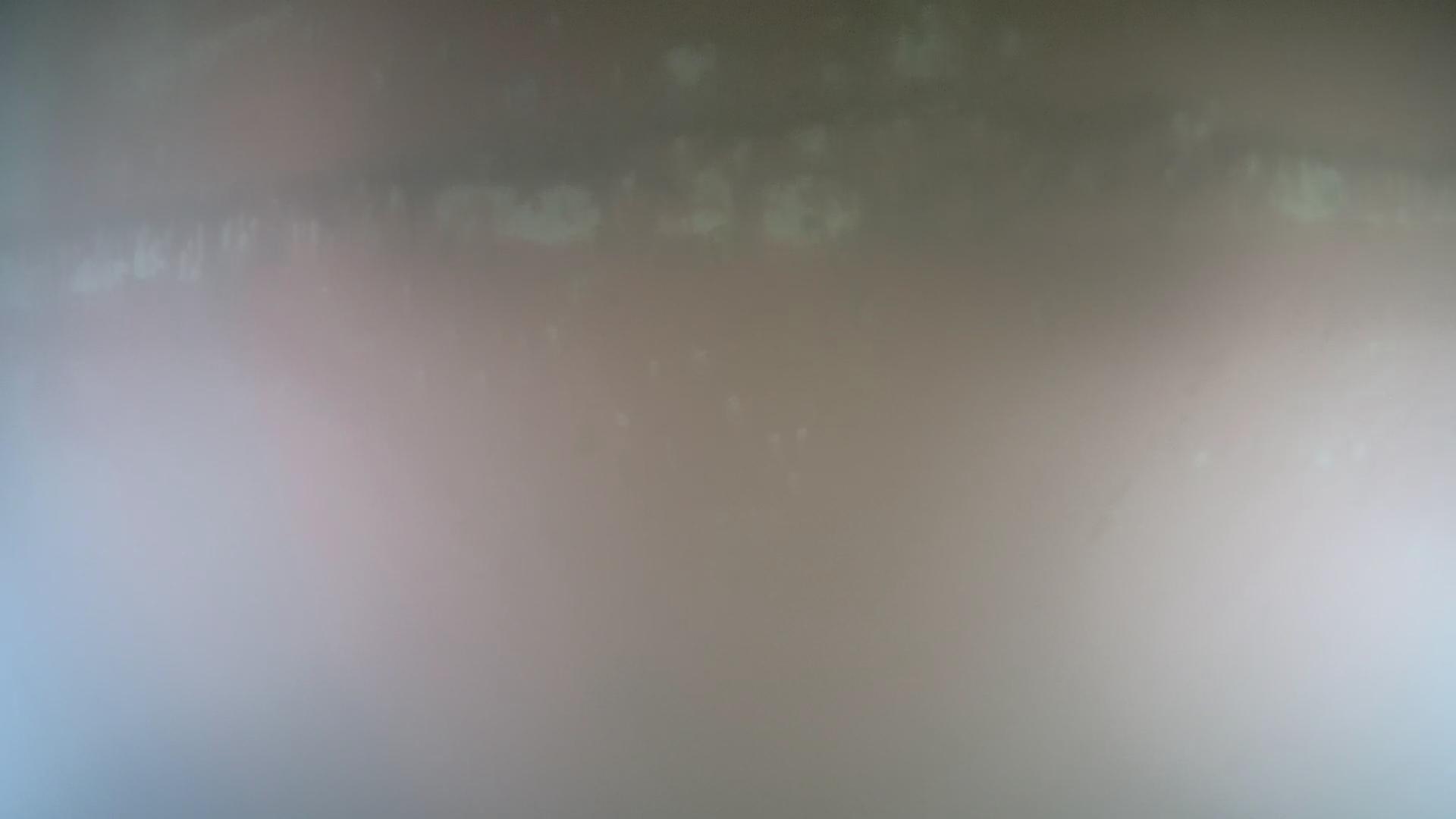}
         \includegraphics[width=\textwidth]{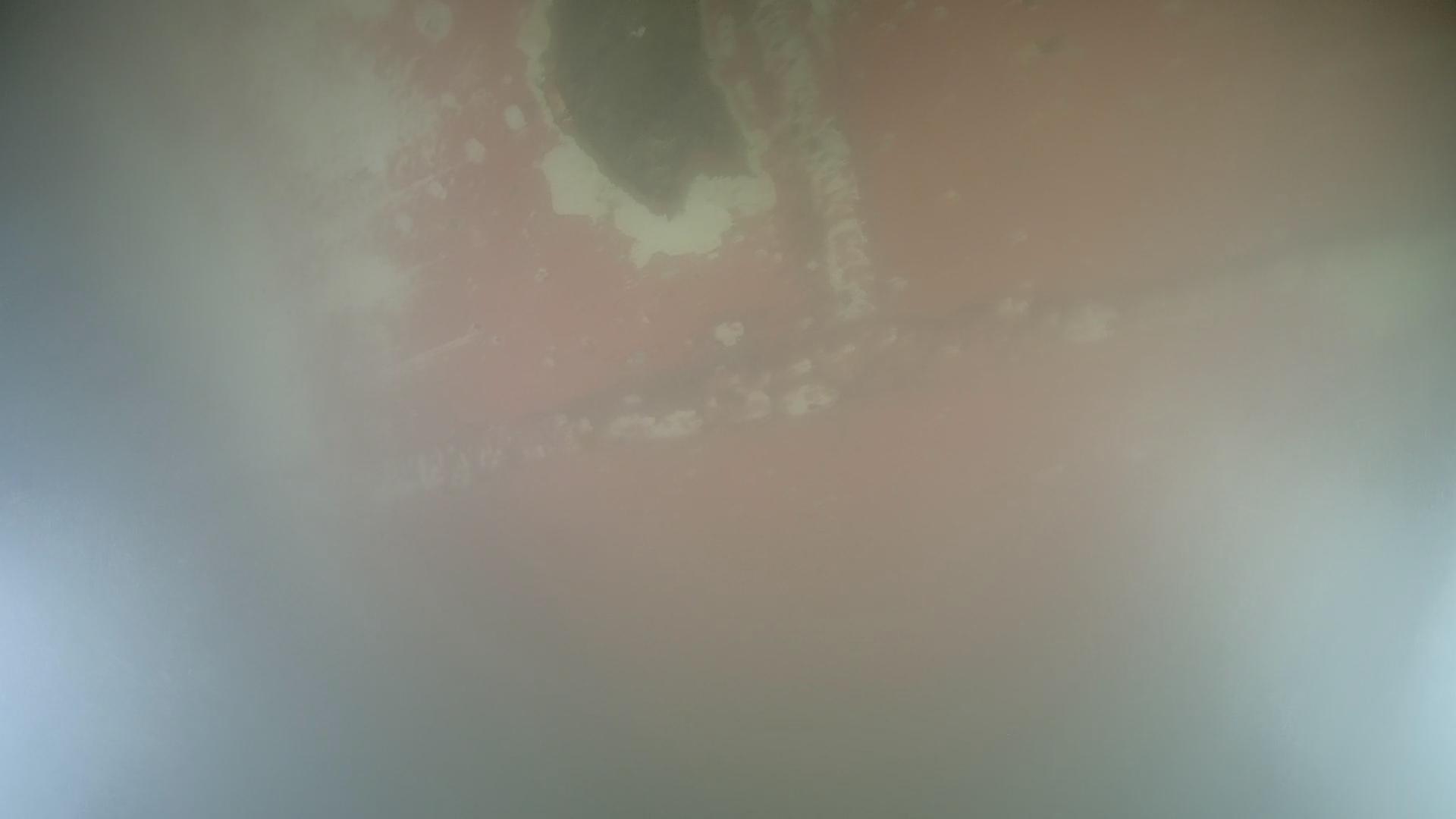}
     \end{subfigure}\\
     \begin{subfigure}[b]{1\textwidth}
         \centering
         \includegraphics[width=\textwidth]{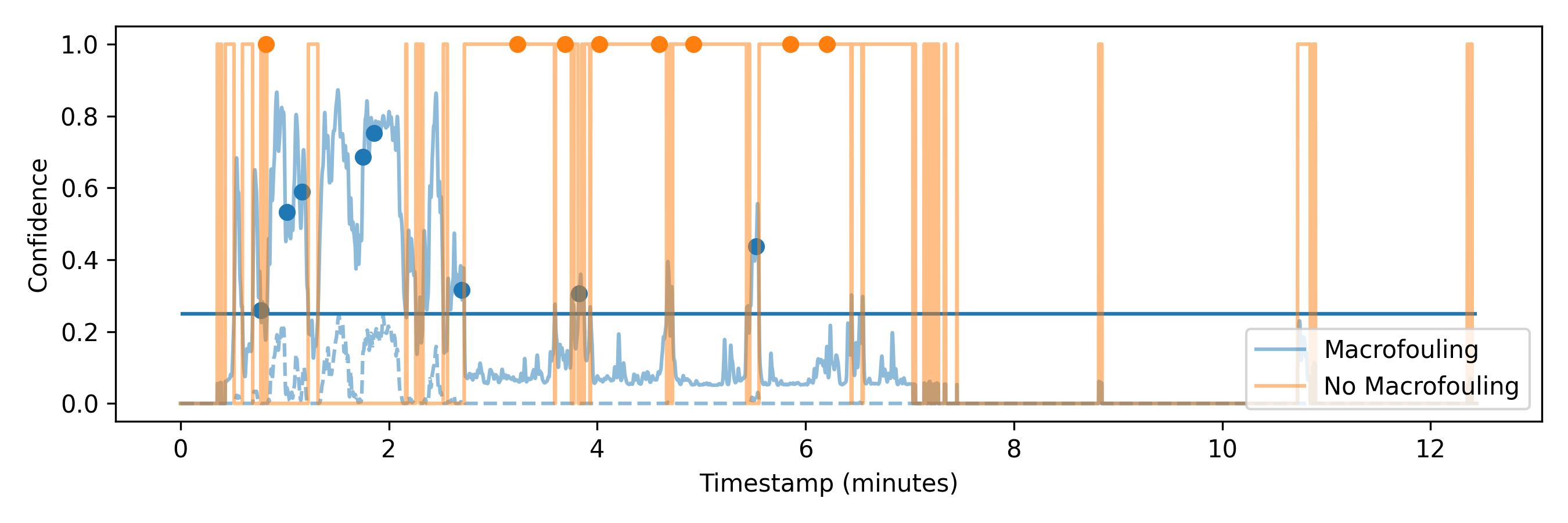}
     \end{subfigure}
        \caption{Analysis of ROV captured video using a ComFe model to detect the presence of biofouling. The dots in the timeseries represent the frames shown above the plot, which are selected using the SKMPS procedure. The dashed lines represent the proportion of fouling coverage identified by the model, and the horizontal lines show the thresholds used to classify frames as having macrofouling present. Timestamps with no hull present were filtered out by the hull detection model.}
        \label{fig:main_text_video_trial_examples_heavy_fouling}
\end{figure*}

A trial was undertaken at a marina to collect video footage with which to qualitatively evaluate the performance of the ComFe biofouling detection models. An underwater ROV was used to inspect two small commercial vessels in waters with high turbidity and poor visibility. The collected footage consisted of a number of transects of the vessels, with varying degrees of fouling being captured depending on the vessel, hull location and water depth.

For this exercise, the ComFe model using the DINOv2 ViT-B/14 (f) w/reg backbone was chosen to balance performance versus computational cost, and an appropriate confidence threshold was determined using the precision recall curve obtained on the test data as shown in \cref{fig:model_suite_average_precision_biofouling}. We aimed for a recall of 90\%, setting the confidence threshold at 0.25 which resulted in a precision of 76\%.

The output from the ComFe model over an example transect is shown in \cref{fig:main_text_video_trial_examples_heavy_fouling}, and further transects are shown in the supporting information. This figure showcases the smoothed confidence over time plots and extracted representative frames using SKMPS. It is found that while mistakes can be made for particular frames, the models can efficiently highlight points of interest within the video clip, such as the particularly fouled sea chest at the beggining of the video. The representative frames selected by SKMPS are also effective for quickly making a determination as to the nature of the fouling present, and if it poses a concern. The full set of transects is provided for download from \href{https://figshare.com/articles/dataset/Automating_the_assessment_of_biofouling_in_images/26537158}{figshare}.\cite{Mannix2024figshare}


\section{Discussion}
\label{sec:discussion}

\subsection{Deployment considerations and costs}

The ViT-B/14 ComFe model used in this paper to process video data was run efficiently using a NVIDIA T4 tensor core with 16GiB of video memory. These are readily available on cloud hosting platforms for deploying machine learning models and currently cost around \$1 USD per hour to run. This setup could process an hour of video within around 25 minutes, evaluating ten frames per second. Further throughput could be obtained by reducing the number of frames per second, or using the ViT-S rather than the ViT-B model variant, although this would compromise the quality of model outputs.

This has the potential to save significant amounts of time for agencies that currently manually review footage from biofouling inspections, and could also be used by industry groups themselves to improve confidence in assessments provided after various activities, such as in-water cleaning.

\subsection{Towards vessel-level estimates of fouling coverage}

The regulatory requirement for a clean hull standard in New Zealand specifies guidances in terms of acceptable levels of cover.\cite{georgiades2017evidence} The models developed in this paper address the need to identify regions of a vessel's hull which may contain fouling levels of concern, but are not able to provide vessel-level estimates of fouling coverage, which requires a number of additional considerations. These include (i) determining the area within a still frame that is covered by fouling vs not covered by fouling, (ii) determining the area sampled in a video, as simply looking at still frames will result in overlap of estimates, and (iii) extrapolating these transects to the whole-of-vessel level.

The first point (i) might be answerable using monocular depth perception models, of which it has been found that DINOv2 can perform quite well at in terrestrial settings with computer vision benchmarking datasets containing common scenes, such as lounge rooms or city scapes. If the appropriate dataset was used, this may translate well to the setting of vessel hulls underwater, but this has yet to be attempted. Solving point (ii) is more challenging, as it may require a the video data to be processed into a 3D surface, that to do easily may require additional metadata on diver or ROV positioning relative to the hull. Other imaging approaches, such as sonar, may be more effective at gathering such an estimate, and may also be potentially used to estimate the level of fouling present on a vessel's hull.\cite{zabin2018will} Further data collection is likely required to address the third point, which will be more achievable if the first two challenges can be addressed.

\subsection{Improving the pretrained backbone}

The ComFe approach employed in this paper uses a frozen pretrained model backbone, where the weights of the backbone are unchanged throughout training the ComFe head. While the DINOv2 models used in the work are quite effective---they have been trained on a wide variety of computer vision datasets, from iNaturalist\cite{vanhorn2018inaturalist} to ImageNet\cite{ILSVRC15}---they have not been trained on any imagery that could be considered in-domain for the problem of detecting biofouling on vessel hulls. Further work could likely improve the backbone model embeddings by using self-supervised\cite{oquab2023dinov2} learning approaches to finetune the DINOv2 models on the biofouling dataset. Alternatively, the backbone could also be improved by undertaking feature distillation using recently developed techniques with larger model variants over the biofouling data, which could potentially be a more stable approach.\cite{mannix2024faithfullabelfreeknowledgedistillation}

\section{Conclusion}

In this study we have successfully applied the ComFe interpretable-by-design approach to the challenge of identifying the presence of biofouling in images of vessel hulls, using DINOv2 ViT foundation models. We demonstrate that this approach can achieve improved performance relative to black-box-approaches, and also prior works using large CNN ensembles, and further show that the approach can efficiently screen video footage transects to identify key regions that may be of concern. All code, models and data are publicly released, and we hope this will generate further interest in developing solutions to globally improve biofouling management.

\section*{Acknowledgements}

We acknowledge the Traditional Custodians and Owners of the lands on which the work detailed in this paper was undertaken---the Wurundjeri Woi Wurrung and Bunurong people of the Kulin nation, and the Larrakia people---and pay our respects to their Elders past and present.

We would like to thank everyone that provided data towards this project. We would like to thank Ian Davidson, Oliver Floerl and Shaun Cunningham from Cawthron, and Samantha Happy from Auckland Council. We would like to thank Ashley Coutts from Biofouling Solutions. We would like to thank Sarah Bailey from Fisheries and Oceans Canada. We would like to thank Lina Ceballos and Chris Scianni from the California State Lands Commision. We would like to thank Kevin Ellard from the Tasmanian Department of Primary Industries. We would like to thank Abagail Robinson, Daniel Kluza, Charlie Parker and Katherine Walls from the New Zealand's Ministry of Primary Industries. 

This research was undertaken using the LIEF HPC-GPGPU Facility hosted at the University of Melbourne. This Facility was established with the assistance of LIEF Grant LE170100200. 
This research was also undertaken with the assistance of resources and services from the National Computational Infrastructure (NCI), which is supported by the Australian Government.
Evelyn Mannix was supported by a Australian Government Research Training Program Scholarship to complete this work.

\section*{Data availability}


Data and model weights are available from doi.org/10.6084/m9.figshare.26537158.v4. Code for training ComFe models is available at github.com/emannix/comfe-component-features.

\section*{Author contributions statement}

E.M. developed the methodology, applied it to the data, analysed the results and wrote the manuscript. B.W. collected the ROV video data, advised on project direction and led the gathering of data from other organisations.  All authors reviewed the manuscript.

\section*{Competing interests}

The authors declare no competing interests.



\bibliography{refs.bib}

\begin{thebibliography}{10}
\urlstyle{rm}
\expandafter\ifx\csname url\endcsname\relax
  \def\url#1{\texttt{#1}}\fi
\expandafter\ifx\csname urlprefix\endcsname\relax\def\urlprefix{URL }\fi
\expandafter\ifx\csname doiprefix\endcsname\relax\def\doiprefix{DOI: }\fi
\providecommand{\bibinfo}[2]{#2}
\providecommand{\eprint}[2][]{\url{#2}}

\bibitem{international2012guidelines}
\bibinfo{author}{IMO}.
\newblock \emph{\bibinfo{title}{Guidelines for the Control and Management of Ships' Biofouling: To Minimize the Transfer of Invasive Aquatic Species}} (\bibinfo{publisher}{IMO}, \bibinfo{year}{2012}).

\bibitem{georgiades2021role}
\bibinfo{author}{Georgiades, E.} \emph{et~al.}
\newblock \bibinfo{journal}{\bibinfo{title}{The role of vessel biofouling in the translocation of marine pathogens: management considerations and challenges}}.
\newblock {\emph{\JournalTitle{Frontiers in Marine Science}}} \textbf{\bibinfo{volume}{8}}, \bibinfo{pages}{660125} (\bibinfo{year}{2021}).

\bibitem{callow2002marine}
\bibinfo{author}{Callow, M.~E.} \& \bibinfo{author}{Callow, J.~A.}
\newblock \bibinfo{journal}{\bibinfo{title}{Marine biofouling: a sticky problem}}.
\newblock {\emph{\JournalTitle{Biologist}}} \textbf{\bibinfo{volume}{49}}, \bibinfo{pages}{1--5} (\bibinfo{year}{2002}).

\bibitem{schultz2011economic}
\bibinfo{author}{Schultz, M.~P.}, \bibinfo{author}{Bendick, J.}, \bibinfo{author}{Holm, E.} \& \bibinfo{author}{Hertel, W.}
\newblock \bibinfo{journal}{\bibinfo{title}{Economic impact of biofouling on a naval surface ship}}.
\newblock {\emph{\JournalTitle{Biofouling}}} \textbf{\bibinfo{volume}{27}}, \bibinfo{pages}{87--98} (\bibinfo{year}{2011}).

\bibitem{davidson2016mini}
\bibinfo{author}{Davidson, I.} \emph{et~al.}
\newblock \bibinfo{journal}{\bibinfo{title}{Mini-review: Assessing the drivers of ship biofouling management--aligning industry and biosecurity goals}}.
\newblock {\emph{\JournalTitle{Biofouling}}} \textbf{\bibinfo{volume}{32}}, \bibinfo{pages}{411--428} (\bibinfo{year}{2016}).

\bibitem{zabin2018will}
\bibinfo{author}{Zabin, C.} \emph{et~al.}
\newblock \bibinfo{journal}{\bibinfo{title}{How will vessels be inspected to meet emerging biofouling regulations for the prevention of marine invasions?}}
\newblock {\emph{\JournalTitle{Management of Biological Invasions}}} \textbf{\bibinfo{volume}{9}}, \bibinfo{pages}{195--208} (\bibinfo{year}{2018}).

\bibitem{DAFF2023biofouling}
\bibinfo{author}{{Department of Agriculture, Fisheries and Forestry}}.
\newblock \bibinfo{journal}{\bibinfo{title}{Australian biofouling management requirements.}}
\newblock {\emph{\JournalTitle{{Department of Agriculture, Fisheries and Forestry}}}}  (\bibinfo{year}{2023}).

\bibitem{scianni2021yes}
\bibinfo{author}{Scianni, C.}, \bibinfo{author}{Lubarsky, K.}, \bibinfo{author}{Ceballos-Osuna, L.} \& \bibinfo{author}{Bates, T.}
\newblock \bibinfo{journal}{\bibinfo{title}{Yes, we canz: initial compliance and lessons learned from regulating vessel biofouling management in california and new zealand.}}
\newblock {\emph{\JournalTitle{Management of Biological Invasions}}} \textbf{\bibinfo{volume}{12}} (\bibinfo{year}{2021}).

\bibitem{mannix2021automating}
\bibinfo{author}{Mannix, E.~J.}, \bibinfo{author}{Wei, S.}, \bibinfo{author}{A.~Woodham, B.}, \bibinfo{author}{Wilkinson, P.} \& \bibinfo{author}{Robinson, A.~P.}
\newblock \bibinfo{journal}{\bibinfo{title}{Automating the assessment of biofouling in images using expert agreement as a gold standard}}.
\newblock {\emph{\JournalTitle{Scientific reports}}} \textbf{\bibinfo{volume}{11}}, \bibinfo{pages}{2739} (\bibinfo{year}{2021}).

\bibitem{krause2023semantic}
\bibinfo{author}{Krause, L.~M.} \emph{et~al.}
\newblock \bibinfo{journal}{\bibinfo{title}{Semantic segmentation for fully automated macrofouling analysis on coatings after field exposure}}.
\newblock {\emph{\JournalTitle{Biofouling}}} \textbf{\bibinfo{volume}{39}}, \bibinfo{pages}{64--79} (\bibinfo{year}{2023}).

\bibitem{rudin2019stop}
\bibinfo{author}{Rudin, C.}
\newblock \bibinfo{journal}{\bibinfo{title}{Stop explaining black box machine learning models for high stakes decisions and use interpretable models instead}}.
\newblock {\emph{\JournalTitle{Nature machine intelligence}}} \textbf{\bibinfo{volume}{1}}, \bibinfo{pages}{206--215} (\bibinfo{year}{2019}).

\bibitem{Neuhaus_2023_ICCV}
\bibinfo{author}{Neuhaus, Y.}, \bibinfo{author}{Augustin, M.}, \bibinfo{author}{Boreiko, V.} \& \bibinfo{author}{Hein, M.}
\newblock \bibinfo{title}{Spurious features everywhere - large-scale detection of harmful spurious features in imagenet}.
\newblock In \emph{\bibinfo{booktitle}{Proceedings of the IEEE/CVF International Conference on Computer Vision (ICCV)}}, \bibinfo{pages}{20235--20246} (\bibinfo{year}{2023}).

\bibitem{moser2017quantifying}
\bibinfo{author}{Moser, C.~S.} \emph{et~al.}
\newblock \bibinfo{journal}{\bibinfo{title}{Quantifying the extent of niche areas in the global fleet of commercial ships: the potential for “super-hot spots” of biofouling}}.
\newblock {\emph{\JournalTitle{Biological Invasions}}} \textbf{\bibinfo{volume}{19}}, \bibinfo{pages}{1745--1759} (\bibinfo{year}{2017}).

\bibitem{chen2019looks}
\bibinfo{author}{Chen, C.} \emph{et~al.}
\newblock \bibinfo{title}{This looks like that: Deep learning for interpretable image recognition} (\bibinfo{year}{2019}).
\newblock \eprint{1806.10574}.

\bibitem{mannix2024scalable}
\bibinfo{author}{Mannix, E.~J.}, \bibinfo{author}{Hodgkinson, L.} \& \bibinfo{author}{Bondell, H.}
\newblock \bibinfo{title}{Comfe: An interpretable head for vision transformers} (\bibinfo{year}{2025}).
\newblock \eprint{2403.04125}.

\bibitem{oquab2023dinov2}
\bibinfo{author}{Oquab, M.} \emph{et~al.}
\newblock \bibinfo{journal}{\bibinfo{title}{Dinov2: Learning robust visual features without supervision}}.
\newblock {\emph{\JournalTitle{arXiv preprint arXiv:2304.07193}}}  (\bibinfo{year}{2023}).

\bibitem{bhalla2024interpreting}
\bibinfo{author}{Bhalla, U.}, \bibinfo{author}{Oesterling, A.}, \bibinfo{author}{Srinivas, S.}, \bibinfo{author}{Calmon, F.} \& \bibinfo{author}{Lakkaraju, H.}
\newblock \bibinfo{title}{Interpreting {CLIP} with sparse linear concept embeddings (spli{CE})}.
\newblock In \emph{\bibinfo{booktitle}{The Thirty-eighth Annual Conference on Neural Information Processing Systems}} (\bibinfo{year}{2024}).

\bibitem{floerl2005risk}
\bibinfo{author}{Floerl, O.}, \bibinfo{author}{Inglis, G.~J.} \& \bibinfo{author}{Hayden, B.~J.}
\newblock \bibinfo{journal}{\bibinfo{title}{A risk-based predictive tool to prevent accidental introductions of nonindigenous marine species}}.
\newblock {\emph{\JournalTitle{Environmental management}}} \textbf{\bibinfo{volume}{35}}, \bibinfo{pages}{765--778} (\bibinfo{year}{2005}).

\bibitem{liu2019binormal}
\bibinfo{author}{Liu, Z.} \& \bibinfo{author}{Bondell, H.~D.}
\newblock \bibinfo{journal}{\bibinfo{title}{Binormal precision--recall curves for optimal classification of imbalanced data}}.
\newblock {\emph{\JournalTitle{Statistics in Biosciences}}} \textbf{\bibinfo{volume}{11}}, \bibinfo{pages}{141--161} (\bibinfo{year}{2019}).

\bibitem{Mannix2024figshare}
\bibinfo{author}{Mannix, E.}
\newblock \bibinfo{journal}{\bibinfo{title}{{Automating the assessment of biofouling in images and video footage}}}.
\newblock {\emph{\JournalTitle{Figshare}}} \doiprefix\url{10.6084/m9.figshare.26537158.v4} (\bibinfo{year}{2024}).

\bibitem{darcet2023vision}
\bibinfo{author}{Darcet, T.}, \bibinfo{author}{Oquab, M.}, \bibinfo{author}{Mairal, J.} \& \bibinfo{author}{Bojanowski, P.}
\newblock \bibinfo{journal}{\bibinfo{title}{Vision transformers need registers}}.
\newblock {\emph{\JournalTitle{arXiv preprint arXiv:2309.16588}}}  (\bibinfo{year}{2023}).

\bibitem{shankar2021image}
\bibinfo{author}{Shankar, V.} \emph{et~al.}
\newblock \bibinfo{title}{Do image classifiers generalize across time?}
\newblock In \emph{\bibinfo{booktitle}{Proceedings of the IEEE/CVF International Conference on Computer Vision}}, \bibinfo{pages}{9661--9669} (\bibinfo{year}{2021}).

\bibitem{hardle1992kernel}
\bibinfo{author}{H{\"a}rdle, W.} \& \bibinfo{author}{Vieu, P.}
\newblock \bibinfo{journal}{\bibinfo{title}{Kernel regression smoothing of time series}}.
\newblock {\emph{\JournalTitle{Journal of Time Series Analysis}}} \textbf{\bibinfo{volume}{13}}, \bibinfo{pages}{209--232} (\bibinfo{year}{1992}).

\bibitem{mannix2024efficient}
\bibinfo{author}{Mannix, E.} \& \bibinfo{author}{Bondell, H.}
\newblock \bibinfo{title}{A mixture of exemplars approach for efficient out-of-distribution detection with foundation models} (\bibinfo{year}{2025}).
\newblock \eprint{2311.17093}.

\bibitem{georgiades2017evidence}
\bibinfo{author}{Georgiades, E.} \& \bibinfo{author}{Kluza, D.}
\newblock \bibinfo{journal}{\bibinfo{title}{Evidence-based decision making to underpin the thresholds in new zealand's craft risk management standard: biofouling on vessels arriving to new zealand}}.
\newblock {\emph{\JournalTitle{Marine Technology Society Journal}}} \textbf{\bibinfo{volume}{51}}, \bibinfo{pages}{76--88} (\bibinfo{year}{2017}).

\bibitem{vanhorn2018inaturalist}
\bibinfo{author}{Horn, G.~V.} \emph{et~al.}
\newblock \bibinfo{title}{The inaturalist species classification and detection dataset} (\bibinfo{year}{2018}).
\newblock \eprint{1707.06642}.

\bibitem{ILSVRC15}
\bibinfo{author}{Russakovsky, O.} \emph{et~al.}
\newblock \bibinfo{journal}{\bibinfo{title}{{ImageNet Large Scale Visual Recognition Challenge}}}.
\newblock {\emph{\JournalTitle{International Journal of Computer Vision (IJCV)}}} \textbf{\bibinfo{volume}{115}}, \bibinfo{pages}{211--252}, \doiprefix\url{10.1007/s11263-015-0816-y} (\bibinfo{year}{2015}).

\bibitem{mannix2024faithfullabelfreeknowledgedistillation}
\bibinfo{author}{Mannix, E.~J.}, \bibinfo{author}{Hodgkinson, L.} \& \bibinfo{author}{Bondell, H.}
\newblock \bibinfo{title}{Preserving angles improves feature distillation of foundation models} (\bibinfo{year}{2025}).
\newblock \eprint{2411.15239}.

\end{thebibliography}

\newpage
\newpage

\maketitlesupplementaryalt

\appendix

\renewcommand{\thefigure}{S\arabic{figure}}
\renewcommand{\thetable}{S\arabic{table}}

\section{Detecting biofouling --- class prototypes}
\label{app:biofouling_class_prototypes}

\begin{figure*}[!h]
     \centering
     \begin{subfigure}[b]{0.45\textwidth}
         \centering
         \includegraphics[width=\textwidth]{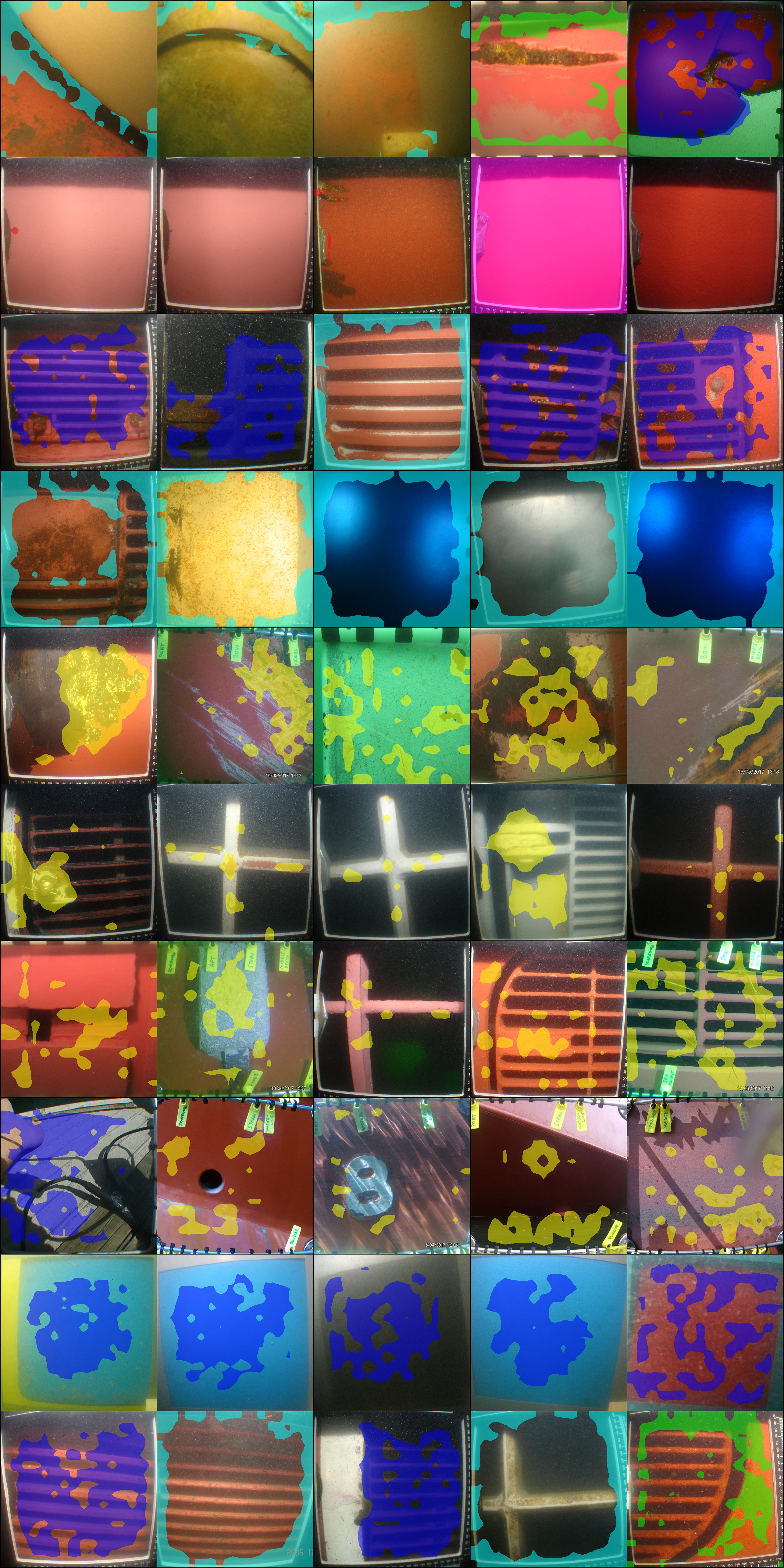}
         \caption{No Macrofouling}
     \end{subfigure}%
     \hfill{}
     \begin{subfigure}[b]{0.45\textwidth}
         \centering
         \includegraphics[width=\textwidth]{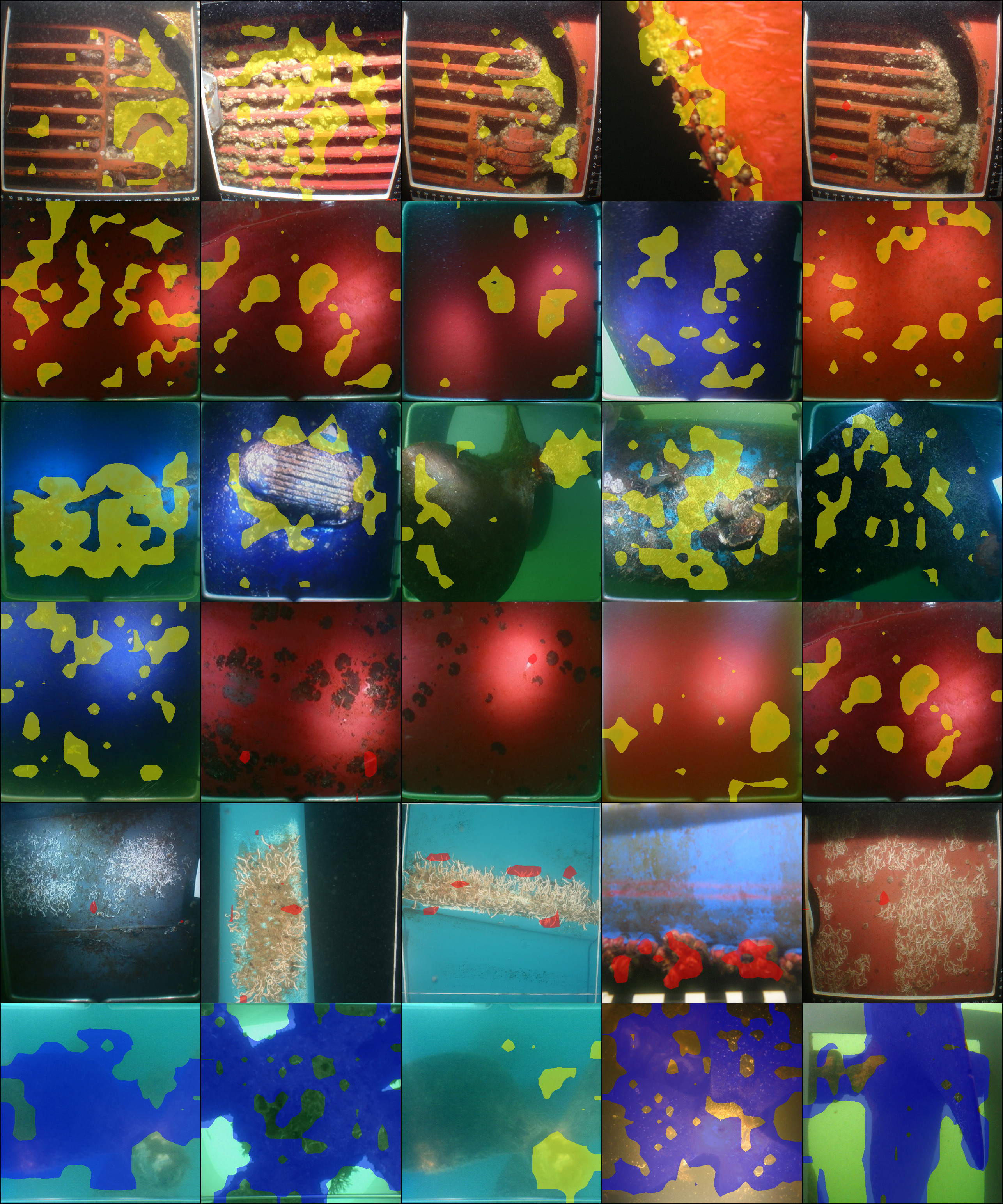}
         \caption{Macrofouling Present}
     \end{subfigure}
        \caption{Class prototype exemplars for detecting the presence of biofouling. Each row represents a different class prototype, with the most commonly used prototype occurring first. Five exemplars from the training dataset are used to visualise each class prototype. }
        \label{fig:class_prototype_exemplars_biofouling}
\end{figure*}

\clearpage
\section{Detecting biofouling --- further transects}
\label{app:biofouling_further_transects}

\begin{figure*}[!h]
     \centering
     \begin{subfigure}[b]{0.11\textwidth}
         \centering
        \begin{overpic}[height=0.6\textwidth, width=\textwidth]{example_images/1280px-HD_transparent_picture.png}
         \put (0,20) {\parbox{1.5cm}{\centering \small Fouling}}
        \end{overpic}
        \begin{overpic}[height=0.6\textwidth, width=\textwidth]{example_images/1280px-HD_transparent_picture.png}
         \put (0,20) {\parbox{1.5cm}{\centering \small No\\Fouling}}
        \end{overpic}
     \end{subfigure}%
     \begin{subfigure}[b]{0.11\textwidth}
         \centering
         \includegraphics[width=\textwidth]{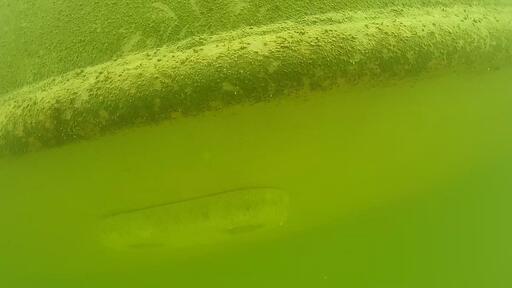}
         \includegraphics[width=\textwidth]{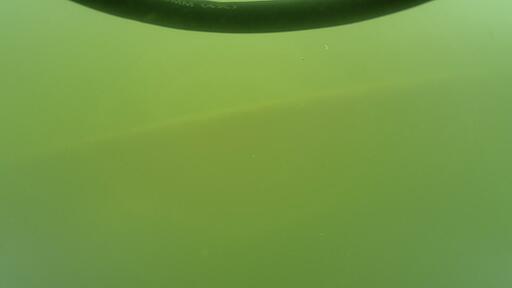}
     \end{subfigure}%
     \begin{subfigure}[b]{0.11\textwidth}
         \centering
         \includegraphics[width=\textwidth]{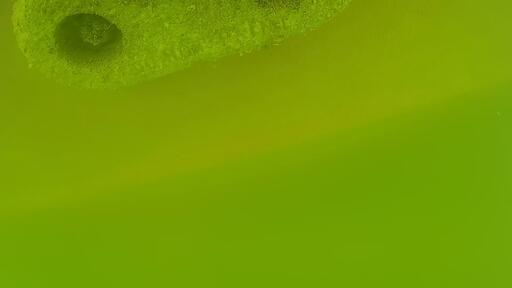}
         \includegraphics[width=\textwidth]{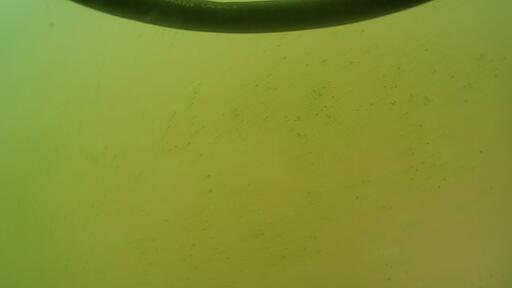}
     \end{subfigure}%
     \begin{subfigure}[b]{0.11\textwidth}
         \centering
         \includegraphics[width=\textwidth]{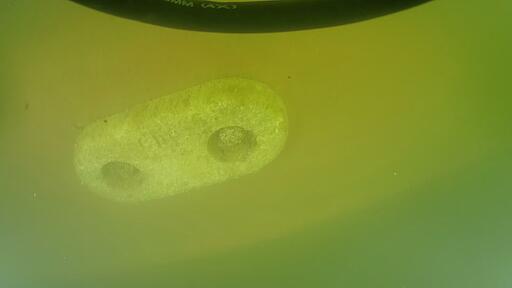}
         \includegraphics[width=\textwidth]{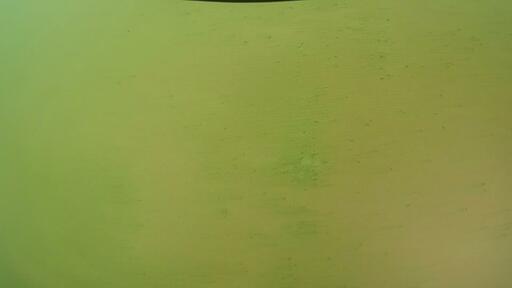}
     \end{subfigure}%
     \begin{subfigure}[b]{0.11\textwidth}
         \centering
         \includegraphics[width=\textwidth]{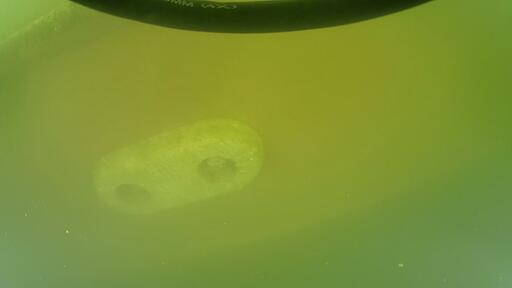}
         \includegraphics[width=\textwidth]{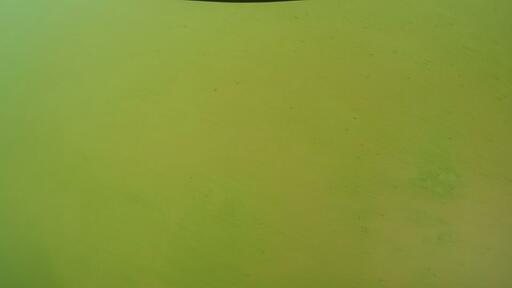}
     \end{subfigure}%
     \begin{subfigure}[b]{0.11\textwidth}
         \centering
         \includegraphics[width=\textwidth]{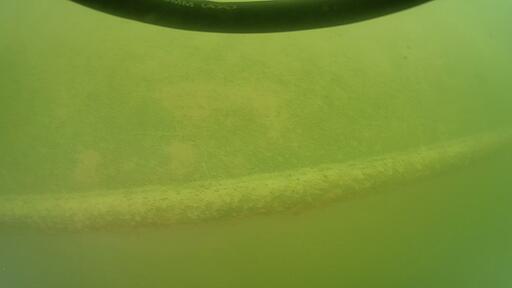}
         \includegraphics[width=\textwidth]{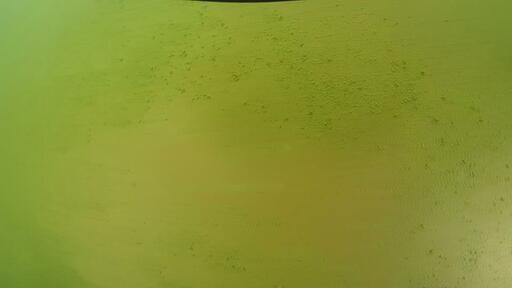}
     \end{subfigure}%
     \begin{subfigure}[b]{0.11\textwidth}
         \centering
         \includegraphics[width=\textwidth]{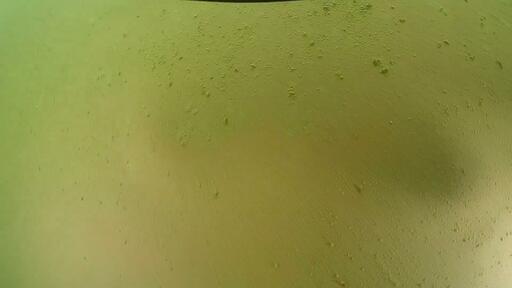}
         \includegraphics[width=\textwidth]{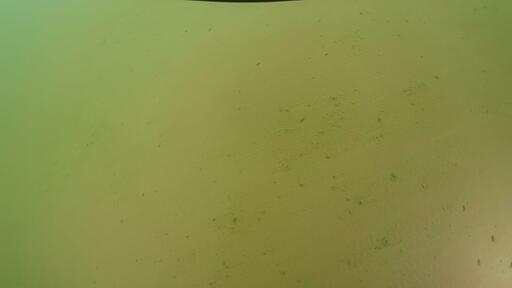}
     \end{subfigure}%
     \begin{subfigure}[b]{0.11\textwidth}
         \centering
         \includegraphics[width=\textwidth]{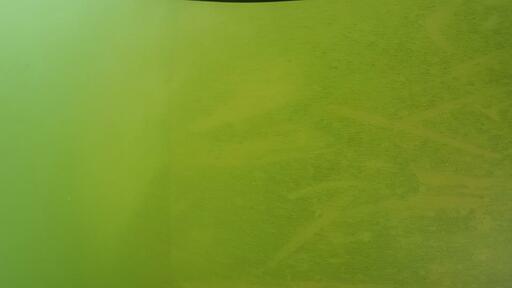}
         \includegraphics[width=\textwidth]{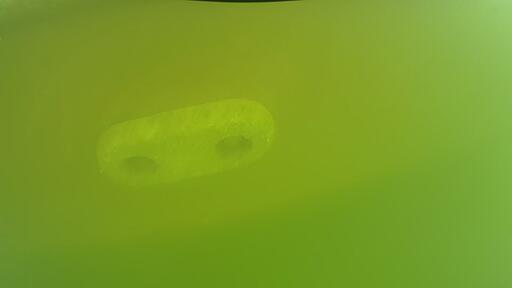}
     \end{subfigure}%
     \begin{subfigure}[b]{0.11\textwidth}
         \centering
         \includegraphics[width=\textwidth]{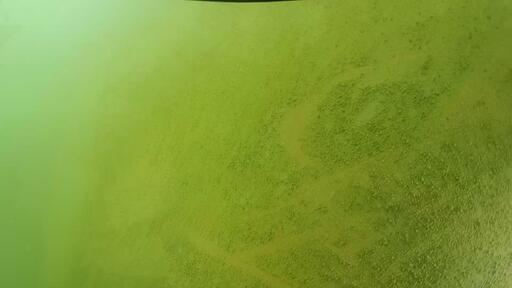}
         \includegraphics[width=\textwidth]{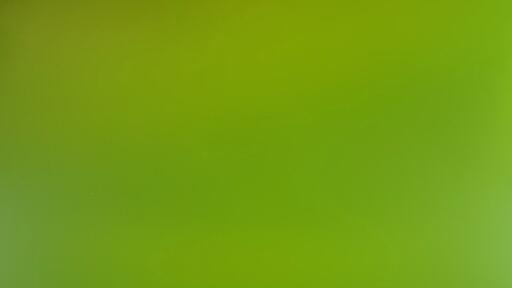}
     \end{subfigure}\\
     \begin{subfigure}[b]{1\textwidth}
         \centering
         \includegraphics[width=\textwidth]{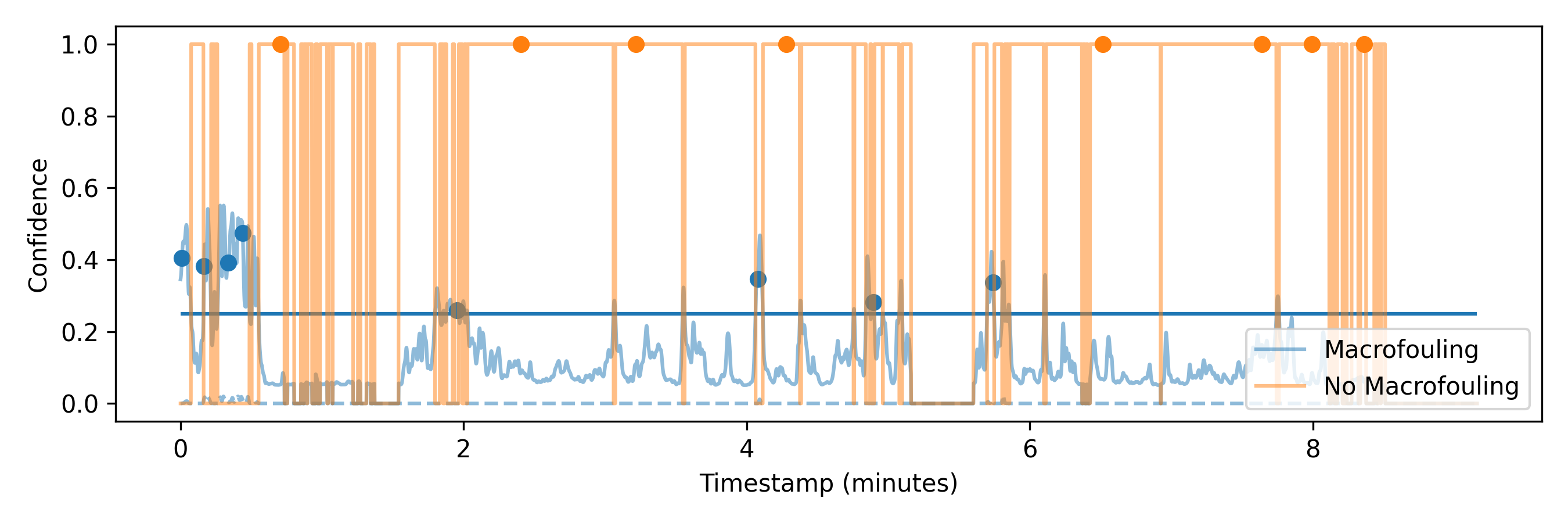}
     \end{subfigure}
        \caption[Analysis of ROV captured video with light fouling for macrofouling using the ComFe model. ]{Analysis of ROV captured video with light fouling for macrofouling using the ComFe model. The dots in the timeseries represent the frames shown above the plot, which are selected using the SKMPS method. The dashed lines represent the proportion of fouling coverage predicted by the ComFe head, and the horizontal lines show the thresholds used to classify frames as having macrofouling present.}
        \label{fig:prototype_examples}
\end{figure*}

\begin{figure*}[!h]
     \centering
     \begin{subfigure}[b]{0.11\textwidth}
         \centering
        \begin{overpic}[height=0.6\textwidth, width=\textwidth]{example_images/1280px-HD_transparent_picture.png}
         \put (0,20) {\parbox{1.5cm}{\centering \small Fouling}}
        \end{overpic}
        \begin{overpic}[height=0.6\textwidth, width=\textwidth]{example_images/1280px-HD_transparent_picture.png}
         \put (0,20) {\parbox{1.5cm}{\centering \small No\\Fouling}}
        \end{overpic}
     \end{subfigure}%
     \begin{subfigure}[b]{0.11\textwidth}
         \centering
         \includegraphics[width=\textwidth]{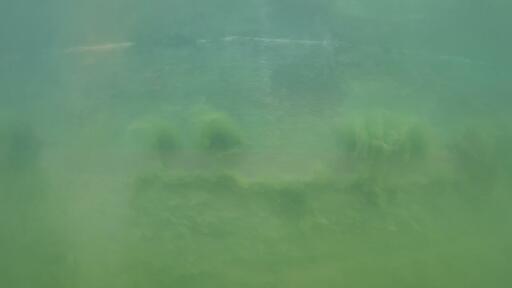}
         \includegraphics[width=\textwidth]{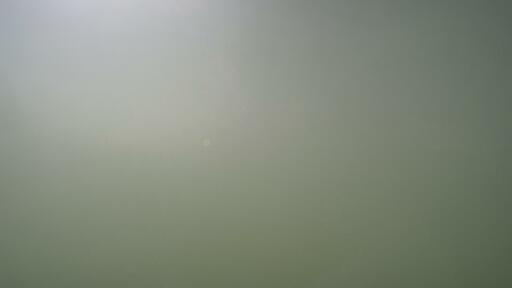}
     \end{subfigure}%
     \begin{subfigure}[b]{0.11\textwidth}
         \centering
         \includegraphics[width=\textwidth]{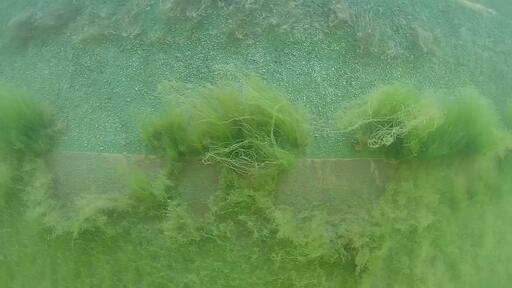}
         \includegraphics[width=\textwidth]{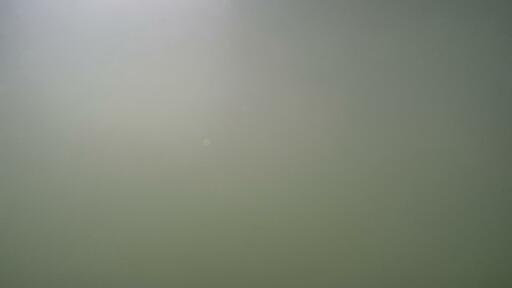}
     \end{subfigure}%
     \begin{subfigure}[b]{0.11\textwidth}
         \centering
         \includegraphics[width=\textwidth]{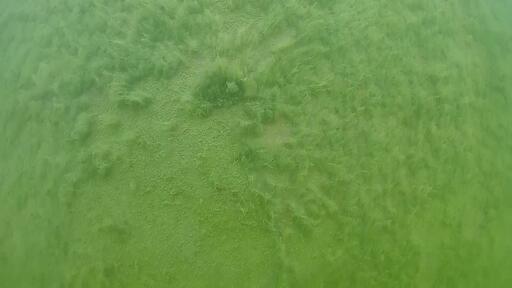}
         \includegraphics[width=\textwidth]{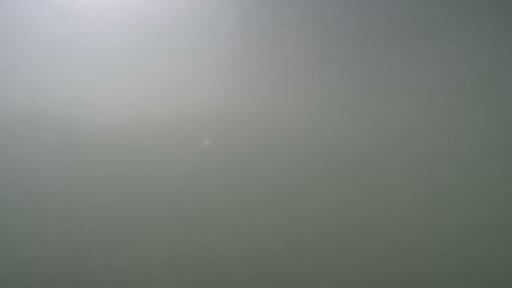}
     \end{subfigure}%
     \begin{subfigure}[b]{0.11\textwidth}
         \centering
         \includegraphics[width=\textwidth]{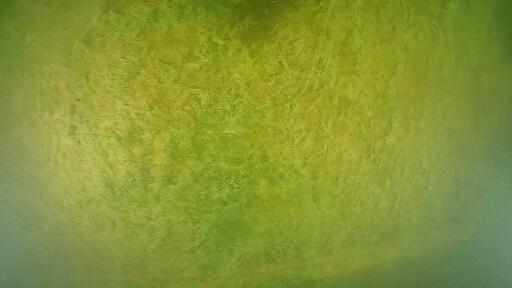}
         \includegraphics[width=\textwidth]{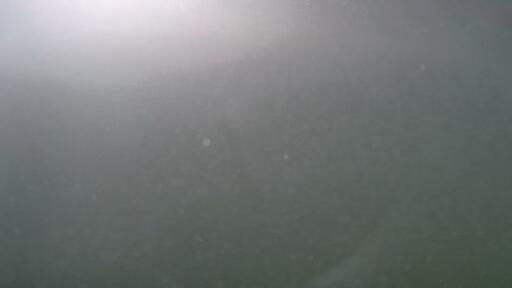}
     \end{subfigure}%
     \begin{subfigure}[b]{0.11\textwidth}
         \centering
         \includegraphics[width=\textwidth]{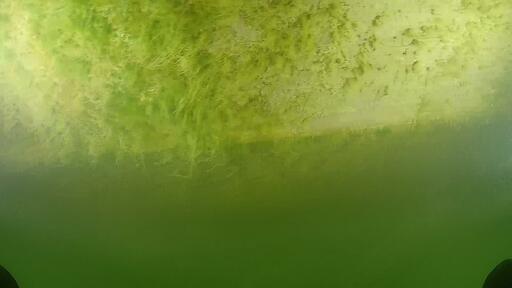}
         \includegraphics[width=\textwidth]{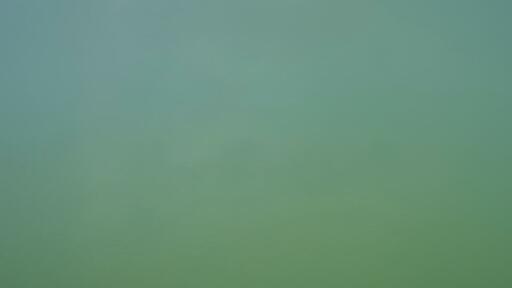}
     \end{subfigure}%
     \begin{subfigure}[b]{0.11\textwidth}
         \centering
         \includegraphics[width=\textwidth]{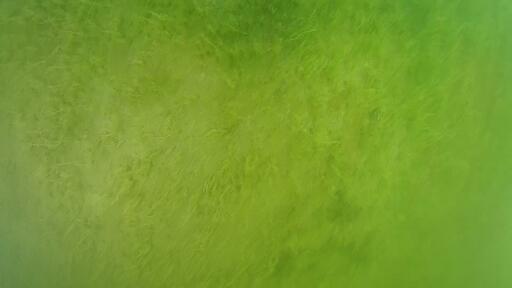}
         \includegraphics[width=\textwidth]{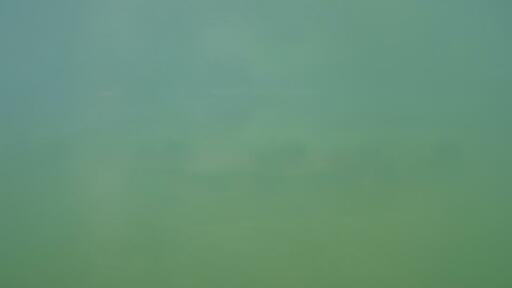}
     \end{subfigure}%
     \begin{subfigure}[b]{0.11\textwidth}
         \centering
         \includegraphics[width=\textwidth]{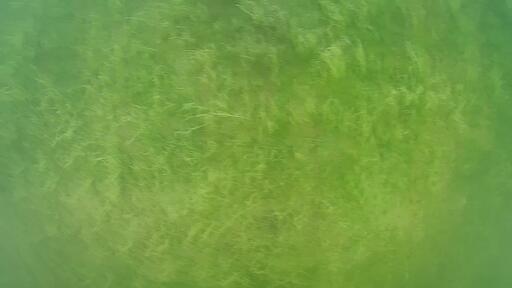}
         \includegraphics[width=\textwidth]{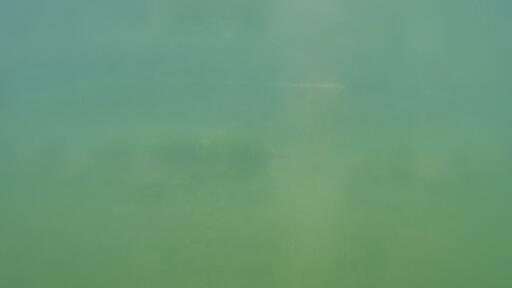}
     \end{subfigure}%
     \begin{subfigure}[b]{0.11\textwidth}
         \centering
         \includegraphics[width=\textwidth]{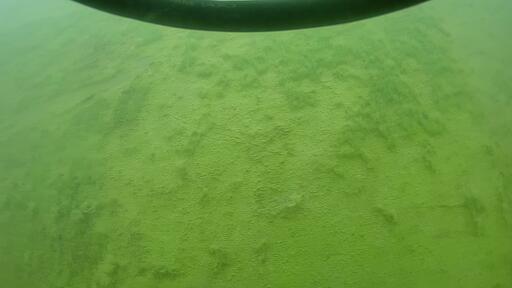}
         \includegraphics[width=\textwidth]{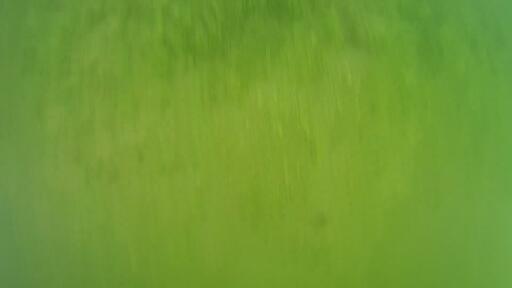}
     \end{subfigure}\\
     \begin{subfigure}[b]{1\textwidth}
         \centering
         \includegraphics[width=\textwidth]{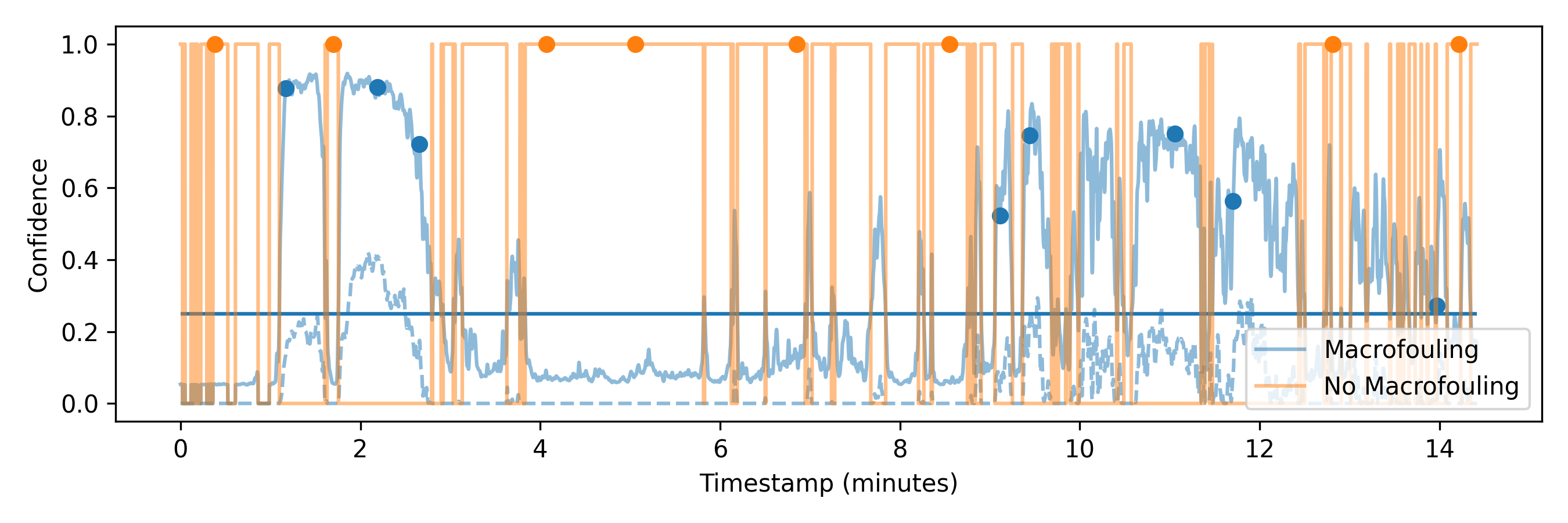}
     \end{subfigure}
        \caption[Analysis of ROV captured video with heavy fouling for macrofouling using the ComFe model.]{Analysis of ROV captured video with heavy fouling for macrofouling using the ComFe model. The dots in the timeseries represent the frames shown above the plot, which are selected using the SKMPS method. The dashed lines represent the proportion of fouling coverage predicted by the ComFe head, and the horizontal lines show the thresholds used to classify frames as having macrofouling present.}
        \label{fig:video_trial_examples_heavy_fouling}
\end{figure*}

\FloatBarrier

\clearpage
\section{Detecting vessel hulls}
\label{app:detecting_vessel_hulls}

To enable effective deployment of the biofouling model, footage not containing a vessel hull was filtered using a hull detection model. A specialised dataset was created for this purpose, which included the dataset used to train the biofouling model in addition to a diverse image set sampled from videos collected by divers and ROVs which contained frames where a vessel hull was not present. This included both instances where there was no objects visible within the water, or where above-water footage was taken. A sample of images from this dataset is shown in \cref{tbl:dataset_hulldetect}.

\begin{figure*}[!b]
     \centering
     \begin{subfigure}[b]{0.3\textwidth}
         \centering
         \caption{No vessel hull present}
         \includegraphics[width=\textwidth]{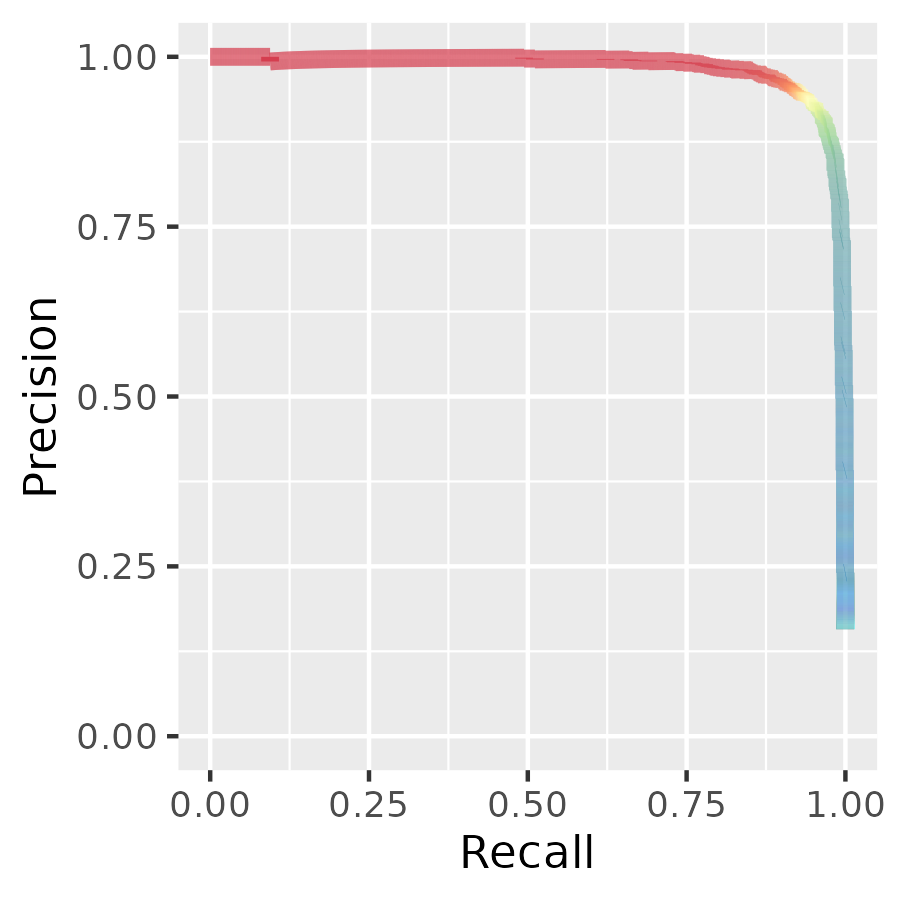}
     \end{subfigure}%
     \begin{subfigure}[b]{0.09\textwidth}
         \picdims[height=4.5cm]{\textwidth}{3.5cm}{plots/2024_biofouling_prauc_legend.png}
         \vspace{1cm}
     \end{subfigure} \\
        \caption[Precision-recall curve for the best ComFe model for detecting the presence of a vessel hull. ]{Precision-recall curve for the best ComFe model for detecting the presence of a vessel hull. The model was trained using a ComFe head on the DINOv2 ViT-B/14 (f) w/reg backbone.}
        \label{fig:model_suite_average_precision}
\end{figure*}

\begin{figure*}[!tb]
     \centering
     \begin{subfigure}[b]{0.12\textwidth}
         \centering
        \begin{overpic}[height=1.25cm, width=\textwidth]{example_images/1280px-HD_transparent_picture.png}
         \put (0,30) {\parbox{2cm}{\centering Input image}}
        \end{overpic}
        \begin{overpic}[height=1.25cm, width=\textwidth]{example_images/1280px-HD_transparent_picture.png}
         \put (0,30) {\parbox{2cm}{\centering Class\\prediction}}
        \end{overpic}
     \end{subfigure}
     \begin{subfigure}[b]{0.12\textwidth}
         \centering
         \caption{Hull}
         \label{fig:prototype_examples_pets_2}
         \picdims[height=1.25cm]{\textwidth}{1.25cm}{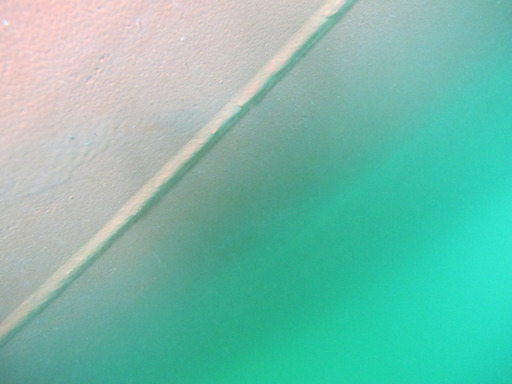}
         \picdims[height=1.25cm]{\textwidth}{1.25cm}{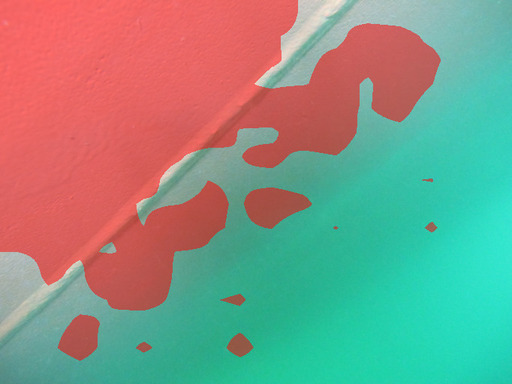}
     \end{subfigure}
     \begin{subfigure}[b]{0.12\textwidth}
         \centering
         \caption{Hull}
         \label{fig:prototype_examples_planes_1}
         \picdims[height=1.25cm]{\textwidth}{1.25cm}{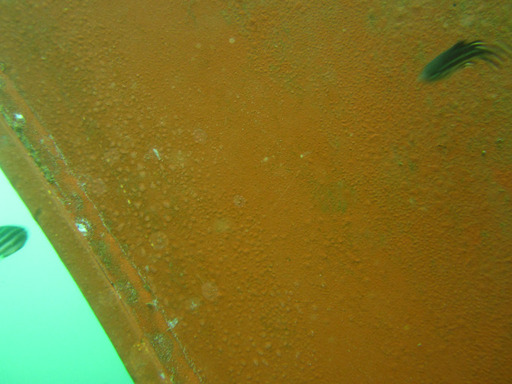}
         \picdims[height=1.25cm]{\textwidth}{1.25cm}{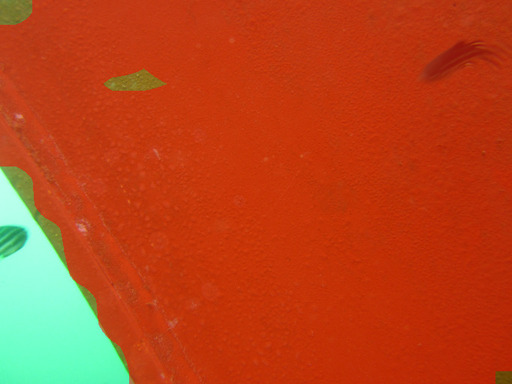}
     \end{subfigure}
     \begin{subfigure}[b]{0.12\textwidth}
         \centering
         \caption{Hull}
         \label{fig:prototype_examples_planes_2}
         \picdims[height=1.25cm]{\textwidth}{1.25cm}{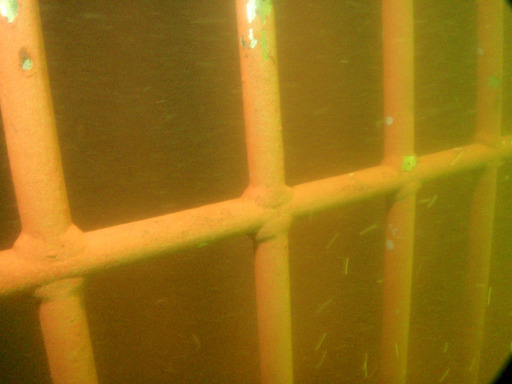}
         \picdims[height=1.25cm]{\textwidth}{1.25cm}{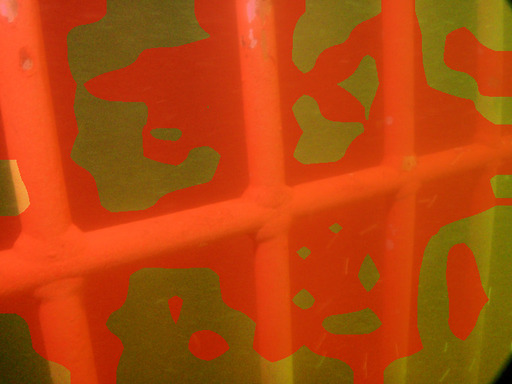}
     \end{subfigure}
     \begin{subfigure}[b]{0.12\textwidth}
         \centering
         \caption{No hull}
         \label{fig:prototype_examples_cars_1}
         \picdims[height=1.25cm]{\textwidth}{1.25cm}{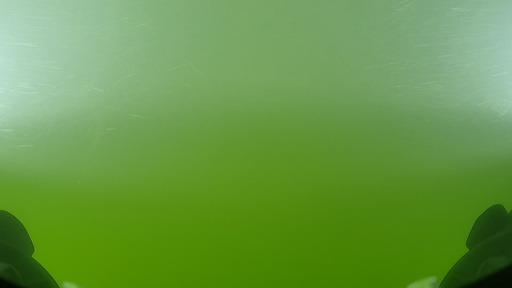}
         \picdims[height=1.25cm]{\textwidth}{1.25cm}{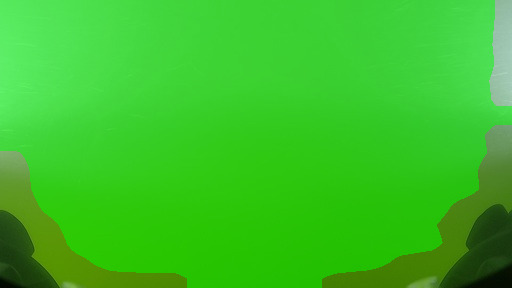}
     \end{subfigure}
     \begin{subfigure}[b]{0.12\textwidth}
         \centering
         \caption{No hull}
         \label{fig:prototype_examples_cars_2}
         \picdims[height=1.25cm]{\textwidth}{1.25cm}{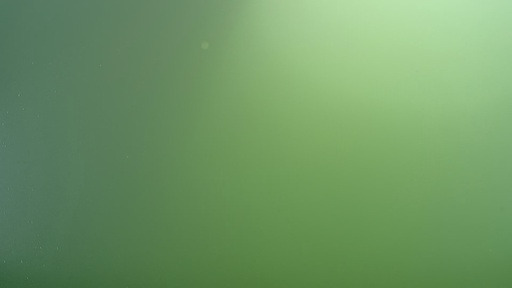}
         \picdims[height=1.25cm]{\textwidth}{1.25cm}{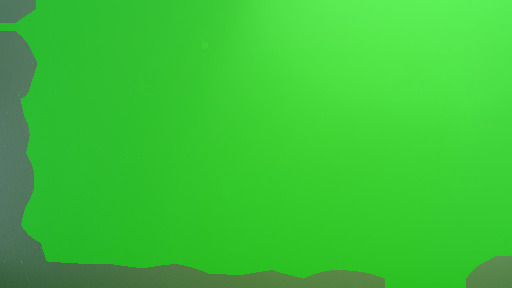}
     \end{subfigure}
     \begin{subfigure}[b]{0.12\textwidth}
         \centering
         \caption{No hull}
         \label{fig:prototype_examples_pets_1}
         \picdims[height=1.25cm]{\textwidth}{1.25cm}{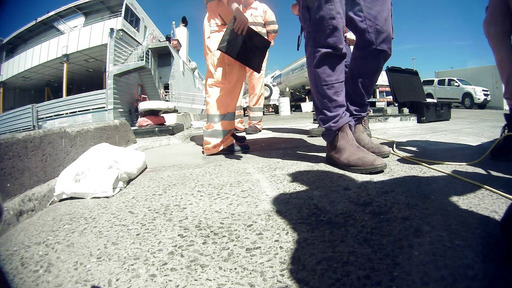}
         \picdims[height=1.25cm]{\textwidth}{1.25cm}{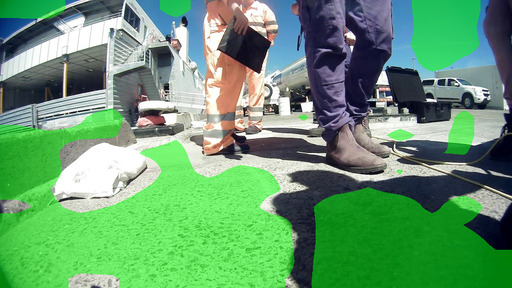}
     \end{subfigure}
        \caption[Segmentation masks predicted by ComFe for the presence of an underwater vessel hull.]{Segmentation masks predicted by ComFe for the presence of an underwater vessel hull. Red areas in the class prediction images depict the hull present class, green areas are for the hull absent class, and the transparent patches are assigned to the background (non-informative) class.}
        \label{fig:salient_features_hulldetect}
\end{figure*}

\begin{figure*}[!tb]
     \centering
     \begin{subfigure}[b]{0.11\textwidth}
         \centering
        \begin{overpic}[height=0.6\textwidth, width=\textwidth]{example_images/1280px-HD_transparent_picture.png}
         \put (0,20) {\parbox{1.5cm}{\centering \small Hull}}
        \end{overpic}
        \begin{overpic}[height=0.6\textwidth, width=\textwidth]{example_images/1280px-HD_transparent_picture.png}
         \put (0,20) {\parbox{1.5cm}{\centering \small No Hull}}
        \end{overpic}
     \end{subfigure}%
     \begin{subfigure}[b]{0.11\textwidth}
         \centering
         \includegraphics[width=\textwidth]{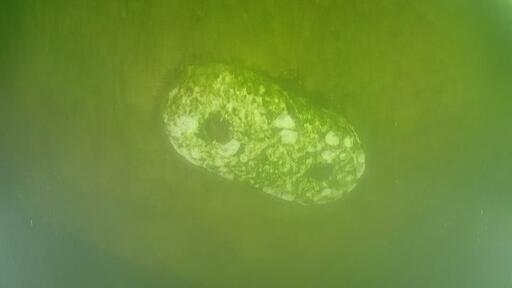}
         \includegraphics[width=\textwidth]{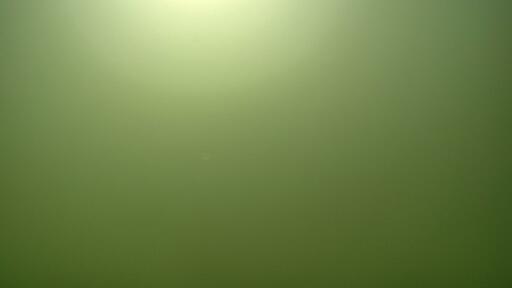}
     \end{subfigure}%
     \begin{subfigure}[b]{0.11\textwidth}
         \centering
         \includegraphics[width=\textwidth]{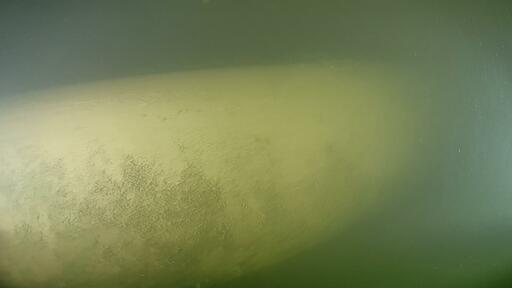}
         \includegraphics[width=\textwidth]{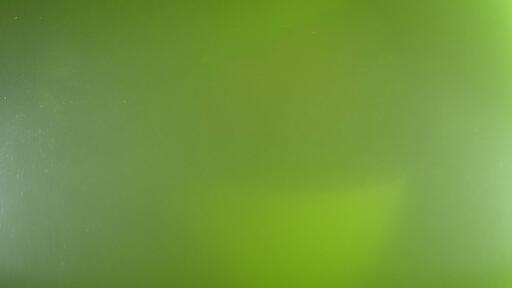}
     \end{subfigure}%
     \begin{subfigure}[b]{0.11\textwidth}
         \centering
         \includegraphics[width=\textwidth]{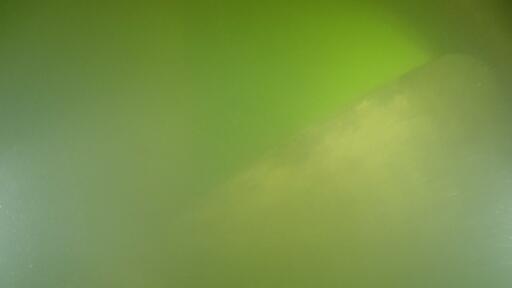}
         \includegraphics[width=\textwidth]{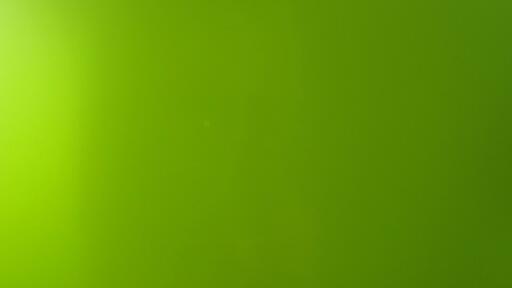}
     \end{subfigure}%
     \begin{subfigure}[b]{0.11\textwidth}
         \centering
         \includegraphics[width=\textwidth]{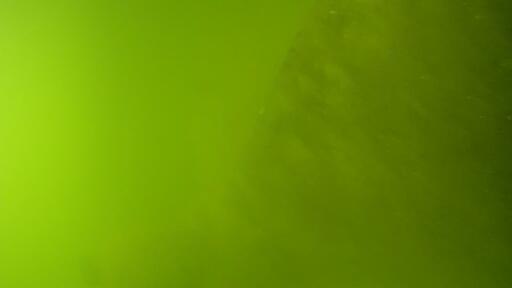}
         \includegraphics[width=\textwidth]{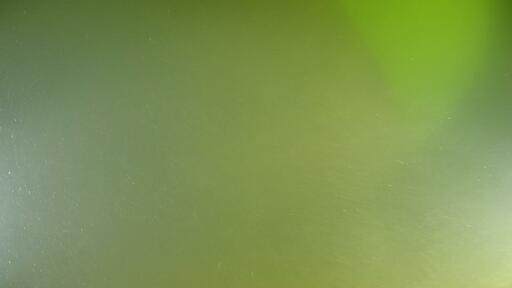}
     \end{subfigure}%
     \begin{subfigure}[b]{0.11\textwidth}
         \centering
         \includegraphics[width=\textwidth]{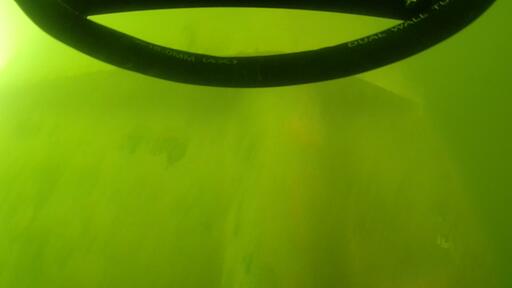}
         \includegraphics[width=\textwidth]{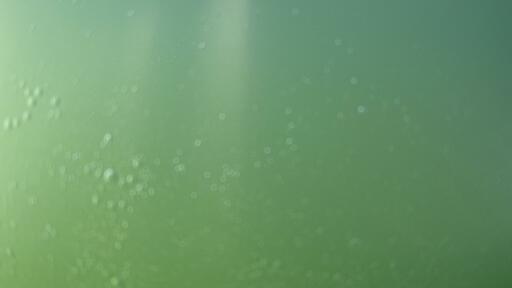}
     \end{subfigure}%
     \begin{subfigure}[b]{0.11\textwidth}
         \centering
         \includegraphics[width=\textwidth]{figures-video/day2-3.mkv/frames/underwater_structure_detected/Frame-16122}
         \includegraphics[width=\textwidth]{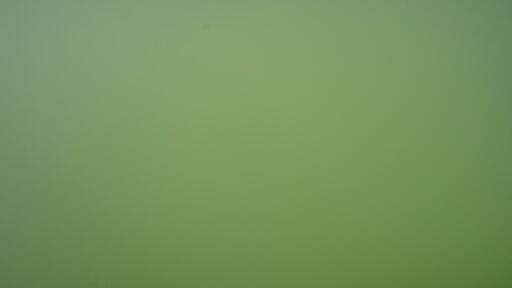}
     \end{subfigure}%
     \begin{subfigure}[b]{0.11\textwidth}
         \centering
         \includegraphics[width=\textwidth]{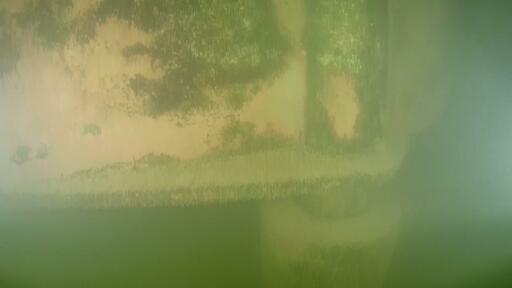}
         \includegraphics[width=\textwidth]{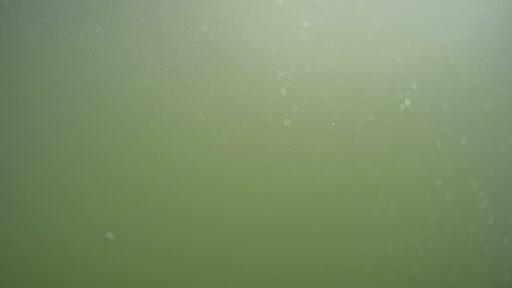}
     \end{subfigure}%
     \begin{subfigure}[b]{0.11\textwidth}
         \centering
         \includegraphics[width=\textwidth]{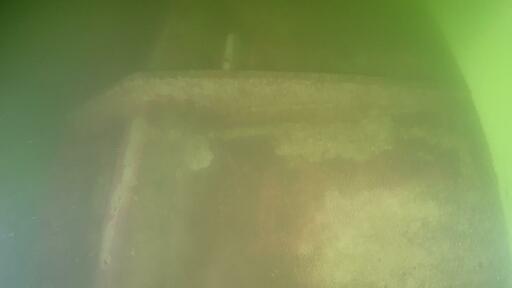}
         \includegraphics[width=\textwidth]{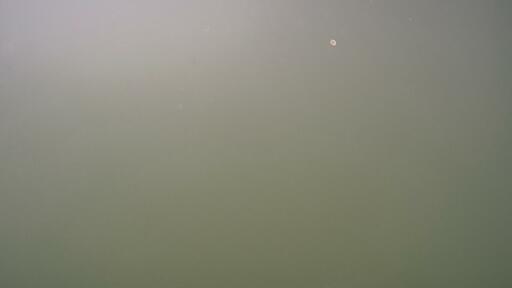}
     \end{subfigure}\\
     \begin{subfigure}[b]{1\textwidth}
         \centering
         \includegraphics[width=\textwidth]{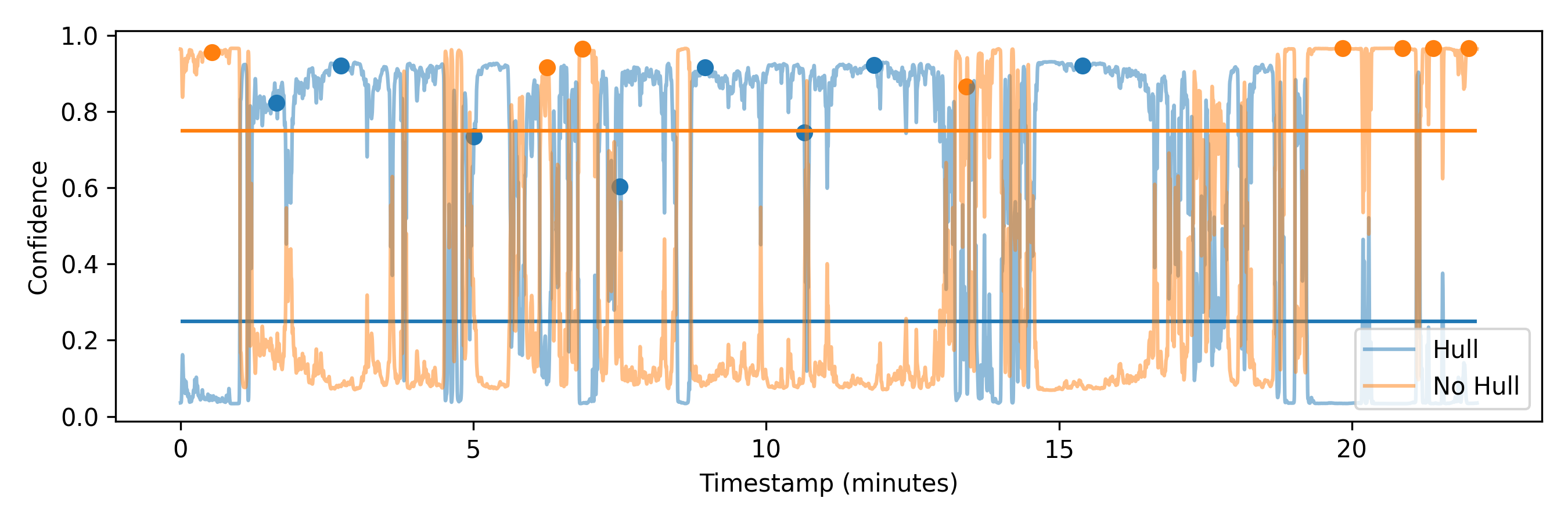}
     \end{subfigure}
        \caption[Analysis of ROV captured video for presence of underwater hull using the ComFe model.]{Analysis of ROV captured video for presence of underwater hull using the ComFe model. The dots in the timeseries represent the frames shown above the plot, which are selected using the SKMPS method. The horizontal lines shows the thresholds used to classify frames as having hull or no hull present.}
        \label{fig:prototype_examples_hull_detect}
\end{figure*}

\begin{figure*}[!h]
     \centering
     \begin{subfigure}[b]{0.11\textwidth}
         \centering
        \begin{overpic}[height=0.6\textwidth, width=\textwidth]{example_images/1280px-HD_transparent_picture.png}
         \put (0,20) {\parbox{1.5cm}{\centering \small Hull}}
        \end{overpic}
        \begin{overpic}[height=0.6\textwidth, width=\textwidth]{example_images/1280px-HD_transparent_picture.png}
         \put (0,20) {\parbox{1.5cm}{\centering \small No Hull}}
        \end{overpic}
     \end{subfigure}%
     \begin{subfigure}[b]{0.11\textwidth}
         \centering
         \includegraphics[width=\textwidth]{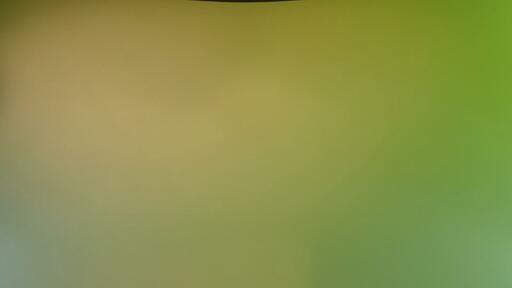}
         \includegraphics[width=\textwidth]{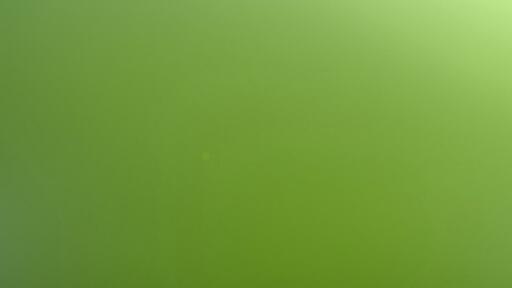}
     \end{subfigure}%
     \begin{subfigure}[b]{0.11\textwidth}
         \centering
         \includegraphics[width=\textwidth]{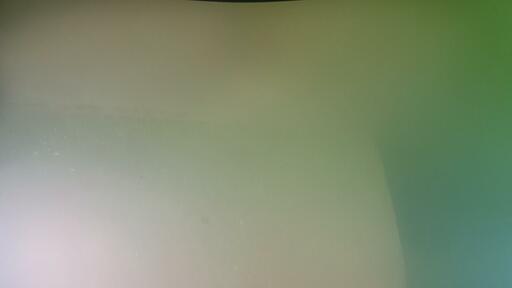}
         \includegraphics[width=\textwidth]{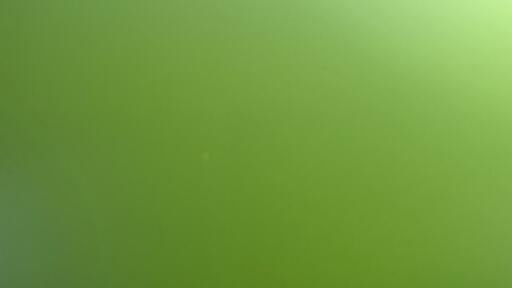}
     \end{subfigure}%
     \begin{subfigure}[b]{0.11\textwidth}
         \centering
         \includegraphics[width=\textwidth]{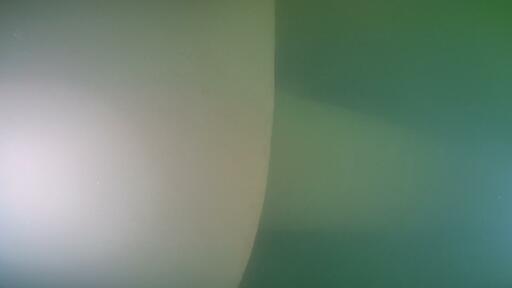}
         \includegraphics[width=\textwidth]{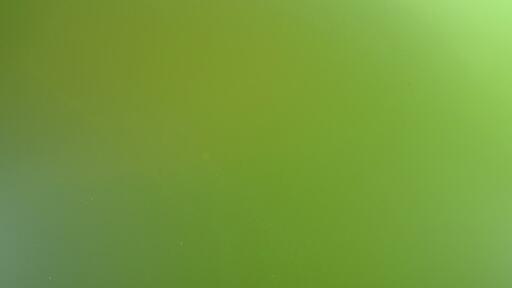}
     \end{subfigure}%
     \begin{subfigure}[b]{0.11\textwidth}
         \centering
         \includegraphics[width=\textwidth]{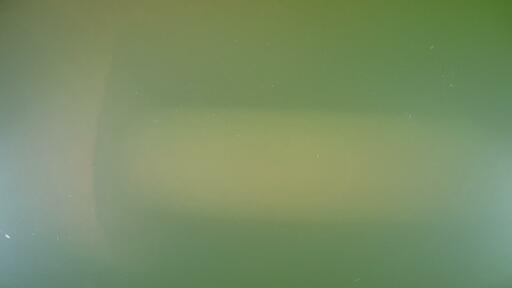}
         \includegraphics[width=\textwidth]{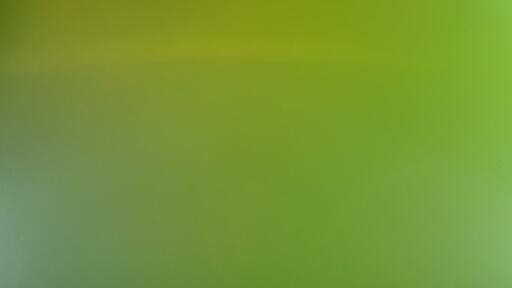}
     \end{subfigure}%
     \begin{subfigure}[b]{0.11\textwidth}
         \centering
         \includegraphics[width=\textwidth]{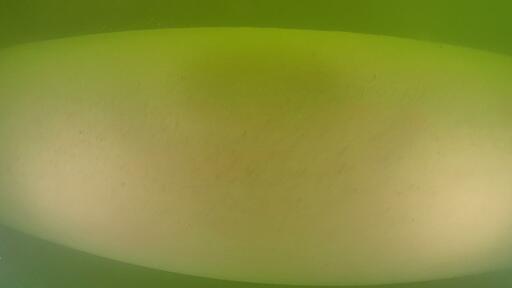}
         \includegraphics[width=\textwidth]{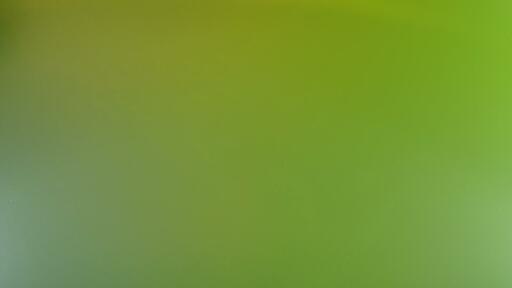}
     \end{subfigure}%
     \begin{subfigure}[b]{0.11\textwidth}
         \centering
         \includegraphics[width=\textwidth]{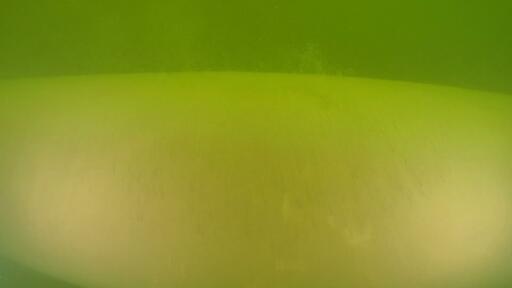}
         \includegraphics[width=\textwidth]{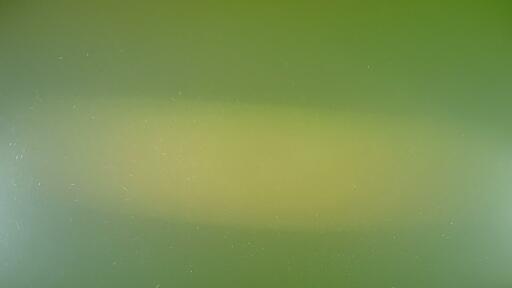}
     \end{subfigure}%
     \begin{subfigure}[b]{0.11\textwidth}
         \centering
         \includegraphics[width=\textwidth]{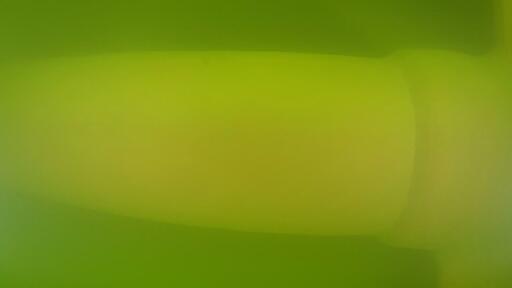}
         \includegraphics[width=\textwidth]{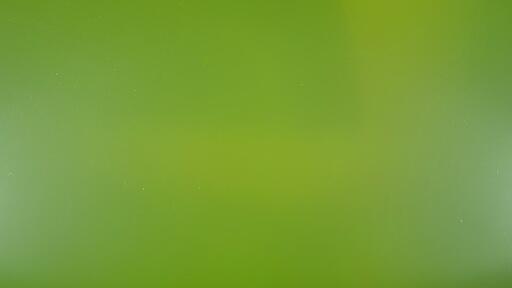}
     \end{subfigure}%
     \begin{subfigure}[b]{0.11\textwidth}
         \centering
         \includegraphics[width=\textwidth]{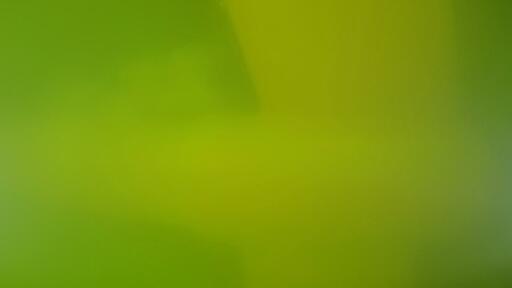}
         \includegraphics[width=\textwidth]{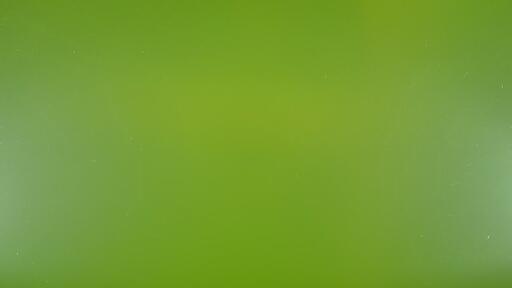}
     \end{subfigure}\\
     \begin{subfigure}[b]{1\textwidth}
         \centering
         \includegraphics[width=\textwidth]{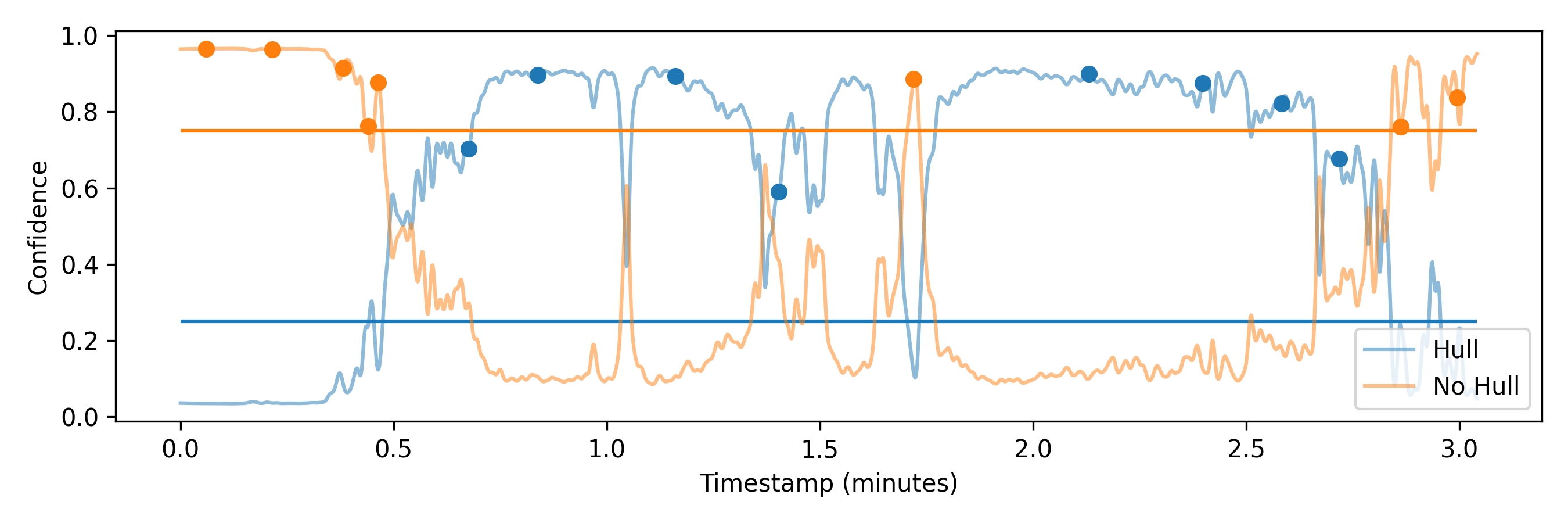}
     \end{subfigure}
        \caption[Analysis of ROV captured video for presence of underwater hull using the ComFe model (smoothed).]{Analysis of ROV captured video for presence of underwater hull using the ComFe model (smoothed). The dots in the timeseries represent the frames shown above the plot, which are selected using the SKMPS method. The horizontal lines shows the thresholds used to classify frames as having hull or no hull present.}
        \label{fig:prototype_examples_hull_detect_smoothed}
\end{figure*}

The models were trained as described in \cref{sec:implementation}, using hull present, hull absent, and a background class. This was an easier challenge to detecting biofouling, and the best model achieved an average precision of 0.984 for detecting imagery not containing a vessel hull (\cref{fig:model_suite_average_precision}). When applying the model to video footage, a confidence threshold of 0.75 was used, which provided a recall of 92.0\% and a precision of 95.1\%. The regions detected as informative under this ComFe model are shown in \cref{fig:salient_features_hulldetect}. This hull detection model performed quite well in the ROV trial, even under challenging conditions, and example transects are shown in \cref{fig:prototype_examples_hull_detect} and \cref{fig:prototype_examples_hull_detect_smoothed}.

\begin{table*}[!b]
\caption[Brief summary of the hull detection dataset.]{Brief summary of the hull detection dataset. We show example training (first row) and testing (second row) images, and a table of the counts of each category in the dataset below. }
\label{tbl:dataset_hulldetect}
\centering
\includegraphics[width=1\textwidth]{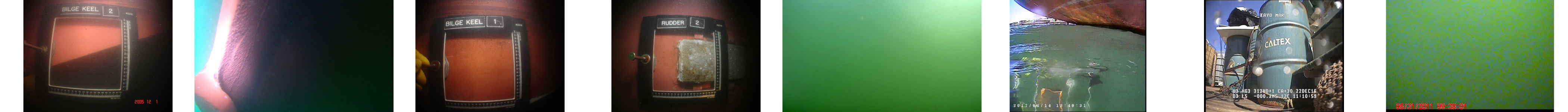}
\includegraphics[width=1\textwidth]{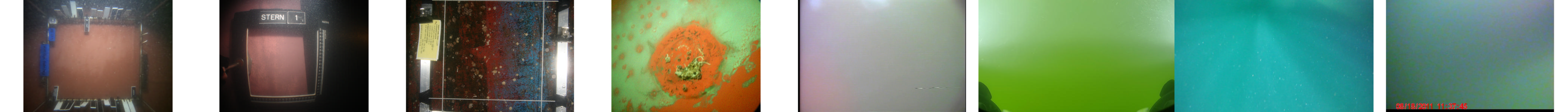}
\begin{tabular}[t]{lrrr}
\toprule
\textbf{Category} & Hull & No Hull & Total \\
\midrule
\textbf{Training}  & 8608 & 2962 & 11570  \\ 
\textbf{Test}  & 8181  & 1528 & 9709\\ 
\bottomrule
\end{tabular}
\end{table*}

\FloatBarrier
\clearpage
\section{Impact of Gaussian kernel smoothing}
\label{app:impact_of_smoothing}

\begin{figure*}[!h]
     \centering
     \begin{subfigure}[b]{0.11\textwidth}
         \centering
        \begin{overpic}[height=0.6\textwidth, width=\textwidth]{example_images/1280px-HD_transparent_picture.png}
         \put (0,20) {\parbox{1.5cm}{\centering \small Hull}}
        \end{overpic}
        \begin{overpic}[height=0.6\textwidth, width=\textwidth]{example_images/1280px-HD_transparent_picture.png}
         \put (0,20) {\parbox{1.5cm}{\centering \small No Hull}}
        \end{overpic}
     \end{subfigure}%
     \begin{subfigure}[b]{0.11\textwidth}
         \centering
         \includegraphics[width=\textwidth]{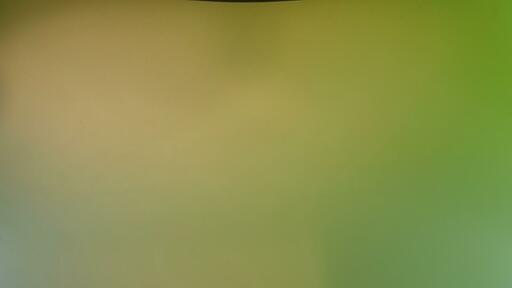}
         \includegraphics[width=\textwidth]{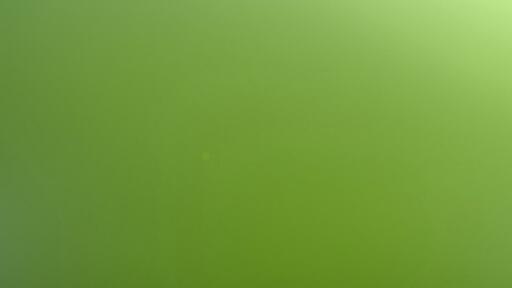}
     \end{subfigure}%
     \begin{subfigure}[b]{0.11\textwidth}
         \centering
         \includegraphics[width=\textwidth]{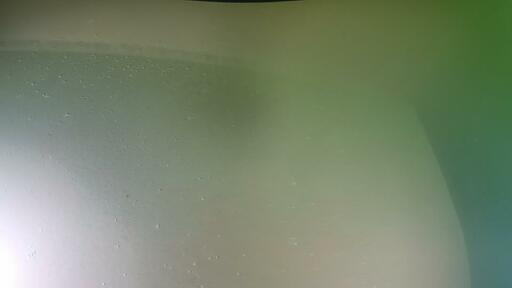}
         \includegraphics[width=\textwidth]{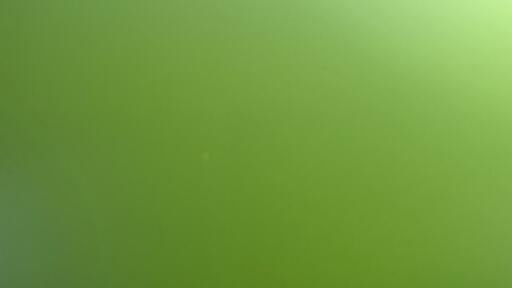}
     \end{subfigure}%
     \begin{subfigure}[b]{0.11\textwidth}
         \centering
         \includegraphics[width=\textwidth]{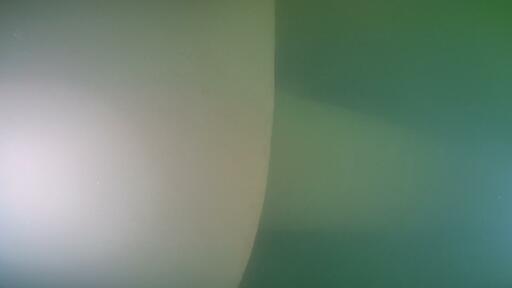}
         \includegraphics[width=\textwidth]{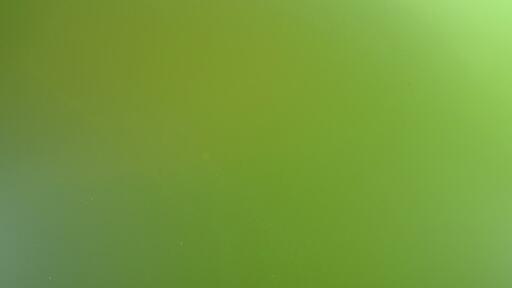}
     \end{subfigure}%
     \begin{subfigure}[b]{0.11\textwidth}
         \centering
         \includegraphics[width=\textwidth]{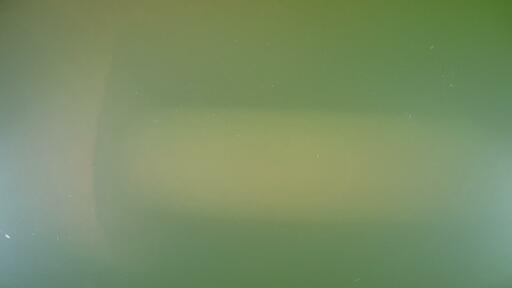}
         \includegraphics[width=\textwidth]{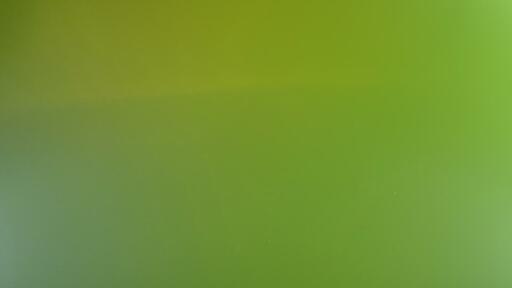}
     \end{subfigure}%
     \begin{subfigure}[b]{0.11\textwidth}
         \centering
         \includegraphics[width=\textwidth]{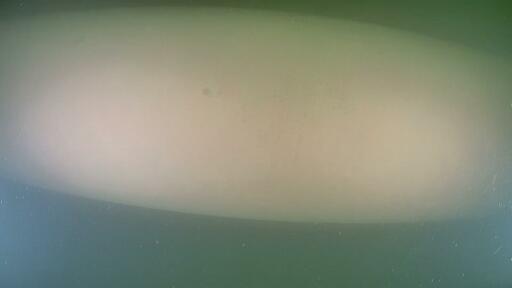}
         \includegraphics[width=\textwidth]{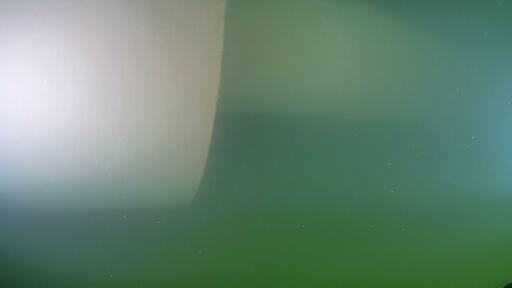}
     \end{subfigure}%
     \begin{subfigure}[b]{0.11\textwidth}
         \centering
         \includegraphics[width=\textwidth]{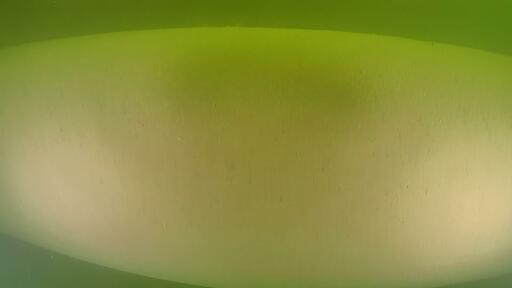}
         \includegraphics[width=\textwidth]{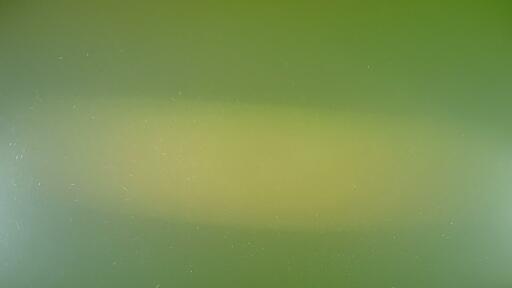}
     \end{subfigure}%
     \begin{subfigure}[b]{0.11\textwidth}
         \centering
         \includegraphics[width=\textwidth]{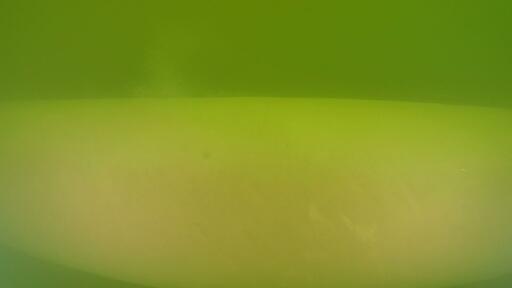}
         \includegraphics[width=\textwidth]{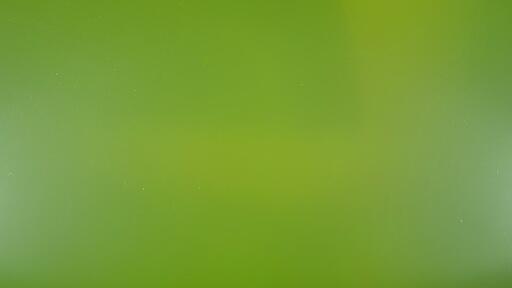}
     \end{subfigure}%
     \begin{subfigure}[b]{0.11\textwidth}
         \centering
         \includegraphics[width=\textwidth]{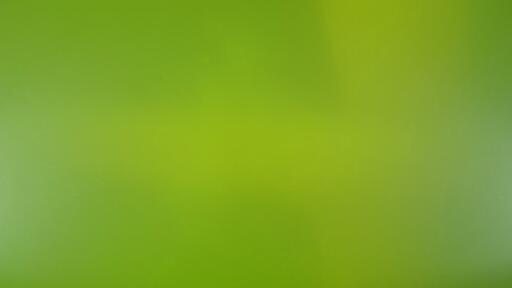}
         \includegraphics[width=\textwidth]{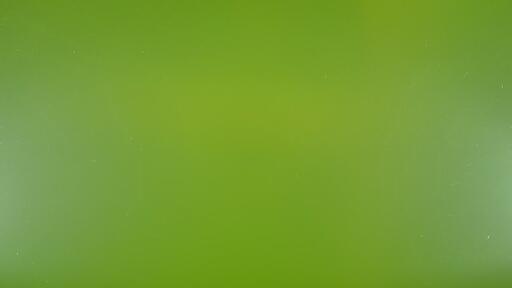}
     \end{subfigure}\\
     \begin{subfigure}[b]{1\textwidth}
         \centering
         \includegraphics[width=\textwidth]{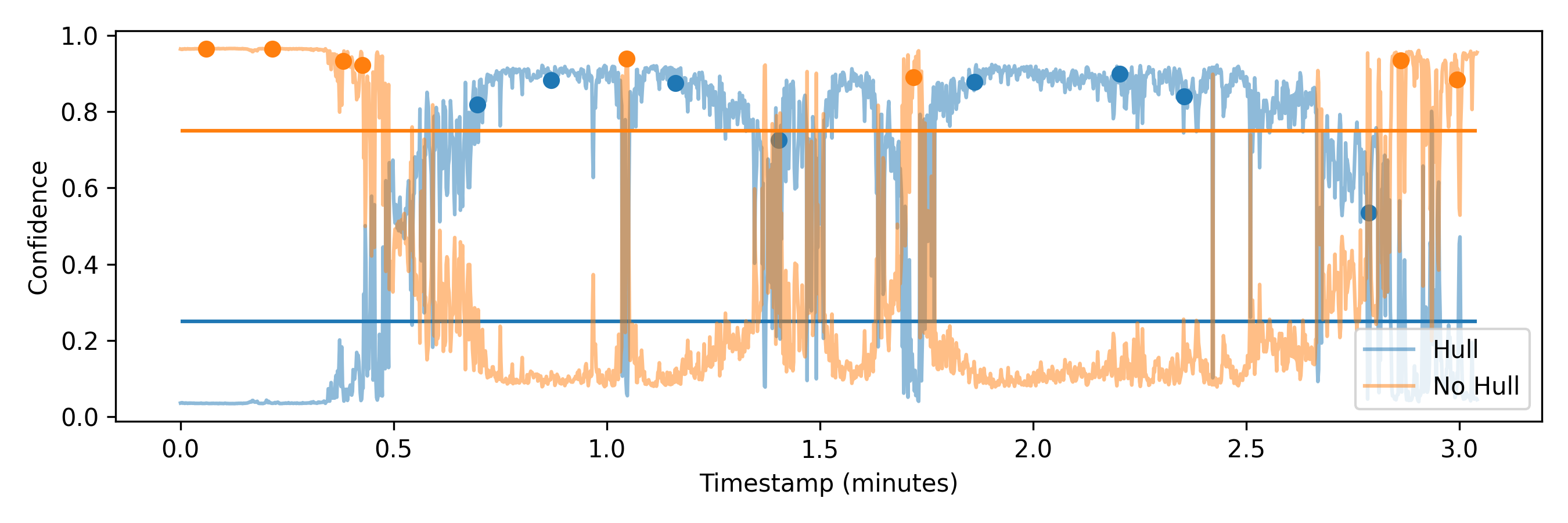}
     \end{subfigure}
        \caption[Analysis of ROV captured video for presence of underwater hull using the ComFe model (unsmoothed).]{Analysis of ROV captured video for presence of underwater hull using the ComFe model (unsmoothed). The dots in the timeseries represent the frames shown above the plot, which are selected using the SKMPS method. The horizontal lines shows the thresholds used to classify frames as having hull or no hull present.}
        \label{fig:prototype_examples_hull_detect_unsmoothed}
\end{figure*}

\end{document}